\documentclass{article}

\usepackage{arxiv}

\usepackage[utf8]{inputenc} 
\usepackage[T1]{fontenc}    
\usepackage{hyperref}       
\usepackage{url}            
\usepackage{array}          
\usepackage{adjustbox}      
\usepackage{booktabs}       
\usepackage{caption}        
\usepackage[table]{xcolor}  
\usepackage{amsfonts}       
\usepackage{amsmath}        
\usepackage{nicefrac}       
\usepackage{microtype}      
\usepackage{cleveref}       
\usepackage{placeins}       
\usepackage{lipsum}         
\usepackage{graphicx}
\usepackage{natbib}
\usepackage{doi}
\usepackage{float}
\usepackage{wrapfig}

\title{DuET: Dual Expert Trajectories for Diffusion Image Editing}

\usepackage{authblk}

\author[1,2]{Lidia Troeshestova}
\author[2]{Alexander Ustyuzhanin}
\author[2]{Sergey Kastryulin}
\affil[1]{HSE University}
\affil[2]{Yandex}

\renewcommand{\shorttitle}{DuET}

\hypersetup{
pdftitle={DuET: Dual Expert Trajectories for Diffusion Image Editing},
pdfsubject={cs.CV},
pdfauthor={Lidia Troeshestova, Alexander Ustyuzhanin, Sergey Kastryulin},
pdfkeywords={diffusion, image editing, instruction editing, T2I},
}

\begin{document}
\maketitle

\begin{abstract}
    Recent diffusion editors perform diverse instruction-based edits while conditioning on the source image at every denoising step.
	Yet persistent source-image conditioning can limit how fully an edit is executed and how natural the result appears, especially when the target scene diverges substantially from the input.
	We introduce \textbf{DuET} (\textbf{Du}al \textbf{E}xpert \textbf{T}rajectories), a training-free inference method that temporarily relaxes source-image conditioning by transitioning through a text-to-image phase before returning to edit mode ($\mathrm{E}\to\mathrm{T2I}\to\mathrm{E}$), allowing the denoising trajectory to move toward the target distribution while retaining the structural benefits of image-conditioned editing.
	Without modifying model weights or increasing sampling cost, DuET consistently improves instruction relevance, semantic fidelity, and perceptual quality across diverse models and benchmarks. Fixed switching schedules obtain these gains at a modest, predictable cost in source-image preservation; we show this cost is not fundamental. A per-edit variant, \emph{Selective DuET}, routes on lightweight attention-probe signals read from the edit trajectory and improves fidelity, naturalness, and artifact scores while keeping source preservation perceptually indistinguishable from the baseline.
\end{abstract}


\section{Introduction}

\begin{figure}[t]
	\centering
	\includegraphics[width=\linewidth]{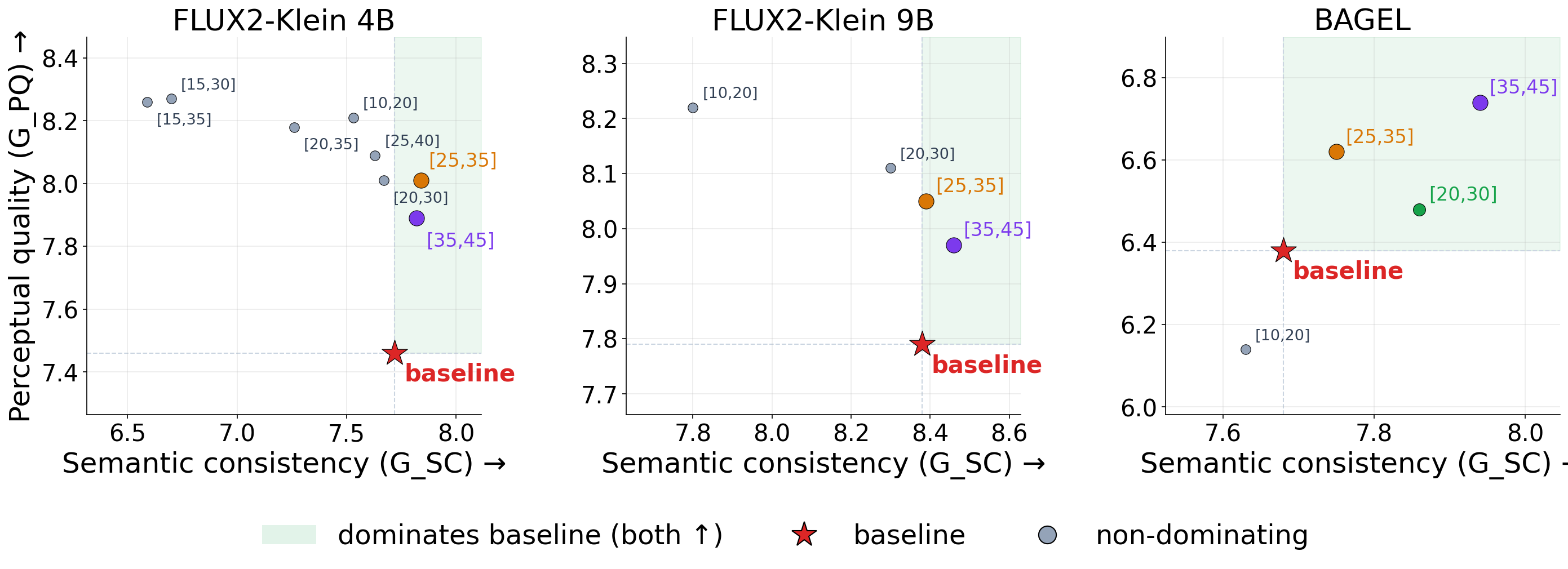}
	\caption{\textbf{DuET ($\mathrm{E}\to\mathrm{T2I}\to\mathrm{E}$) interval placement improves GEdit semantic consistency \emph{and} perceptual quality across all three base models.} G\_SC vs.\ G\_PQ for the no-switching baseline ($\star$) and double-$k$ DuET schedules (T2I active on $[k_1,k_2)$ of the $50$-step trajectory; labels give the interval). Points in the shaded region dominate the baseline on \emph{both} axes; three fixed intervals---$[10,16)$, $[25,35)$ and $[35,45)$---do so for every model.}
	\label{fig:existence}
\end{figure}

Recent diffusion editors such as FLUX2-Klein~\citep{flux2025} perform diverse instruction-based edits while conditioning on the source image throughout denoising~\citep{brooks2023instructpix2pix}.
Yet this persistent source-image anchoring can still limit edit quality: when the target scene diverges substantially from the input, models often under-execute the instruction and produce outputs that are less natural than caption-driven generation, with compositing artifacts, texture clashes, and lighting inconsistencies particularly evident on global rewrites.

Caption-conditioned T2I generation offers a way to relax this tension: without the source image, the denoiser can reorganize the scene more freely toward a target description. Standard edit pipelines nevertheless apply a single mode for the entire trajectory---either edit mode from image and instruction, or T2I mode from a caption---and therefore do not use the model's dual capability to modulate source anchoring during an edit.

We propose to combine both modes within a single edit run. For a short interval mid-trajectory, the sampler switches to caption-only T2I conditioning instead of the source image; it then resumes standard edit mode for the remaining steps. The interval is long enough to improve instruction fidelity and output naturalness when the edit requires substantial scene change, and short enough for image-conditioned editing to recover fine details at the end. This scheduling uses existing dual-mode support without retraining or an additional forward pass: only the active conditioning modality changes over time.

We introduce \textbf{DuET} (\textbf{Du}al \textbf{E}xpert \textbf{T}rajectories), a training-free inference framework for \emph{task and conditioning switching} along a single denoising path. We study when and how to transition between edit mode and caption-only text-to-image (T2I) mode, sweeping over switching intervals, one-way switches, and matched controls that alter only the within-interval conditioning; the main schedules and ablation variants are defined in \cref{sec:method}. The base editor is otherwise unchanged: one forward trajectory, no weight updates. The ``experts'' in DuET are conditioning modes---edit \emph{vs.}\ caption-only---rather than separate networks, complementary to timestep-expert methods that specialize weights while keeping the task fixed (\cref{sec:related}).

Empirically, several switching regimes improve edit relevance and fidelity relative to standard instruction-based sampling; fixed schedules that activate T2I mode earlier or for more steps can modestly reduce source preservation. This trade-off is not fundamental: guided by lightweight attention-probe signals read from the edit trajectory, a per-edit variant---\emph{Selective DuET}---improves fidelity, naturalness, and artifact scores while keeping source preservation perceptually indistinguishable from the baseline. We ablate the constituent conditioning changes and analyze how interval timing and duration shape the fidelity--preservation trade-off (\cref{sec:exps}).

Our contributions are:
\begin{itemize}
	\item A training-free task-switching framework for instruction-based editing that modulates source anchoring by alternating edit mode and caption-only T2I mode along a single denoising trajectory, improving edit relevance and fidelity without finetuning or extra compute.
	\item An evaluation on FLUX2-Klein 4B/9B and BAGEL across several public editing benchmarks, with matched controls and factorial ablations that isolate which conditioning change---dropping the source image \emph{vs.}\ switching the caption---drives the gains and how interval timing shapes the fidelity--preservation trade-off.
	\item \emph{Selective DuET}, a per-edit routing rule over attention-probe signals showing this trade-off is not fundamental: it improves fidelity, naturalness, and artifact scores while keeping source preservation perceptually indistinguishable from the baseline---offered as a proof of existence, not a universal solution.
\end{itemize}

\section{Related Work}
\label{sec:related}

\paragraph{Text-to-image, editing, and unified generation.}
Text-to-image (T2I) diffusion and flow models have driven open-ended image synthesis for several years~\citep{rombach2022ldm, flux2025}. In parallel, instruction-based editing conditions the same class of generators on a source image and a natural-language edit instruction~\citep{brooks2023instructpix2pix}. More recently, unified multimodal models (UMMs) have combined these capabilities within a single architecture and shared weights, supporting both T2I generation and image editing natively~\citep{deng2025bagel}. In practice, however, these capabilities are typically used separately: a model is run either in T2I mode using a caption or in edit mode using an image and instruction. DuET instead combines both modes within a single sampling trajectory. Operating on FLUX2-Klein and BAGEL as black-box generators without finetuning~\citep{flux2025, deng2025bagel}, we switch between edit and T2I conditioning over different sub-intervals of the trajectory.

\paragraph{Timestep experts.}
A substantial line of work specializes the denoiser along the diffusion trajectory. eDiff-I and ERNIE-ViLG~2.0 train an ensemble or mixture of denoising experts for different noise levels~\citep{balaji2022ediffi, feng2023ernievilg}; Switch-DiT, DeMe, and TimeStep~Master route or merge timestep-specific expert weights, often via LoRA~\citep{park2024switchdit, ma2024deme, timestepmaster2025}, and timestep mixtures-of-experts now appear in production-scale models such as Wan~2.2~\citep{wan2025}. All of these specialize the \emph{network weights} for different timesteps while keeping the task and the conditioning fixed. DuET is orthogonal and complementary: its ``experts'' are not different weights but different \emph{conditioning modalities}---edit-mode (source image and instruction) and T2I-mode (a caption of the target)---applied on different sub-intervals of a single trajectory, with no extra parameters and no training. To our knowledge, timestep-expert methods for instruction-based editing have not explored mixing conditioning modalities within a single sampling trajectory.

\paragraph{Inference-time scaling.}
A separate line of work improves a \emph{fixed} generator by spending extra compute at inference time, without retraining. One family runs iterative-refinement loops that re-prompt the model from feedback: OPT2I optimizes the prompt against a consistency score~\citep{manas2024opt2i}, and TIR closes the loop with a multimodal LLM that inspects each output and rewrites the prompt for the next round~\citep{khan2025tir}; more recent variants such as Reflect-DiT condition the transformer on past generations and textual feedback so it can reflect and self-correct~\citep{li2025reflectdit}.
A second family is best-of-$N$ sampling, which draws many candidates and keeps the one a verifier prefers: SANA-1.5 generates a large pool and ranks it with a VLM judge~\citep{xie2025sana15}, while noise-search methods reframe sampling as a search over initial noises guided by reward models~\citep{ma2025inferencescaling}. All of these scale inference compute---typically across multiple forward passes---and often rely on an external evaluator in an outer loop around the sampler. DuET takes the opposite stance: it does not scale compute. Instead, we use an external vision-language model as an \emph{enabler} of modality switching---producing a T2I-style caption of the intended edit so the unified generator can be run in T2I mode on a sub-interval of the trajectory. The sampler still runs in a single pass: edit-mode steps cost what they normally would, and T2I-mode steps cost slightly less (no source-image context); the only change is which conditioning modality is active when.

\section{Method}
\label{sec:method}

\paragraph{Preliminaries.}
Unified multimodal generators expose two conditioning modes under one weight set. \emph{Edit mode} (E) conditions on a source image and an instruction; \emph{text-to-image mode} (T2I) conditions on a caption alone. Let $k \in \{0,\ldots,K\}$ index denoising steps along a single sample trajectory from noise to image; all models in this work use a $K{=}50$-step trajectory. Standard instruction-based editing applies edit mode at every step: given source image $x_{\mathrm{src}}$ and edit instruction $p_{\mathrm{edit}}$,
\begin{equation}
	\text{conditioning}_{\mathrm{edit}}(k) = (x_{\mathrm{src}},\, p_{\mathrm{edit}}) \quad \forall k,
	\label{eq:baseline-edit}
\end{equation}
which we denote as the pure editing schedule $\mathrm{E}$. Standalone text-to-image generation instead keeps T2I mode throughout, conditioning only on a caption $c$:
\begin{equation}
	\text{conditioning}_{\mathrm{t2i}}(k) = c \quad \forall k,
	\label{eq:baseline-t2i}
\end{equation}
denoted $\mathrm{T2I}$. The two tasks thus differ only in what is passed to the same denoiser at each step, not in the underlying weights or sampler.

\begin{figure}[!t]
	\centering
	\includegraphics[width=\linewidth]{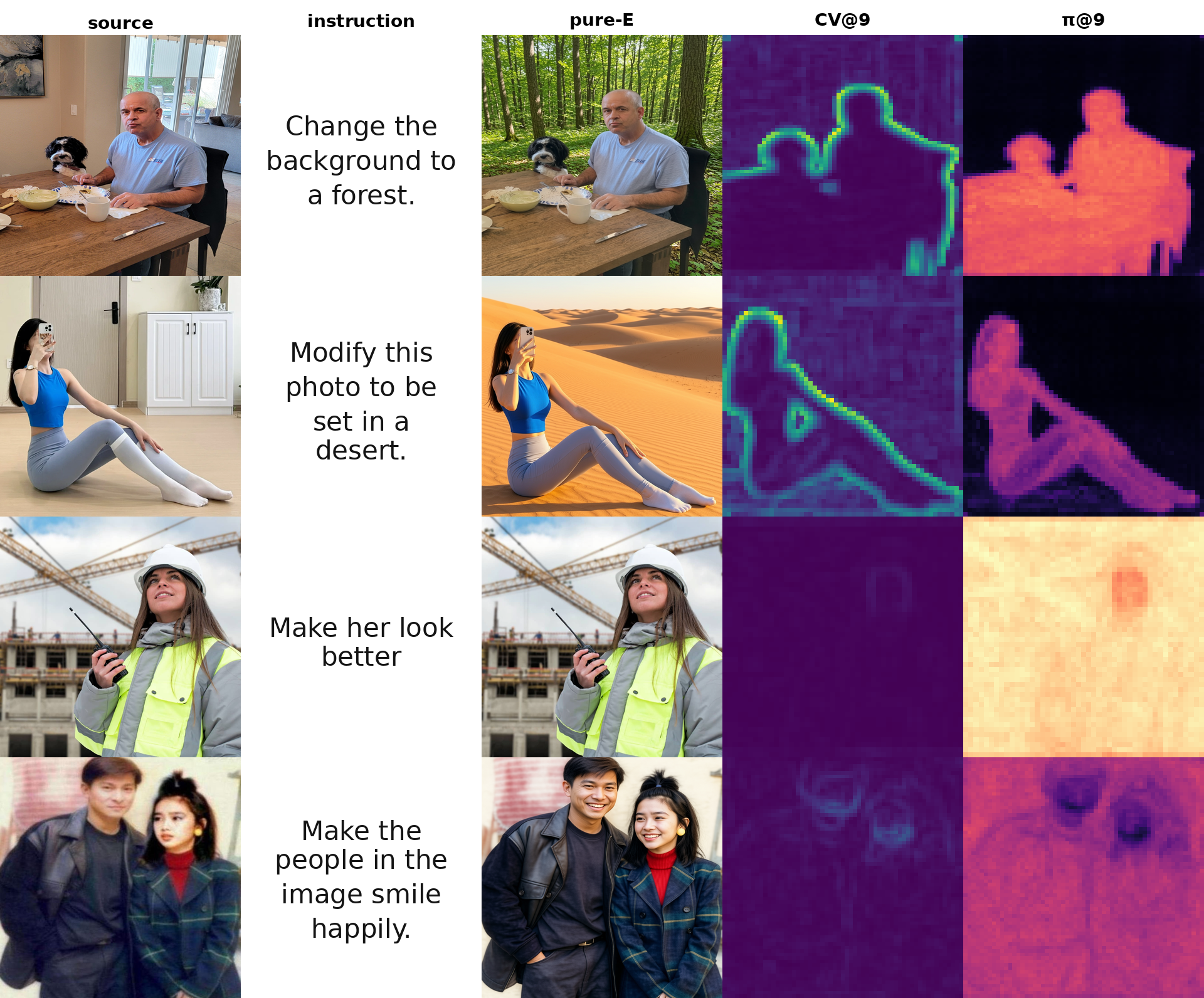}
	\caption{\textbf{What the editor attends to (FLUX2-Klein 4B, pure editing).} For four edits: input, instruction, pure-E result, and two maps read from the per-patch source-image attention $\{a_i\}$ at probe step $k{=}9$ (of 50; \cref{eq:pi-cv})---its local coefficient of variation (\emph{CV@9}, locality) and the preservation-pressure map whose high quantile is $\pi$ (\emph{$\pi$@9}). Low CV \emph{and} low $\pi$ (source already released) flag global rewrites that benefit from a T2I excursion; high CV (localized reliance) or high $\pi$ (strong preservation pressure) flag edits best left to pure editing---the rule behind Selective DuET (\cref{sec:giebench}).}
	\label{fig:observations}
\end{figure}

\paragraph{Where does the editor lean on the source?}
Before switching tasks, we asked what an edit-mode denoiser actually attends to as it works. For each output (latent) patch $i$ we sum its attention over the source-image conditioning tokens and average that across attention heads and a band of deep-semantic layers, giving a per-patch \emph{source-reliance} $a_i$; the map $\{a_i\}_{i=1}^{N}$ is read at an early probe step ($k{=}9$ of the $50$-step trajectory) of a pure-editing pass. We summarize this map with two scalars (\cref{fig:observations}):
\begin{equation}
	\pi \;=\; Q_{0.75}\!\big(\{a_i\}\big), \qquad
	\mathrm{cv} \;=\; \frac{\sigma(\{a_i\})}{\mu(\{a_i\})},
	\label{eq:pi-cv}
\end{equation}
a high-quantile \emph{preservation pressure} $\pi$ (how strongly the still-anchored patches hold onto the source) and a \emph{locality} coefficient of variation $\mathrm{cv}$ (how unevenly that reliance is spread across the canvas). High $\pi$ marks edits that keep large regions bound to the source and will preserve its structure; high $\mathrm{cv}$ marks spatially \emph{localized} edits, where some regions release the source while others stay pinned, as opposed to a spatially uniform (global) change. Empirically these two numbers separate edits that benefit from a T2I excursion from those best left to pure editing, motivating both the fixed DuET schedules below and the adaptive \emph{Selective DuET} variant (\cref{sec:giebench}).

DuET (\textbf{Du}al \textbf{E}xpert \textbf{T}rajectories) is a training-free inference procedure for instruction-based image editing that mixes edit and T2I conditioning within one trajectory---\emph{task switching} along a single sample path rather than specializing network weights.

\paragraph{Task switching.}
Let $\mathcal{C}(k)$ denote the conditioning passed to the denoiser at step~$k$. A \emph{switching schedule} partitions $\{0,\ldots,K\}$ into $m$ contiguous segments $\mathcal{S}_1,\ldots,\mathcal{S}_m$ and assigns each segment a conditioning mode $M_j$ from a finite set $\mathcal{M}$:
\begin{equation}
	\mathcal{C}(k) = \mathrm{cond}(M_j), \quad k \in \mathcal{S}_j,
	\label{eq:task-switch}
\end{equation}
where $\mathrm{cond}(\cdot)$ maps a mode label to the corresponding inputs. Schedules may switch once (e.g., a single-$k$ change from step $k_{\mathrm{sw}}$ onward) or multiple times (e.g., a double-$k$ interval $[k_1,k_2)$ with edit mode before and after). We write a schedule as the sequence of modes active along the trajectory, such as $\mathrm{E}\to\mathrm{T2I}\to\mathrm{E}$ for edit mode outside $[k_1,k_2)$ and a different mode inside. The overall process remains a single forward trajectory; only $\mathcal{C}(k)$ changes at the switch points.

\paragraph{Target caption from a vision-language model.}
When a schedule uses T2I mode, conditioning requires a target caption $c_{\mathrm{tgt}}$ of the \emph{intended} edited outcome. We obtain $c_{\mathrm{tgt}}$ with an external vision-language model (VLM) from the source image $x_{\mathrm{src}}$ and edit instruction $p_{\mathrm{edit}}$. No finetuning or adapter weights are added to the base editor.

\paragraph{Method variants.}
Beyond the pure baselines $\mathrm{E}$ (\cref{eq:baseline-edit}) and $\mathrm{T2I}$ (\cref{eq:baseline-t2i}), we evaluate the conditioning modes in \cref{tab:cond-modes} on the trajectory segments. \textbf{DuET} denotes the main double-$k$ schedule $\mathrm{E}\to\mathrm{T2I}\to\mathrm{E}$: $\mathrm{cond}(\mathrm{E})=(x_{\mathrm{src}}, p_{\mathrm{edit}})$ outside $[k_1,k_2)$ and $\mathrm{cond}(\mathrm{T2I})=c_{\mathrm{tgt}}$ inside. A \textbf{single-$k$} switch applies a mode from step $k_{\mathrm{sw}}$ through the end of sampling (e.g., $\mathrm{E}\to\mathrm{T2I}$ with no return to edit mode). The matched control $\mathrm{E}\to\mathrm{E}^{*}\to\mathrm{E}$ keeps edit mode throughout but replaces $p_{\mathrm{edit}}$ with a VLM-improved instruction $p^{*}$ on $[k_1,k_2)$; it shares the same interval structure and VLM call budget as DuET.

\begin{wraptable}{r}{0.34\textwidth}
	\centering
	\caption{Conditioning inputs per mode.}
	\label{tab:cond-modes}
	\footnotesize
	\setlength{\tabcolsep}{4pt}
	\begin{tabular}{@{}l@{\hspace{0.6em}}l@{}}
		\toprule
		Mode & $\mathrm{cond}(M)$ \\
		\midrule
		$\mathrm{E}$ & $(x_{\mathrm{src}},\, p_{\mathrm{edit}})$ \\
		$\mathrm{E}^{*}$ & $(x_{\mathrm{src}},\, p^{*})$ \\
		$\mathrm{T2I}$ & $c_{\mathrm{tgt}}$ \\
		T2IEP & $p_{\mathrm{edit}}$ \\
		I2IC & $(x_{\mathrm{src}},\, c_{\mathrm{tgt}})$ \\
		\bottomrule
	\end{tabular}
\end{wraptable}%
\ignorespaces To disentangle the two changes DuET makes within $[k_1,k_2)$---dropping the source image and replacing $p_{\mathrm{edit}}$ with $c_{\mathrm{tgt}}$---we include two single-factor variants (\cref{sec:ablation}; full grid in \cref{tab:appx-gedit-qwen}). \textbf{T2IEP} (\emph{T2I with edit prompt}) uses $\mathrm{cond}(\mathrm{T2IEP})=p_{\mathrm{edit}}$: source-image conditioning is removed but the edit instruction is retained. \textbf{I2IC} (\emph{I2I with caption}) uses $\mathrm{cond}(\mathrm{I2IC})=(x_{\mathrm{src}}, c_{\mathrm{tgt}})$: the target caption replaces the instruction while the source image remains. All double-$k$ variants follow the same outside-interval edit conditioning; only the within-interval mode differs.

\paragraph{Switching parameters.}
The placement and width of switching segments are key hyperparameters: when the schedule changes mode, and for how many steps, strongly affects both edit fidelity and source preservation. We grid-search over single-$k$ and double-$k$ configurations and ablate their effect in \cref{sec:exps}.

\paragraph{Selective DuET.}
The schedules above apply a \emph{fixed} interval to every edit, yet \cref{eq:pi-cv} shows edits differ in how much they rely on the source---and that some need no switch at all. \textbf{Selective DuET} makes switching adaptive: it reads the probe signals $(\mathrm{cv},\pi)$ from the ordinary pure-editing pass at an early step and decides \emph{per edit} whether to switch, applying a fixed DuET excursion only to edits that have already released the source broadly and staying in pure edit mode otherwise. The routing adds no forward passes---the probe signals are computed online during the same trajectory---and no retraining; the decision rule and its empirically-calibrated thresholds are given in \cref{sec:giebench}.

\FloatBarrier
\section{Experiments}
\label{sec:exps}

\paragraph{Setup.}
We evaluate DuET on three off-the-shelf unified editors: \textbf{FLUX2-Klein 4B} (primary model; all ablations and the GIE-Bench sweep), \textbf{FLUX2-Klein 9B}, and \textbf{BAGEL}~\citep{flux2025, deng2025bagel}. No model weights are modified, and all models use 50-step sampling. Target captions $c_{\mathrm{tgt}}$ and VLM-improved instructions $p^{*}$ for the $\mathrm{E}^{*}$ control are generated with Gemini-2.5-pro~\citep{comanici2025gemini25}. We report four public benchmarks: \textbf{GEdit}~\citep{liu2025gedit} (GPT-4.1 judge~\citep{openai2025gpt41}; per sample, semantic consistency $\mathrm{G\_SC}{=}\min(\text{fidelity},\text{preservation})$ and perceptual quality $\mathrm{G\_PQ}{=}\min(\text{naturalness},\text{artifacts})$, with overall G\_O, averaged over the dataset), \textbf{ImgEdit}~\citep{ye2025imgedit} (category-wise scores aggregated to an overall; judged by GPT-4o~\citep{openai2024gpt4o}), \textbf{MagicBrush}~\citep{zhang2023magicbrush} (CLIP-T-gen text--image alignment), and \textbf{GIE-Bench}~\citep{qian2025giebench} (functional-correctness Overall judged by GPT-4o, plus preservation SSIM and Unmasked-CLIP, see \cref{tab:main}). GIE-Bench extended sweeps in \cref{fig:giebench} use Gemini-3-Flash (preview)~\citep{google2025gemini3flash} as a judge for functional correctness. The full single-$k$/double-$k$ GIE-Bench grid is run on 4B.

\paragraph{Switching intervals.}
We ablate a broad set of switching schedules on FLUX2-Klein 4B, varying interval placement, mode order (single-$k$ \emph{vs.}\ double-$k$ return to edit mode), and the within-interval conditioning mode itself (\cref{sec:ablation}, \cref{app:results}). For the main reported configuration we then fix the DuET schedule $\mathrm{E}\to\mathrm{T2I}\to\mathrm{E}$ and sweep interval placement across all three base models---several $\Delta k{=}10$ windows and $[10,16)$ on every model, plus a range of interval lengths on 4B (\cref{fig:existence}). For the main comparison we report three intervals---$[10,16)$, $[25,35)$, and $[35,45)$---each of which improves both GEdit semantic consistency and perceptual quality over the no-switching baseline on every model; their scores on ImgEdit, MagicBrush, and GIE-Bench are given in \cref{tab:main}.

\FloatBarrier
\subsection{Fixed-interval DuET: edit fidelity up, preservation down}
\label{sec:main-results}

\Cref{tab:main} and \cref{fig:existence} summarize performance. Complete per-benchmark sweeps are in \cref{app:results}. A fixed DuET interval improves GEdit, ImgEdit, MagicBrush, and GIE-Bench functional correctness on all three base models relative to the pure editing baseline. The $\mathrm{E}\to\mathrm{E}^{*}\to\mathrm{E}$ control indicates that the gains are not explained by improved instructions alone: $\mathrm{E}^{*}$ preserves SSIM and Unmasked-CLIP at near-baseline levels, but does not yield significant improvements in overall edit quality or instruction fidelity (CLIP-T, Func. Corr). DuET's gains come with a consistent preservation cost---SSIM and Unmasked-CLIP drop whenever the T2I switch is active---establishing a correctness--preservation trade-off we examine in \cref{sec:giebench}. The relative ordering of the fixed intervals is model-dependent. On FLUX2-Klein 4B and 9B the short window $[10,16)$ is among the strongest on every fidelity metric (GEdit SC, ImgEdit, MagicBrush, GIE functional correctness) \emph{and} is the best-preserving of the three intervals. On BAGEL the trade-off reappears: $[10,16)$ pushes functional correctness highest but drops preservation the most, while the later windows preserve far more---$[25,35)$ most of all---and $[35,45)$ is strongest on the GEdit and ImgEdit axes. All three windows beat the baseline on the GEdit axes for every model (\cref{fig:existence}).

\begin{table*}[t]
\centering
\caption{\textbf{Main results.} DuET ($\mathrm{E}\to\mathrm{T2I}\to\mathrm{E}$) at fixed intervals $[10,16)$, $[25,35)$ and $[35,45)$ (plus \emph{Selective DuET} on 4B) and the $\mathrm{E}\to\mathrm{E}^{*}\to\mathrm{E}$ ablation at $[25,35)$/$[35,45)$, vs.\ the no-switching baseline on three base models. GEdit: semantic consistency (SC), perceptual quality (PQ), overall (O); ImgEdit: overall; MagicBrush: CLIP-T; GIE-Bench: functional correctness (FC) and preservation (SSIM, Unmasked-CLIP ``U-CLIP'')---all higher${=}$better. DuET consistently improves the edit-quality metrics \emph{at a clear preservation cost} (lower SSIM/U-CLIP), while the $\mathrm{E}^{*}$ ablation keeps preservation near baseline but gains little. Preservation cells are shaded red when SSIM drops more than $0.029$ (the 4B perceptual margin) below that model's baseline.}
\label{tab:main}
\vspace{0.75em}
{\footnotesize
\setlength{\tabcolsep}{3pt}
\centering
\sbox0{\begin{tabular}{llccccccccc}
\toprule
& & \multicolumn{3}{c}{GEdit} & ImgEdit & MagicBrush & \multicolumn{3}{c}{GIE-Bench} \\
\cmidrule(lr){3-5} \cmidrule(lr){6-6} \cmidrule(lr){7-7} \cmidrule(lr){8-10}
& & & & & & & Func.\ Corr. & \multicolumn{2}{c}{Preservation} \\
\cmidrule(lr){8-8} \cmidrule(lr){9-10}
Model & Method & SC & PQ & O & overall & CLIP-T & Overall & SSIM & U-CLIP \\
\midrule
FLUX2-Klein 4B & Baseline & 7.72 & 7.46 & 7.11 & 3.74 & 0.317 & 74.05 & 0.815 & 0.915 \\
 & $\mathrm{E}\to\mathrm{E}^{*}\to\mathrm{E}$ $[25,35)$ & \cellcolor[RGB]{247,252,248}7.77 & \cellcolor[RGB]{253,254,254}7.47 & \cellcolor[RGB]{249,253,250}7.16 & \cellcolor[RGB]{255,199,206}3.69 & 0.317 & \cellcolor[RGB]{255,199,206}73.13 & 0.805 & 0.914 \\
 & $\mathrm{E}\to\mathrm{E}^{*}\to\mathrm{E}$ $[35,45)$ & \cellcolor[RGB]{253,254,253}7.73 & 7.46 & \cellcolor[RGB]{255,199,206}7.09 & \cellcolor[RGB]{255,243,245}3.73 & 0.317 & \cellcolor[RGB]{255,247,248}73.93 & 0.805 & 0.913 \\
 & DuET $[10,16)$ & \cellcolor[RGB]{198,239,206}8.08 & \cellcolor[RGB]{209,242,215}7.90 & \cellcolor[RGB]{198,239,206}7.67 & \cellcolor[RGB]{207,241,214}3.99 & \cellcolor[RGB]{198,239,206}0.322 & \cellcolor[RGB]{198,239,206}76.46 & \cellcolor[RGB]{255,199,206}0.781 & \cellcolor[RGB]{255,199,206}0.906 \\
 & DuET $[25,35)$ & \cellcolor[RGB]{236,249,238}7.84 & \cellcolor[RGB]{198,239,206}8.01 & \cellcolor[RGB]{211,242,217}7.54 & \cellcolor[RGB]{198,239,206}4.04 & \cellcolor[RGB]{209,242,215}0.321 & \cellcolor[RGB]{231,248,235}75.03 & \cellcolor[RGB]{255,199,206}0.749 & \cellcolor[RGB]{255,199,206}0.900 \\
 & DuET $[35,45)$ & \cellcolor[RGB]{239,250,241}7.82 & \cellcolor[RGB]{210,242,216}7.89 & \cellcolor[RGB]{224,246,228}7.41 & \cellcolor[RGB]{220,245,225}3.92 & \cellcolor[RGB]{243,251,245}0.318 & \cellcolor[RGB]{255,199,206}73.13 & \cellcolor[RGB]{255,199,206}0.778 & \cellcolor[RGB]{255,199,206}0.905 \\
 & Selective DuET & \cellcolor[RGB]{221,245,226}7.93 & \cellcolor[RGB]{230,248,233}7.70 & \cellcolor[RGB]{223,246,227}7.42 & \cellcolor[RGB]{226,247,230}3.89 & \cellcolor[RGB]{232,248,235}0.319 & \cellcolor[RGB]{215,243,221}75.72 & 0.786 & 0.908 \\
\midrule
FLUX2-Klein 9B & Baseline & 8.38 & 7.79 & 7.74 & 4.02 & 0.319 & 77.94 & 0.939 & 0.928 \\
 & $\mathrm{E}\to\mathrm{E}^{*}\to\mathrm{E}$ $[25,35)$ & \cellcolor[RGB]{214,243,220}8.50 & \cellcolor[RGB]{255,199,206}7.77 & \cellcolor[RGB]{234,249,237}7.82 & \cellcolor[RGB]{239,250,241}4.08 & 0.319 & \cellcolor[RGB]{240,250,242}78.78 & 0.939 & 0.928 \\
 & $\mathrm{E}\to\mathrm{E}^{*}\to\mathrm{E}$ $[35,45)$ & \cellcolor[RGB]{238,250,240}8.43 & \cellcolor[RGB]{255,199,206}7.77 & \cellcolor[RGB]{244,252,246}7.78 & \cellcolor[RGB]{242,251,243}4.07 & 0.319 & \cellcolor[RGB]{236,249,238}79.04 & 0.939 & 0.927 \\
 & DuET $[10,16)$ & \cellcolor[RGB]{198,239,206}8.55 & \cellcolor[RGB]{228,247,232}7.91 & \cellcolor[RGB]{198,239,206}7.96 & \cellcolor[RGB]{200,239,208}4.23 & \cellcolor[RGB]{198,239,206}0.324 & \cellcolor[RGB]{198,239,206}81.28 & \cellcolor[RGB]{255,199,206}0.900 & \cellcolor[RGB]{255,199,206}0.922 \\
 & DuET $[25,35)$ & \cellcolor[RGB]{251,254,252}8.39 & \cellcolor[RGB]{198,239,206}8.05 & \cellcolor[RGB]{208,241,214}7.92 & \cellcolor[RGB]{198,239,206}4.24 & \cellcolor[RGB]{220,245,225}0.322 & \cellcolor[RGB]{212,243,218}80.44 & \cellcolor[RGB]{255,199,206}0.827 & \cellcolor[RGB]{255,199,206}0.912 \\
 & DuET $[35,45)$ & \cellcolor[RGB]{228,247,231}8.46 & \cellcolor[RGB]{215,243,221}7.97 & \cellcolor[RGB]{208,241,214}7.92 & \cellcolor[RGB]{213,243,219}4.18 & \cellcolor[RGB]{243,251,245}0.320 & \cellcolor[RGB]{224,246,229}79.70 & \cellcolor[RGB]{255,199,206}0.858 & \cellcolor[RGB]{255,199,206}0.915 \\
\midrule
BAGEL & Baseline & 7.68 & 6.38 & 6.64 & 3.32 & 0.320 & 67.16 & 0.960 & 0.928 \\
 & $\mathrm{E}\to\mathrm{E}^{*}\to\mathrm{E}$ $[25,35)$ & \cellcolor[RGB]{246,252,247}7.72 & \cellcolor[RGB]{251,254,252}6.40 & \cellcolor[RGB]{246,252,247}6.69 & \cellcolor[RGB]{252,254,253}3.33 & 0.320 & \cellcolor[RGB]{255,222,226}66.42 & 0.960 & 0.928 \\
 & $\mathrm{E}\to\mathrm{E}^{*}\to\mathrm{E}$ $[35,45)$ & \cellcolor[RGB]{255,199,206}7.67 & \cellcolor[RGB]{248,253,249}6.42 & \cellcolor[RGB]{251,254,252}6.66 & \cellcolor[RGB]{255,199,206}3.30 & 0.320 & \cellcolor[RGB]{255,199,206}65.89 & 0.960 & 0.928 \\
 & DuET $[10,16)$ & \cellcolor[RGB]{213,243,219}7.87 & \cellcolor[RGB]{243,251,245}6.45 & \cellcolor[RGB]{227,247,231}6.81 & \cellcolor[RGB]{227,247,231}3.44 & \cellcolor[RGB]{198,239,206}0.324 & \cellcolor[RGB]{198,239,206}73.03 & \cellcolor[RGB]{255,199,206}0.786 & \cellcolor[RGB]{255,199,206}0.908 \\
 & DuET $[25,35)$ & \cellcolor[RGB]{239,250,241}7.75 & \cellcolor[RGB]{217,244,222}6.62 & \cellcolor[RGB]{222,245,227}6.84 & \cellcolor[RGB]{211,242,217}3.51 & \cellcolor[RGB]{226,247,230}0.322 & \cellcolor[RGB]{220,245,225}70.71 & 0.937 & 0.924 \\
 & DuET $[35,45)$ & \cellcolor[RGB]{198,239,206}7.94 & \cellcolor[RGB]{198,239,206}6.74 & \cellcolor[RGB]{198,239,206}6.99 & \cellcolor[RGB]{198,239,206}3.57 & 0.320 & \cellcolor[RGB]{236,249,239}69.02 & \cellcolor[RGB]{255,199,206}0.922 & \cellcolor[RGB]{255,199,206}0.923 \\
\bottomrule
\end{tabular}}
\ifdim\wd0>\linewidth\relax
  \resizebox{\linewidth}{!}{\usebox0}
\else
  \usebox0
\fi
}
\end{table*}

\begin{figure}[!htb]
\centering

\begingroup
\sffamily
\setlength{\tabcolsep}{0pt}
\renewcommand{\arraystretch}{0}
\setlength{\fboxsep}{0pt}
\newlength{\qinstr}
\newlength{\qcell}
\newlength{\qcelltall}
\setlength{\qinstr}{0.110\linewidth}
\setlength{\qcell}{0.178\linewidth}
\setlength{\qcelltall}{1.472\qcell}
\providecommand{\qimg}[1]{%
  \parbox[c][\qcell][c]{\qcell}{\centering\includegraphics[width=\qcell,height=\qcell,keepaspectratio]{#1}}%
}
\providecommand{\qimgtall}[1]{%
  \parbox[c][\qcelltall][c]{\qcell}{\centering\includegraphics[width=\qcell]{#1}}%
}
\providecommand{\qmissing}{%
  \parbox[c][\qcell][c]{\qcell}{\centering\textcolor{black!10}{\rule{\qcell}{\qcell}}}%
}
\providecommand{\qmissingtall}{%
  \parbox[c][\qcelltall][c]{\qcell}{\centering\textcolor{black!10}{\rule{\qcell}{\qcelltall}}}%
}
\providecommand{\qtextpad}{\vspace{2pt}\vspace{0pt}}
\providecommand{\qtxtinstr}[1]{%
  \parbox[c][\qcell][t]{\dimexpr\qinstr-2pt\relax}{\qtextpad\raggedright #1\par}%
}
\providecommand{\qtxtinstrtall}[1]{%
  \parbox[c][\qcelltall][t]{\dimexpr\qinstr-2pt\relax}{\qtextpad\raggedright #1\par}%
}
\providecommand{\qtxtcap}[1]{%
  \parbox[c][\qcell][t]{\dimexpr\qcell-2pt\relax}{\qtextpad\raggedright #1\par}%
}
\providecommand{\qtxtcaptall}[1]{%
  \parbox[c][\qcelltall][t]{\dimexpr\qcell-2pt\relax}{\qtextpad\raggedright #1\par}%
}
\providecommand{\qhdrinstr}[1]{%
  \parbox[t]{\dimexpr\qinstr-2pt\relax}{\centering\qtextpad #1\par}%
}
\providecommand{\qhdr}[1]{%
  \parbox[t]{\dimexpr\qcell-2pt\relax}{\centering\qtextpad #1\par}%
}

\begin{tabular}{@{}>{\raggedright\arraybackslash}m{\qinstr}*{5}{>{\centering\arraybackslash}m{\qcell}}@{}}
\qhdrinstr{\fontsize{8.5}{9.5}\selectfont\bfseries instruction} & \qhdr{\fontsize{8.5}{9.5}\selectfont\bfseries source} & \qhdr{\fontsize{8.5}{9.5}\selectfont\bfseries pure-E} & \qhdr{\fontsize{8.5}{9.5}\selectfont\bfseries T2I caption} & \qhdr{\fontsize{8.5}{9.5}\selectfont\bfseries E$\to$T2I $k{=}10$} & \qhdr{\fontsize{8.5}{9.5}\selectfont\bfseries DuET $[10,20)$} \\
\noalign{\vskip 1pt}
\qtxtinstr{\fontsize{7.5}{8.5}\selectfont Replace the house in the image with a giant tree.} & \qimg{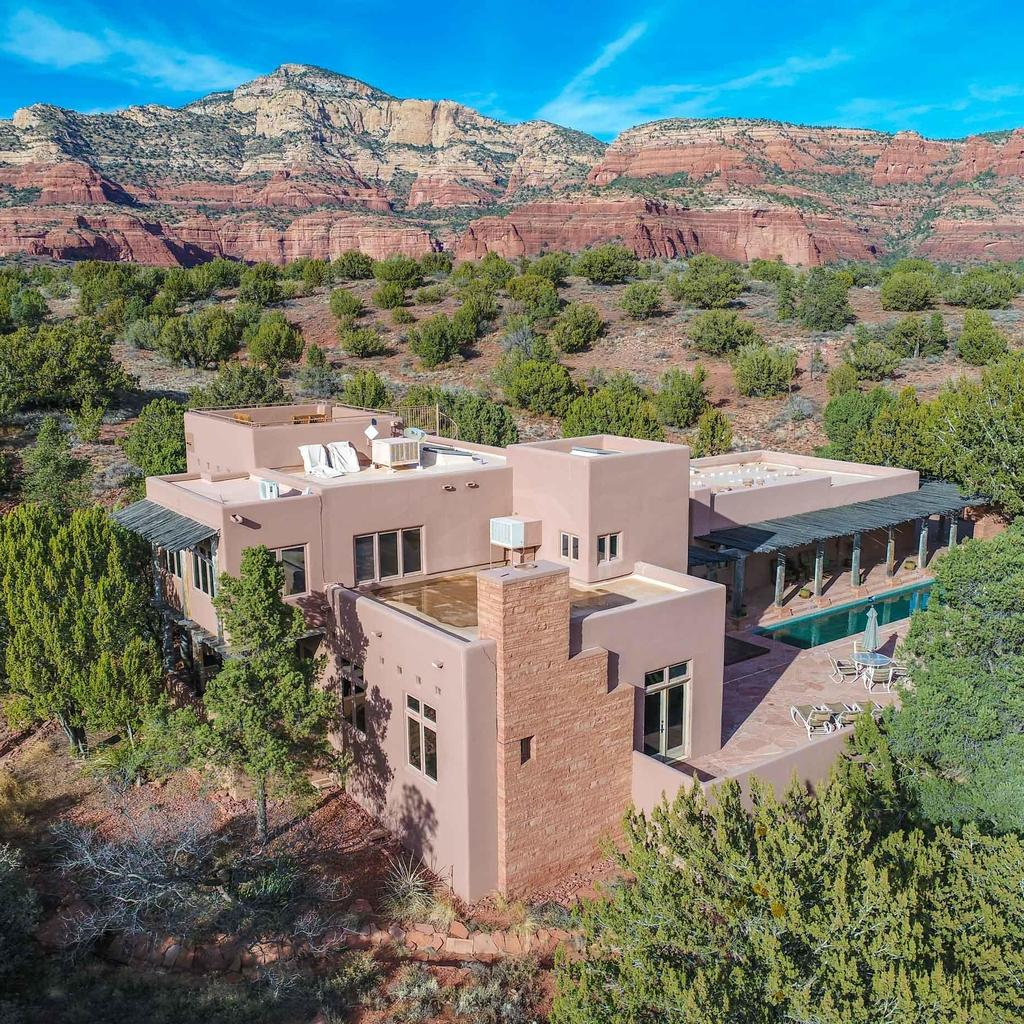} & \qimg{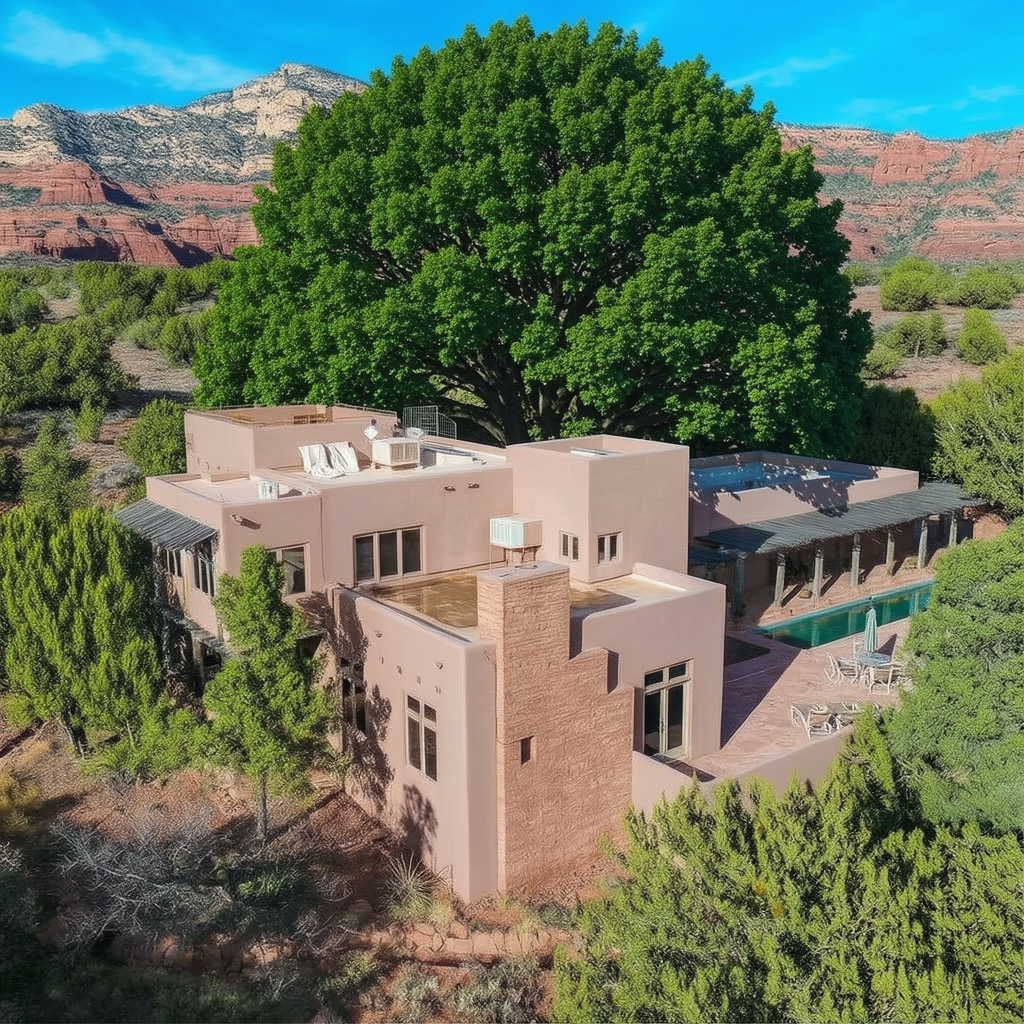} & \qtxtcap{\fontsize{6.5}{7.3}\selectfont An aerial, high-angle drone photograph of a majestic desert landscape where a single, colossal, ancient tree with a massive, thick trunk and a sprawling, dense\,\ldots{}} & \qimg{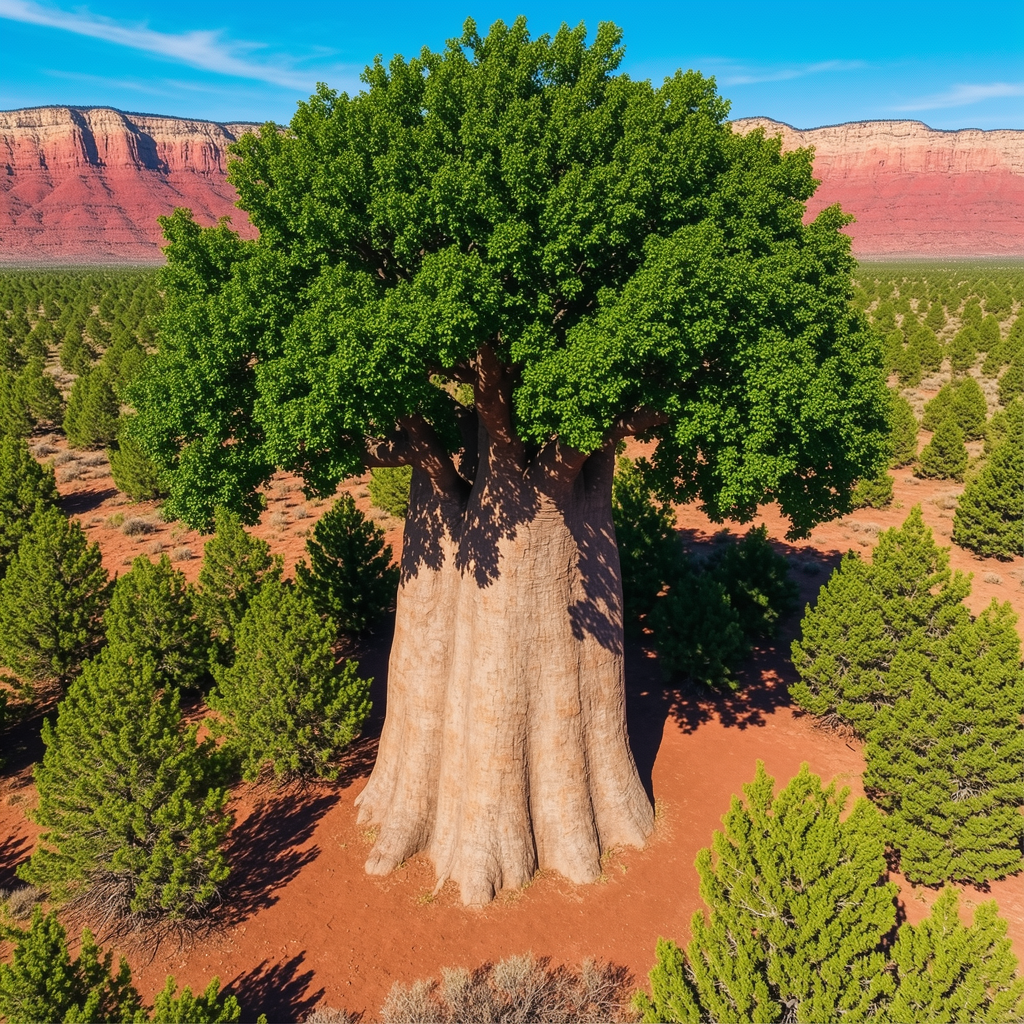} & \qimg{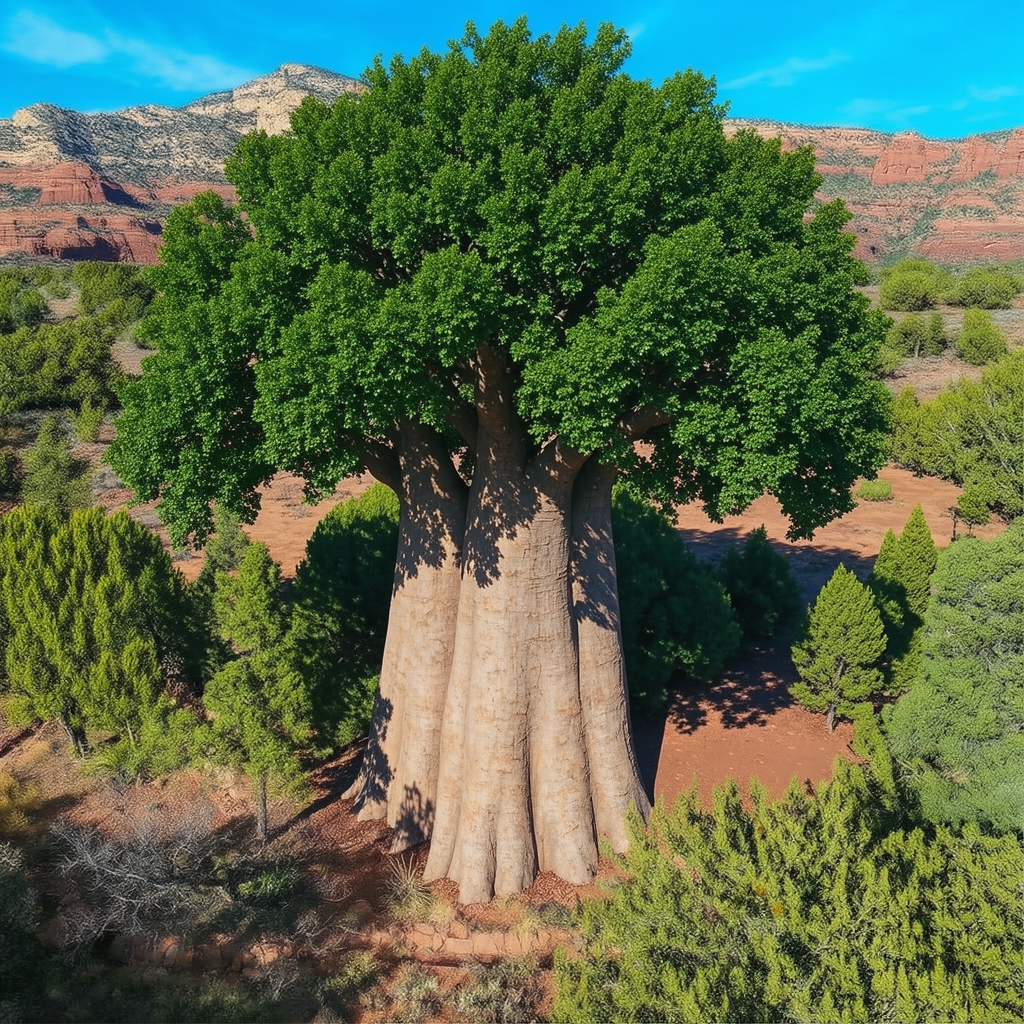} \\
\qtxtinstr{\fontsize{7.5}{8.5}\selectfont Replace the blue bird in the image with a red fox.} & \qimg{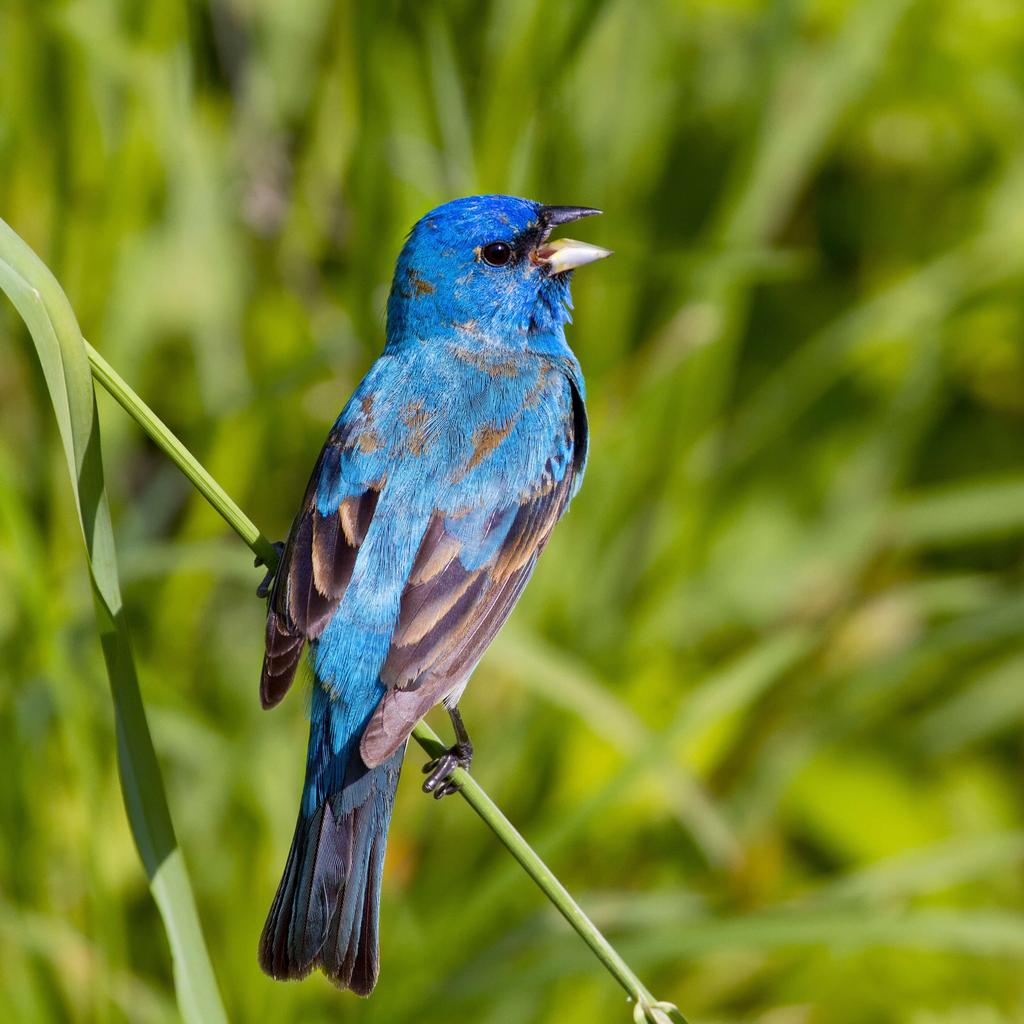} & \qimg{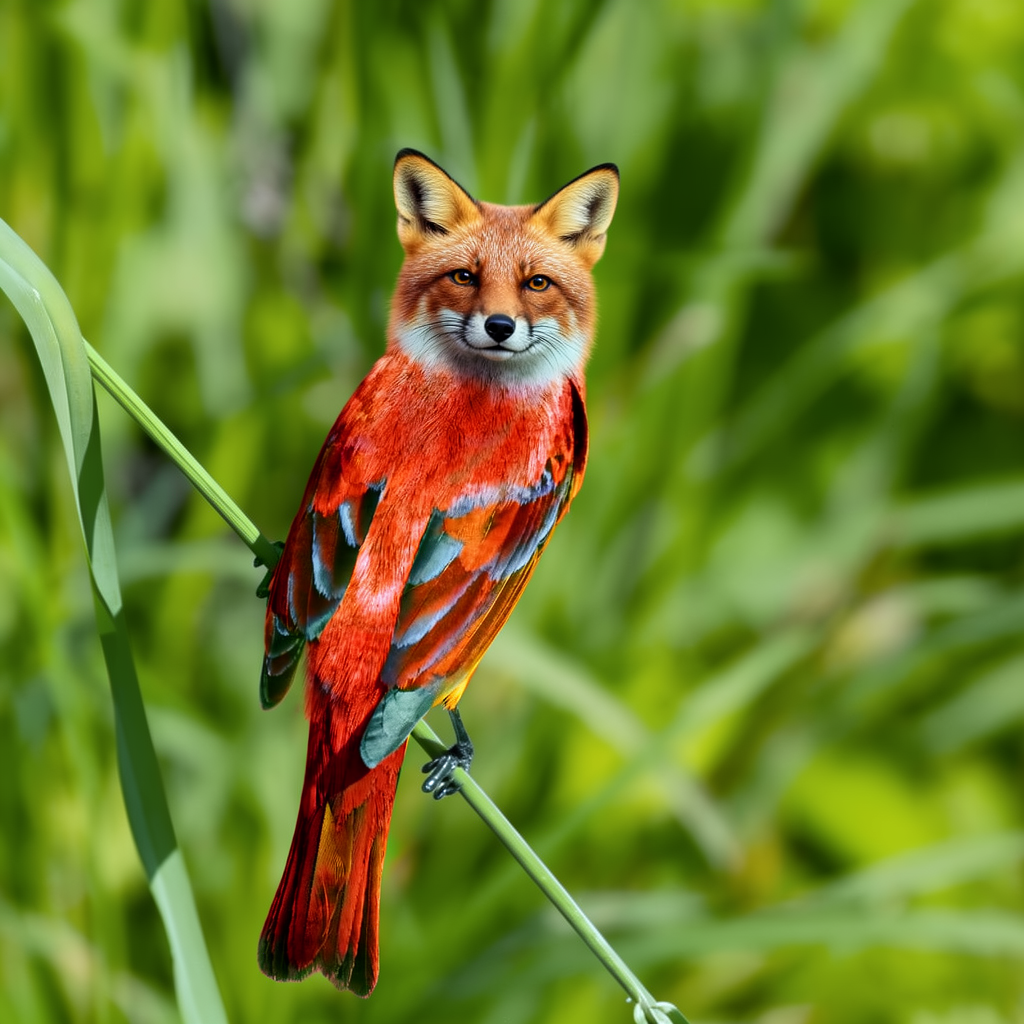} & \qtxtcap{\fontsize{6.5}{7.3}\selectfont A close-up, square-framed wildlife photograph of a vibrant red fox sitting amidst tall, lush green grass. The fox's body is angled away from the camera, but its\,\ldots{}} & \qimg{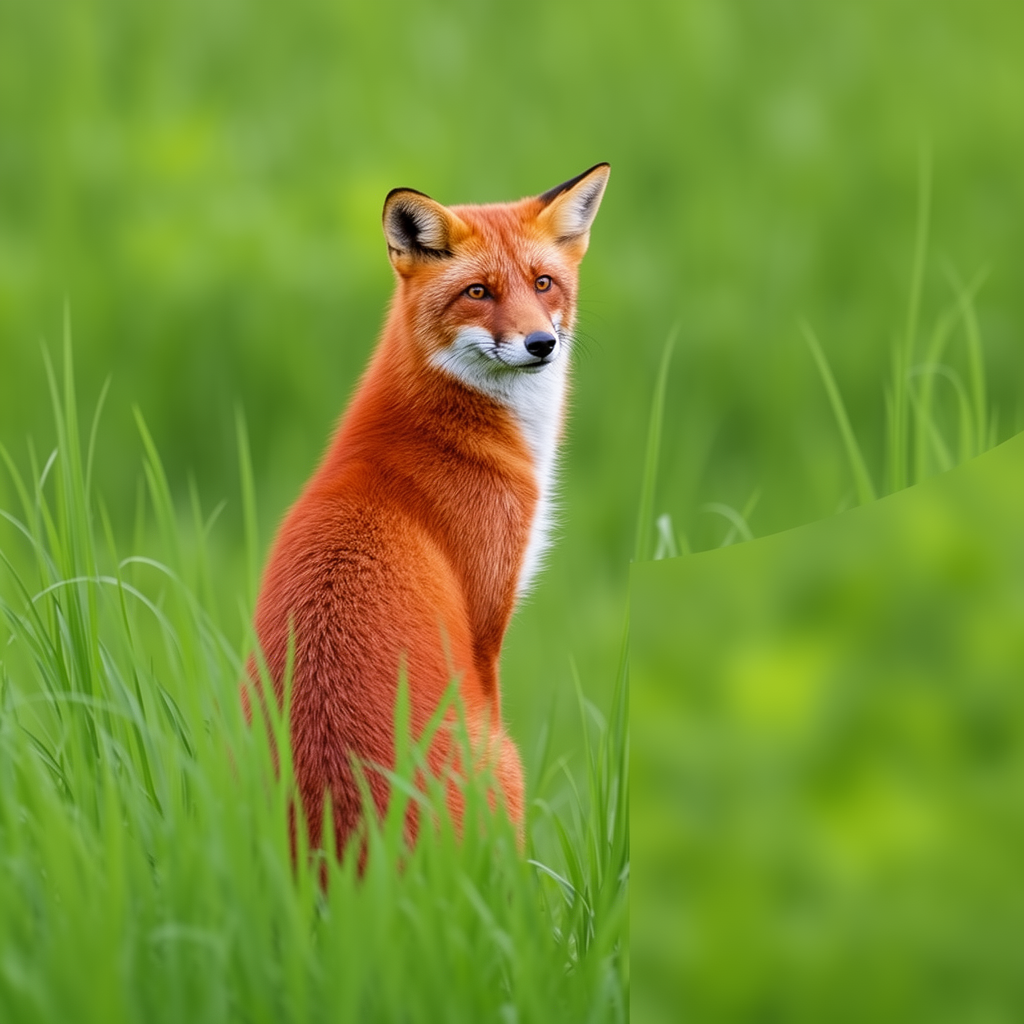} & \qimg{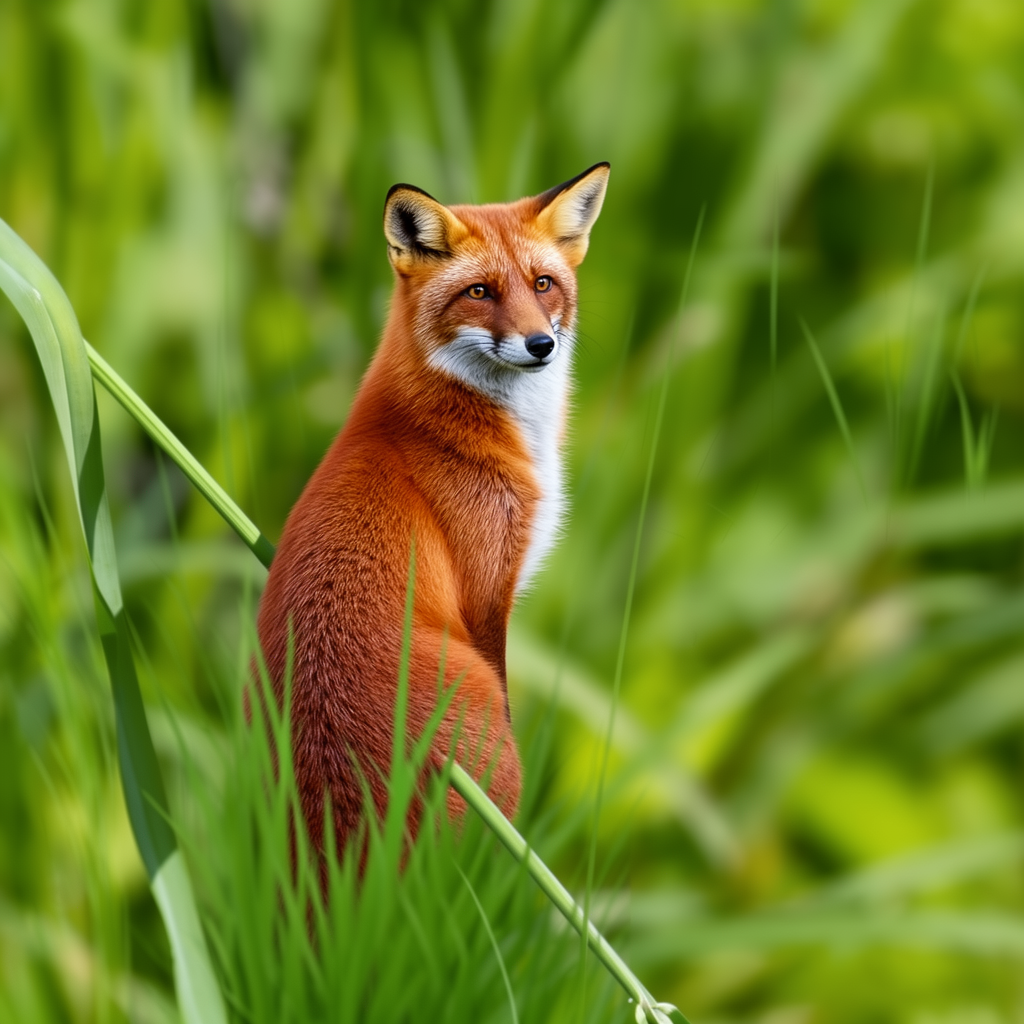} \\
\qtxtinstr{\fontsize{6.8}{7.8}\selectfont Replace the green armchair in the image with a wooden bookshelf.} & \qimg{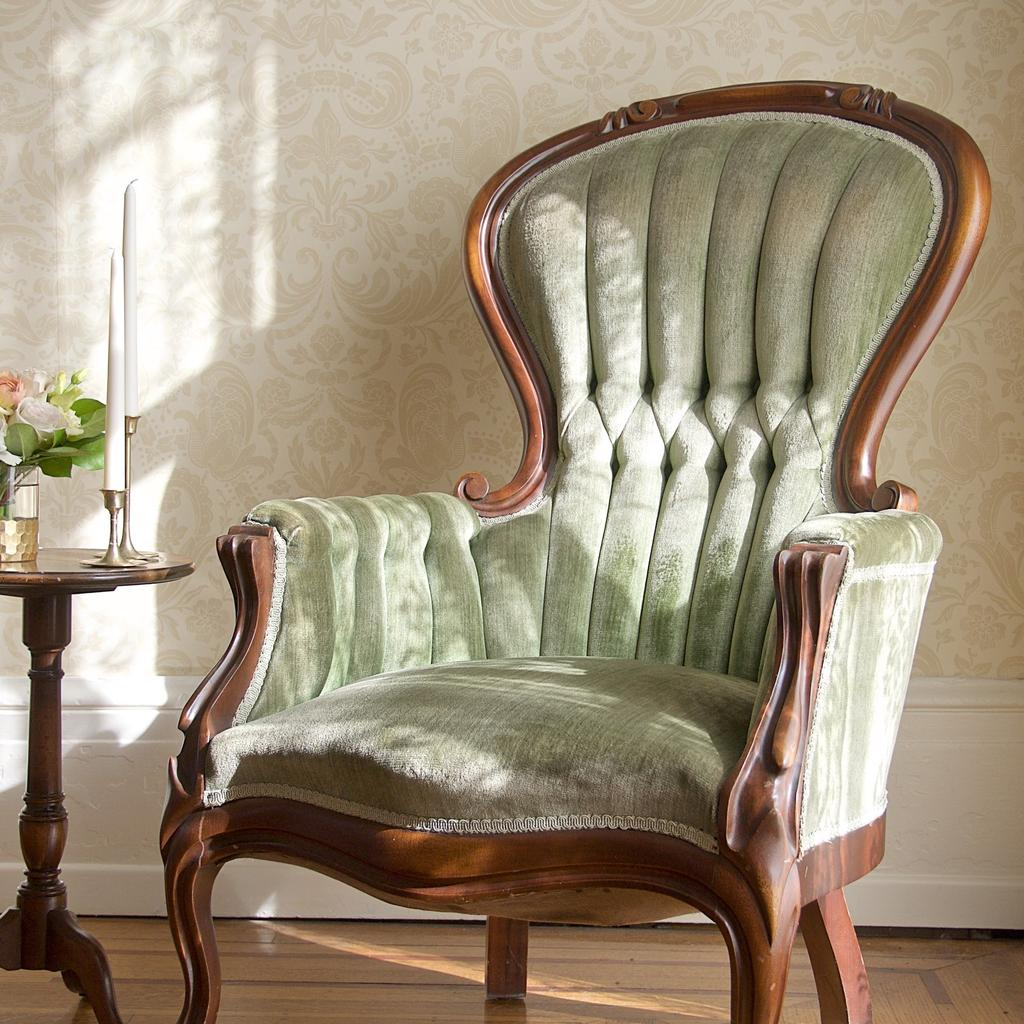} & \qimg{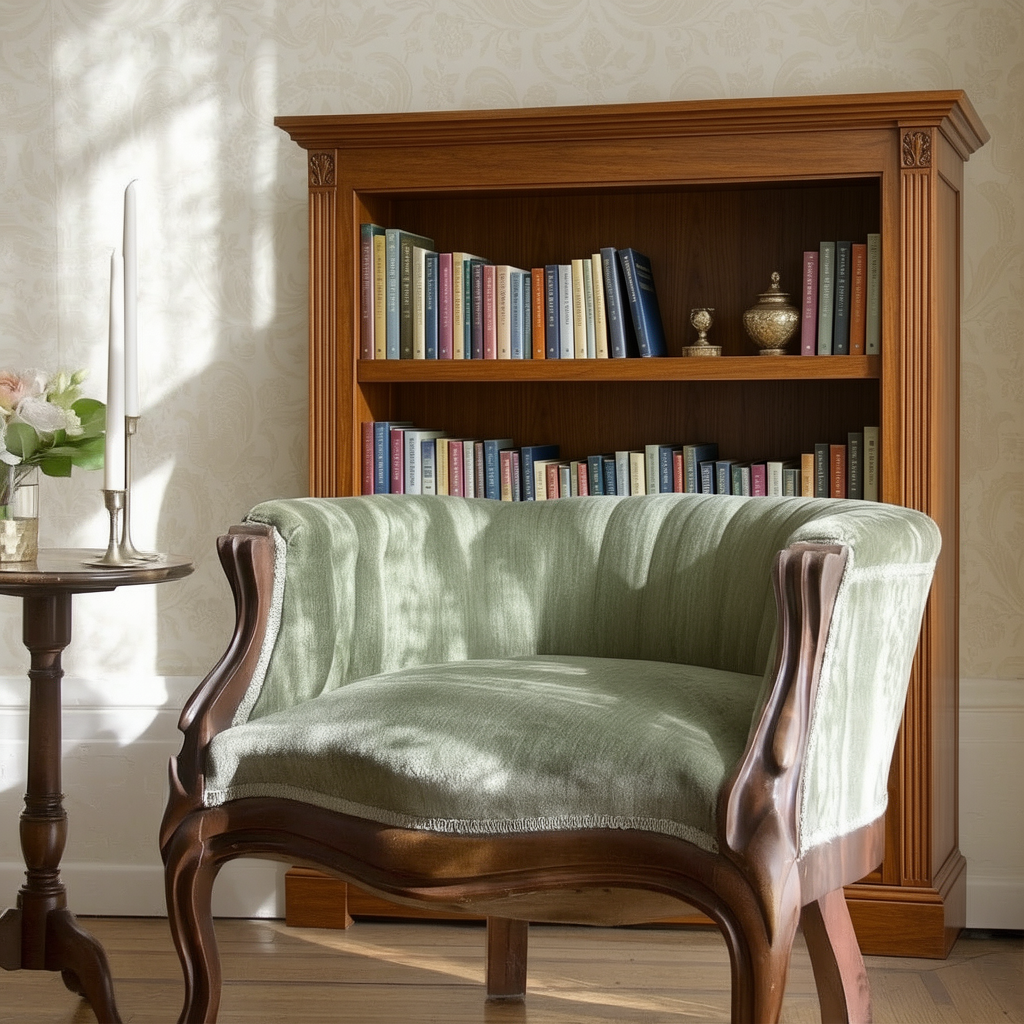} & \qtxtcap{\fontsize{6.5}{7.3}\selectfont A medium shot of a sunlit room corner with an elegant, classic aesthetic, viewed from a slightly right-angled perspective. On the right, a dark wooden bookshelf\,\ldots{}} & \qimg{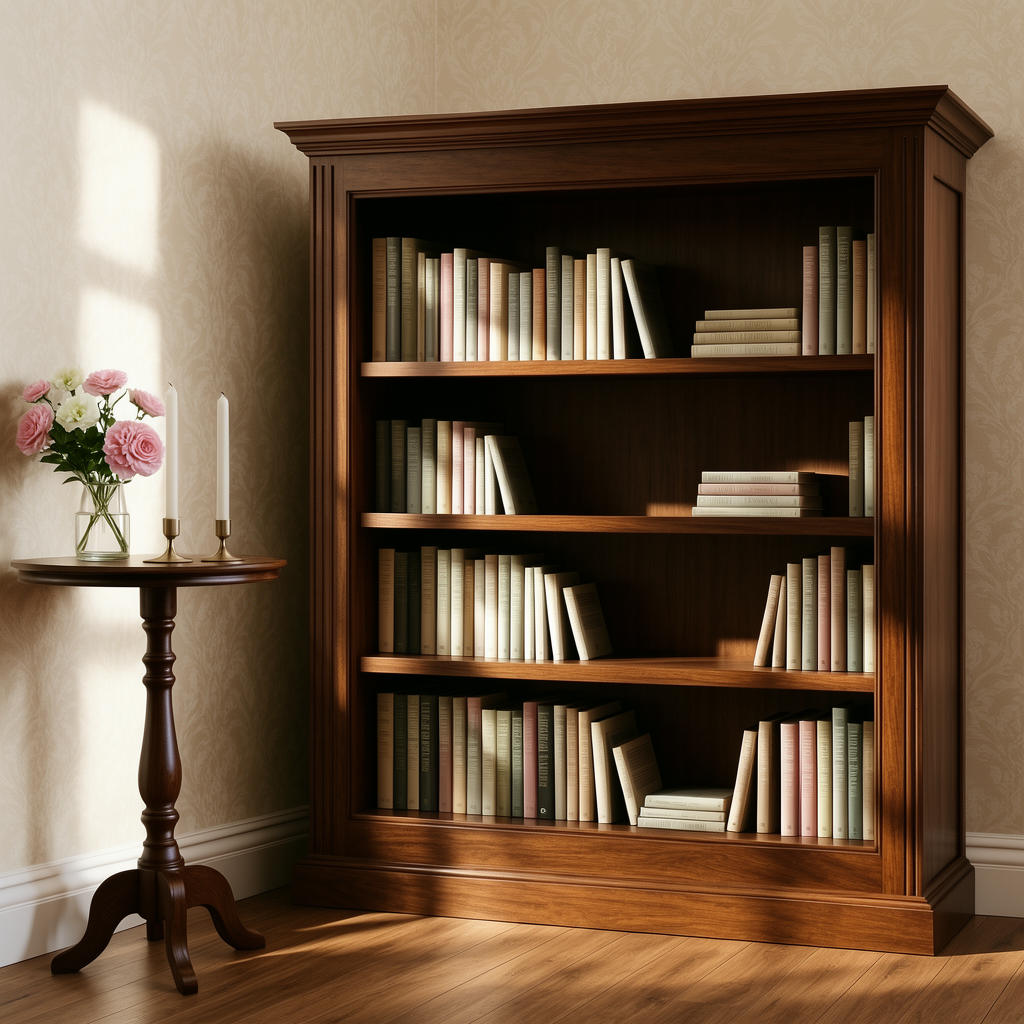} & \qimg{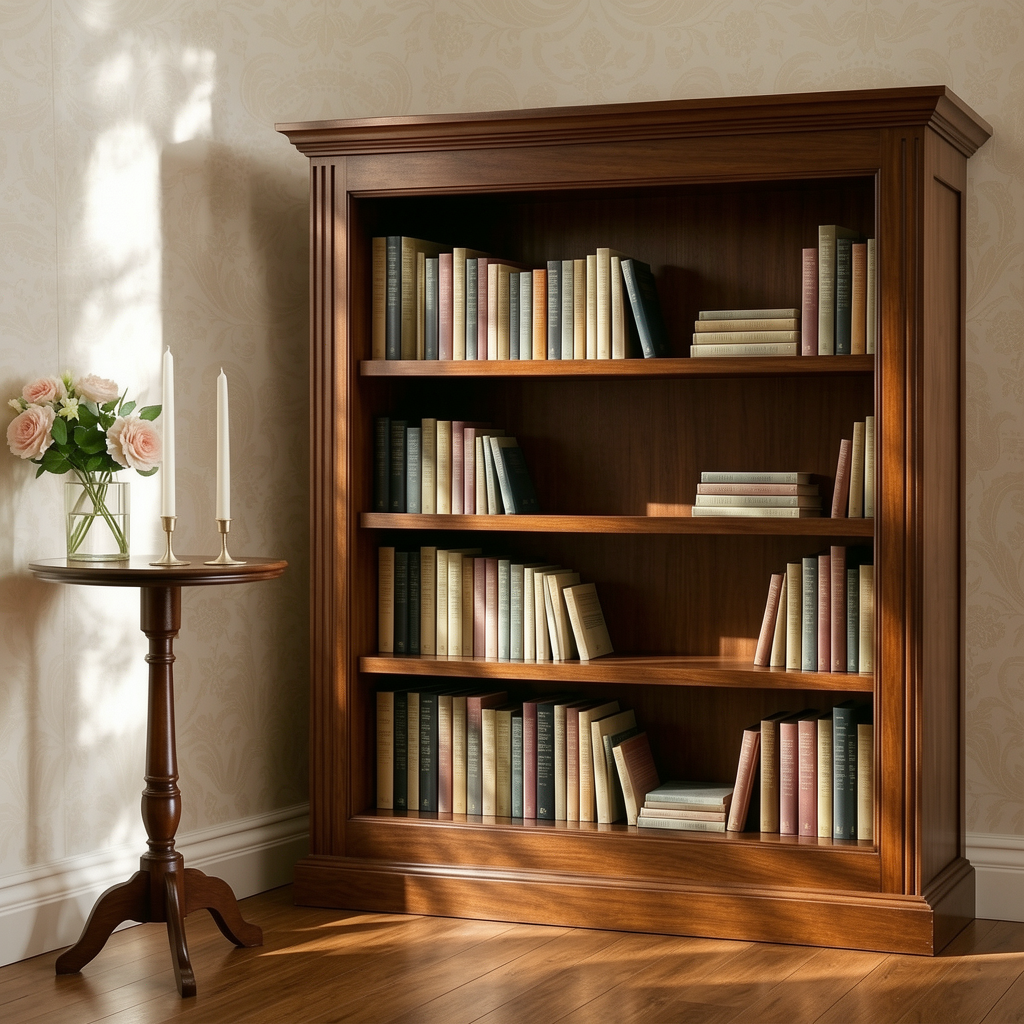} \\
\qtxtinstr{\fontsize{6.8}{7.8}\selectfont Extract the Polaris Ranger XP utility vehicle from the snowy outdoor scene in the image} & \qimg{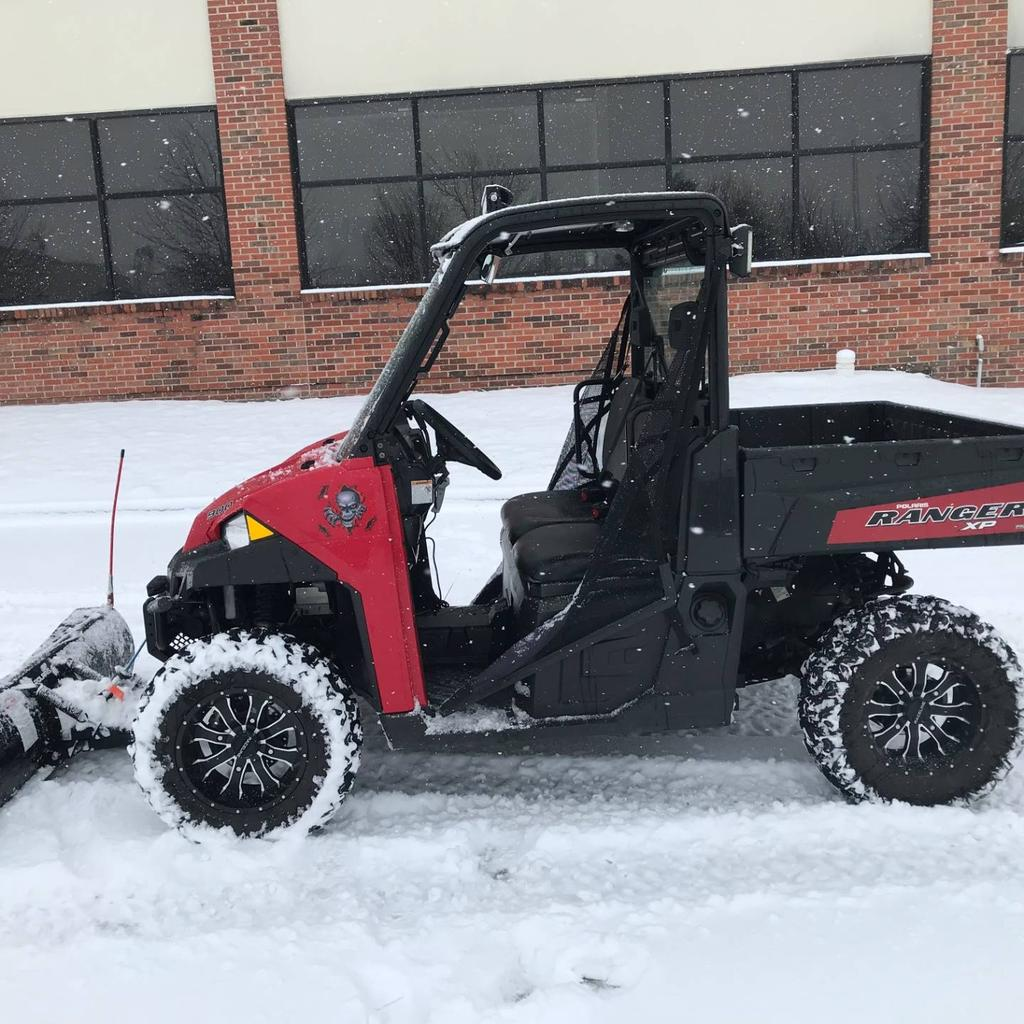} & \qimg{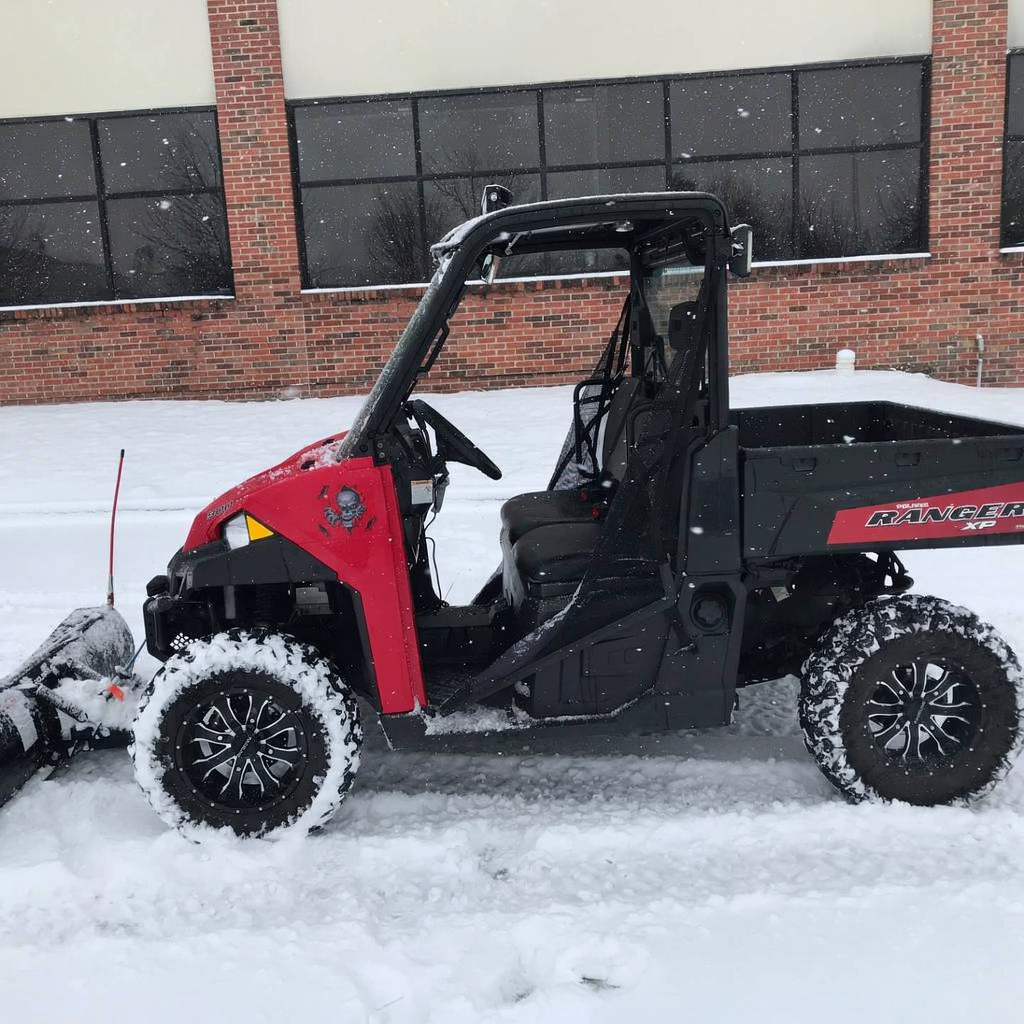} & \qtxtcap{\fontsize{6.5}{7.3}\selectfont A photorealistic shot of a red and black Polaris Ranger XP utility vehicle, isolated against a plain white background. The vehicle is seen from a side profile,\,\ldots{}} & \qimg{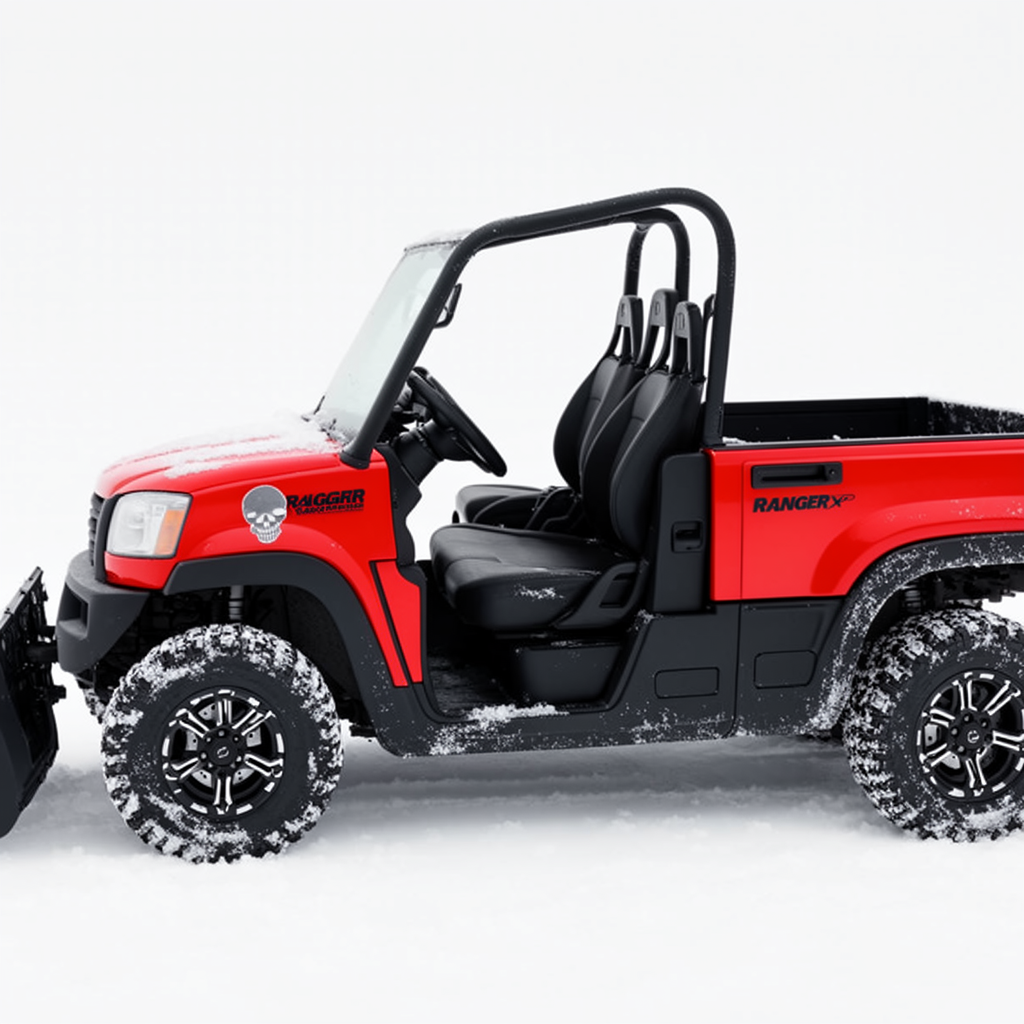} & \qimg{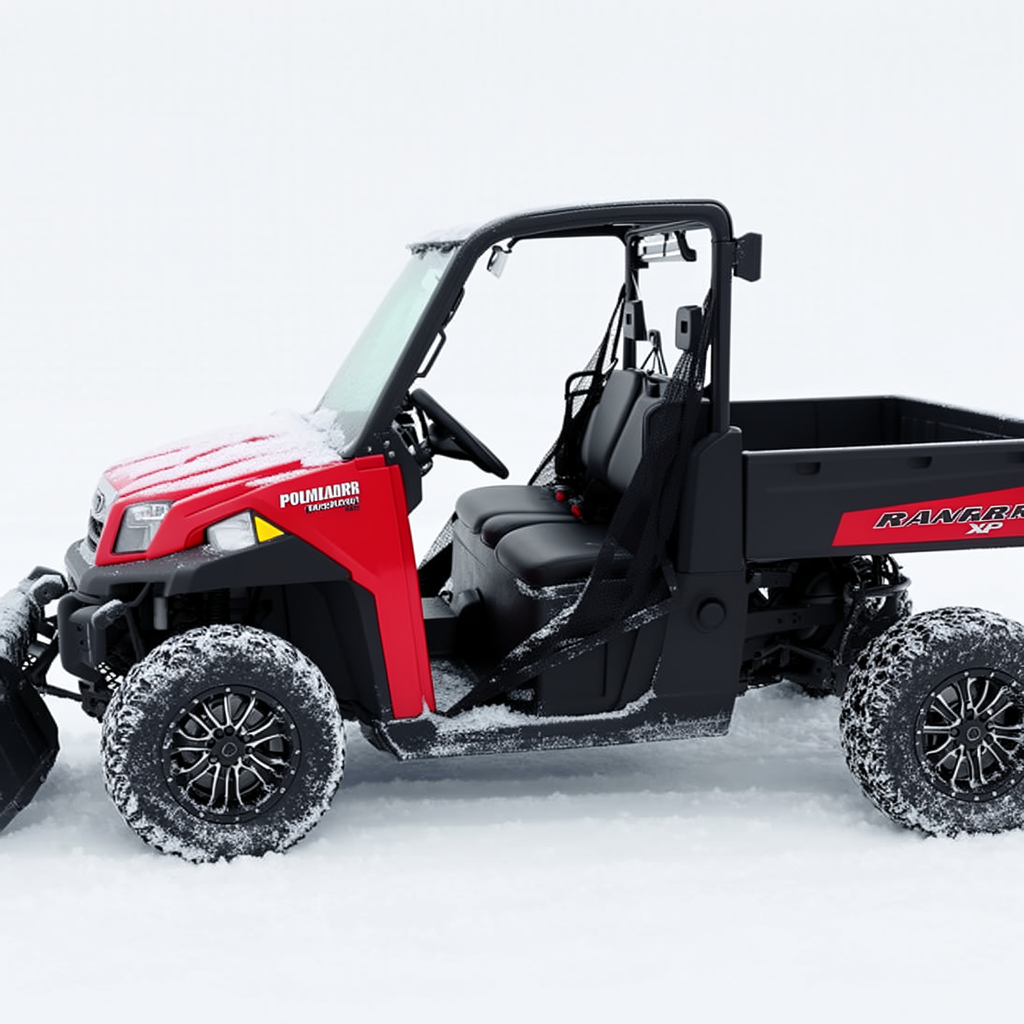} \\
\qtxtinstr{\fontsize{7.5}{8.5}\selectfont Extract the animal in the image} & \qimg{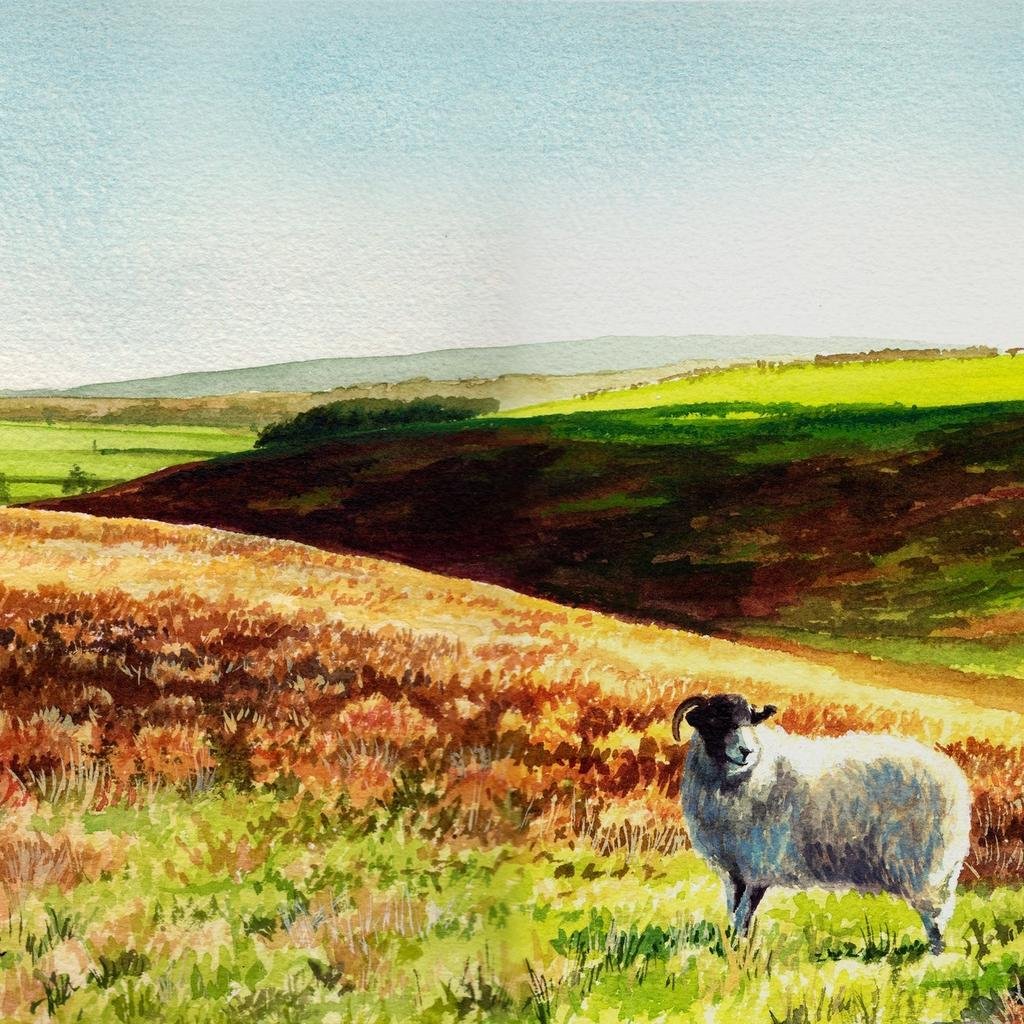} & \qimg{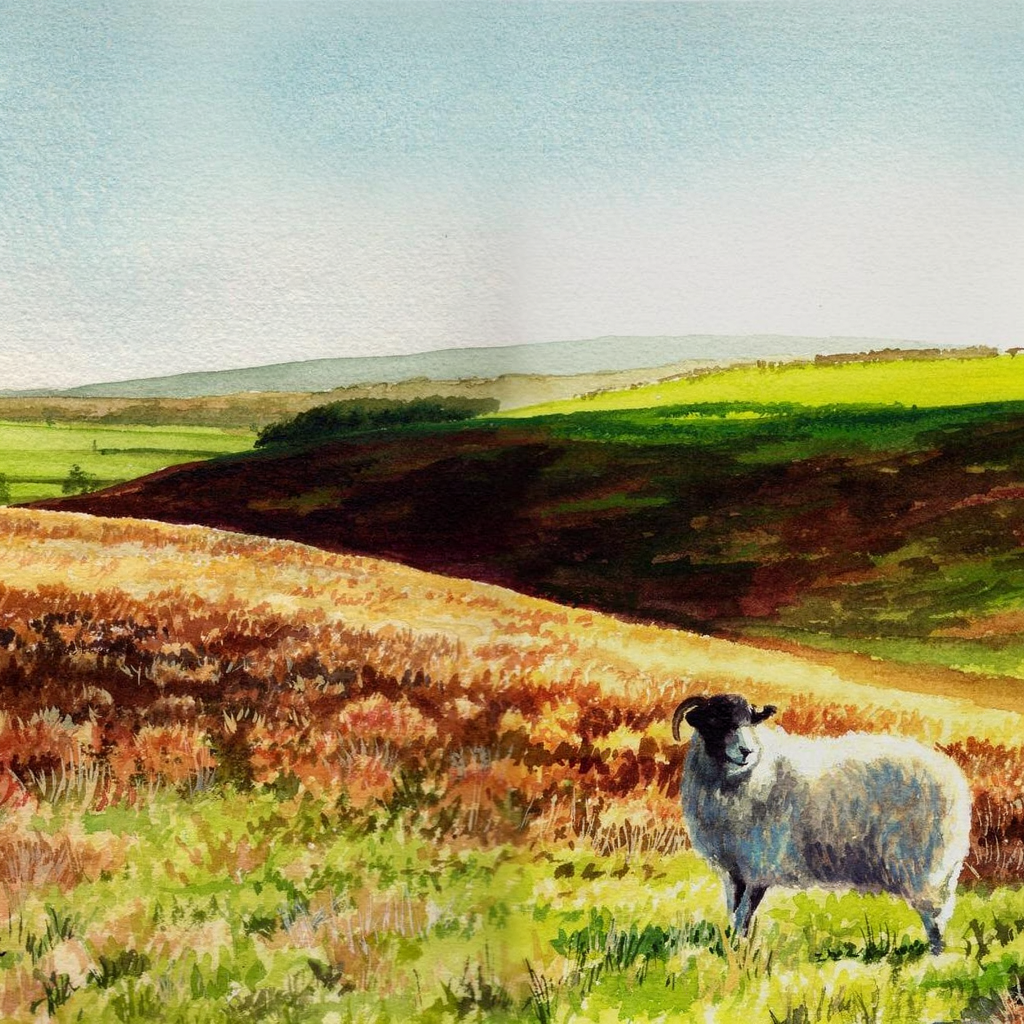} & \qtxtcap{\fontsize{6.5}{7.3}\selectfont A watercolor painting of a single sheep, isolated on a plain white background. The sheep has a black face, small black ears, and two small, dark, curved horns. Its\,\ldots{}} & \qimg{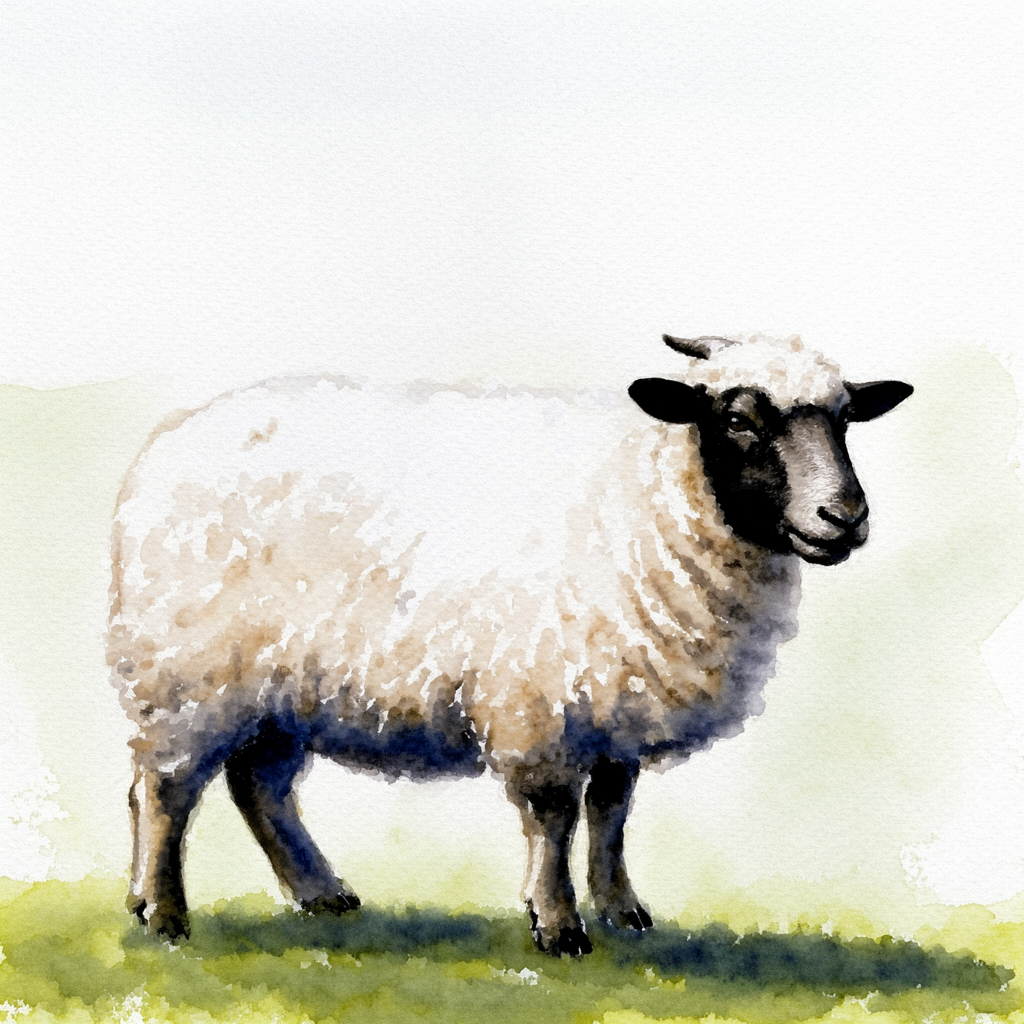} & \qimg{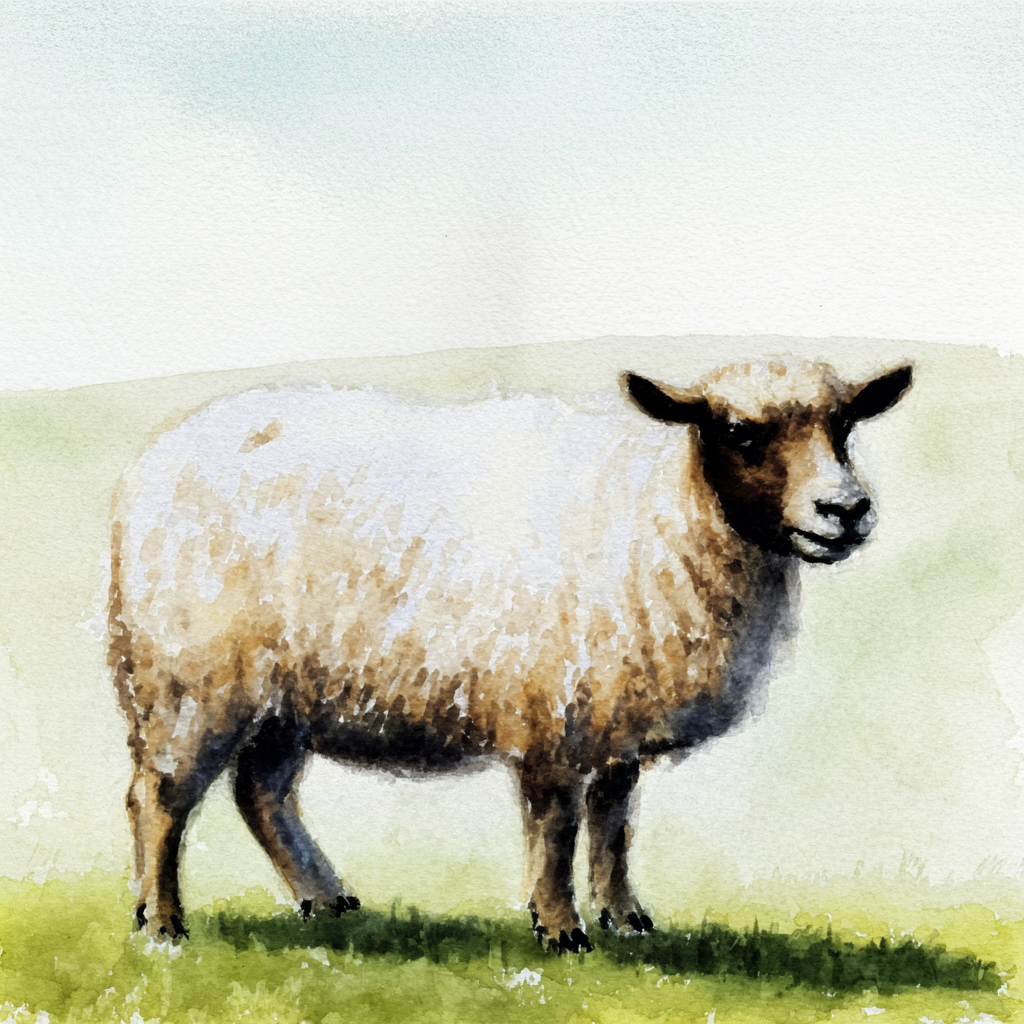} \\
\qtxtinstrtall{\fontsize{6.8}{7.8}\selectfont Change the scene setting to a sunny summer afternoon.} & \qimgtall{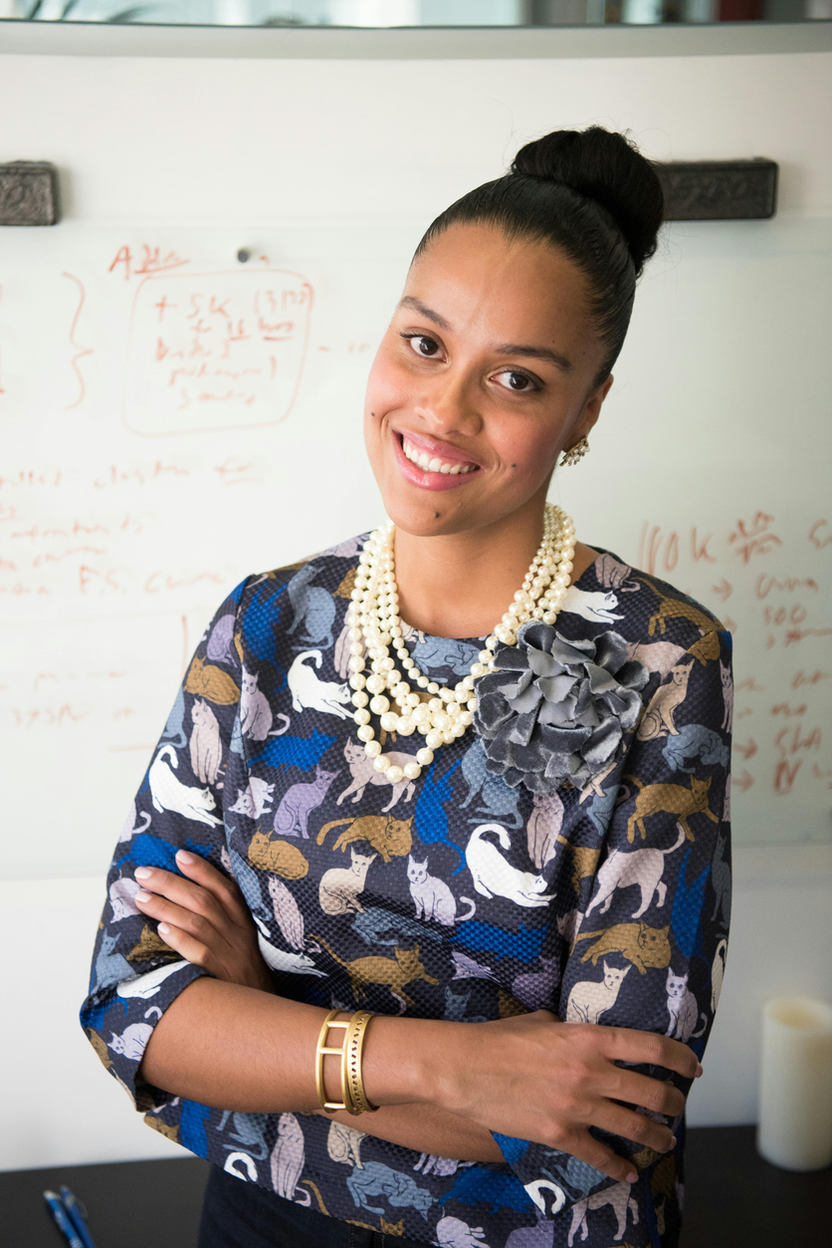} & \qimgtall{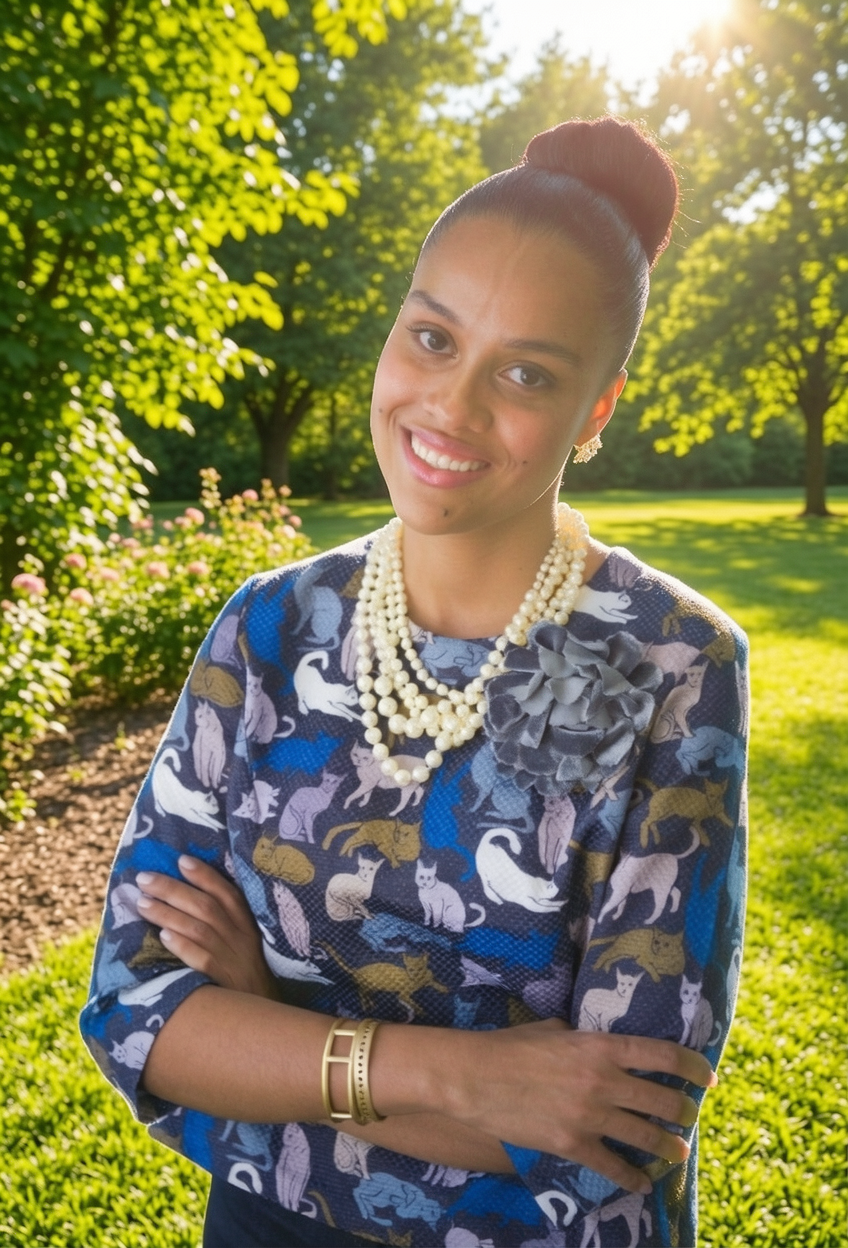} & \qtxtcaptall{\fontsize{6.5}{7.3}\selectfont A photorealistic medium shot of a friendly woman with light brown skin and dark hair pulled up into a neat high bun, standing outdoors on a sunny summer afternoon.\,\ldots{}} & \qimgtall{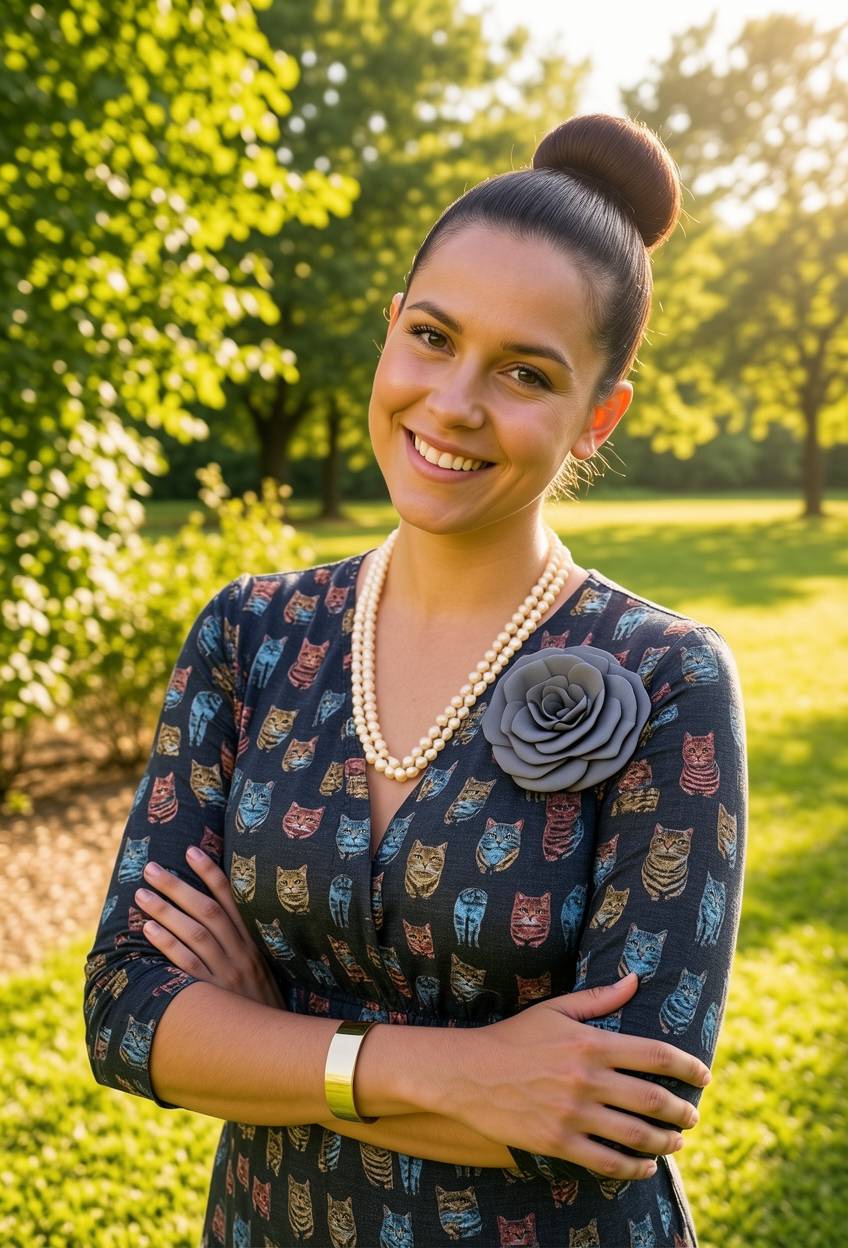} & \qimgtall{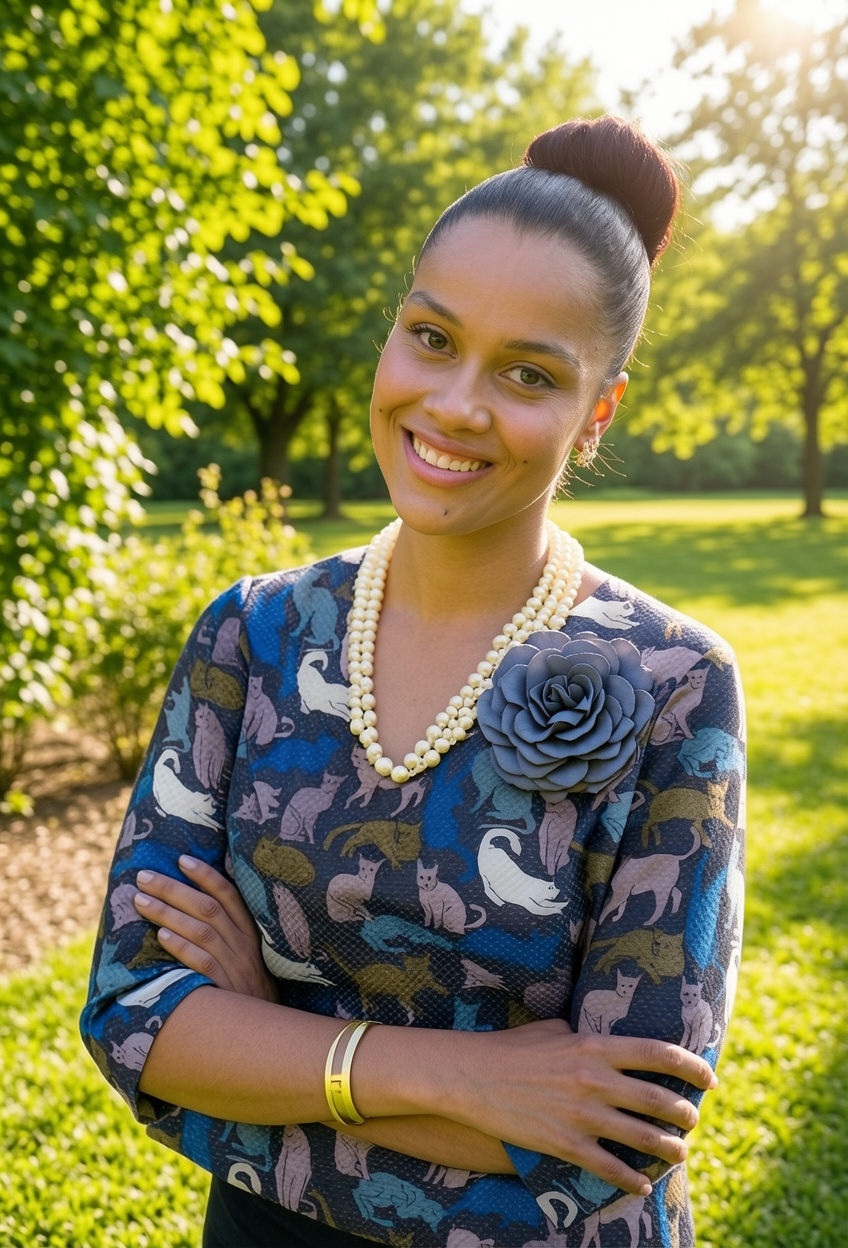} \\
\end{tabular}
\endgroup
\caption{\textbf{Qualitative relevance (FLUX2-Klein 4B).} Each row: source, pure-E, target caption (Gemini), $\mathrm{E}\to\mathrm{T2I}$ at $k{=}10$ without returning to edit mode, and DuET $\mathrm{E}\to\mathrm{T2I}\to\mathrm{E}$ on $[10,20)$. The T2I switch improves semantic relevance on global edits; staying in T2I sacrifices preservation (\cref{fig:giebench}), while DuET resumes edit mode at $k{=}20$.}
\label{fig:qualitative}
\end{figure}

\Cref{fig:qualitative} makes the same trade-off visible on individual edits, comparing the no-switching baseline, a single-$k$ $\mathrm{E}\to\mathrm{T2I}$ switch at $k{=}10$ that stays in T2I to the end, and DuET on $[10,20)$. On global edits---object replacements, extractions, and structural rewrites---the brief T2I interval better matches the intended scene than baseline editing, which sometimes fails to execute the edit. The single-$k$ control also improves relevance, but by staying in T2I it re-renders the whole image and sacrifices preservation relative to DuET, which resumes edit mode at $k{=}20$. For this early interval fine details are reduced (candle length in row 3, text on the vehicle in row 4, the necklace and facial features in the last row) while edit-fidelity gains are more pronounced (\cref{tab:appx-imgedit}); DuET still mitigates facial degradation relative to the single-$k$ variant. The rest of this section asks where this preservation cost is perceptible (\cref{sec:giebench}) and which conditioning changes produce the gains (\cref{sec:ablation,sec:t2i-vs-i2c}).

\subsection{Preservation is perceptual: Selective DuET}
\label{sec:giebench}

GIE-Bench scores functional correctness (whether the edit follows the instruction) and preservation (whether unedited regions remain unchanged) separately, reporting preservation as pixel SSIM. On our fixed intervals SSIM drops whenever the T2I switch is active (\cref{tab:main})---but pixel SSIM overstates this cost: small structural changes that a human still reads as faithful nonetheless lower it. To tell perceptible from imperceptible preservation loss we introduce a \emph{GPT preservation score}: a GEdit-style VLM judge that rates only whether unedited content \emph{looks} unchanged (the preservation half of the GEdit instruction, without the fidelity part), on a $0$--$10$ scale.

\begin{figure}[!htbp]
	\centering
	\begin{minipage}[t]{0.49\linewidth}
		\centering
		\includegraphics[width=\linewidth]{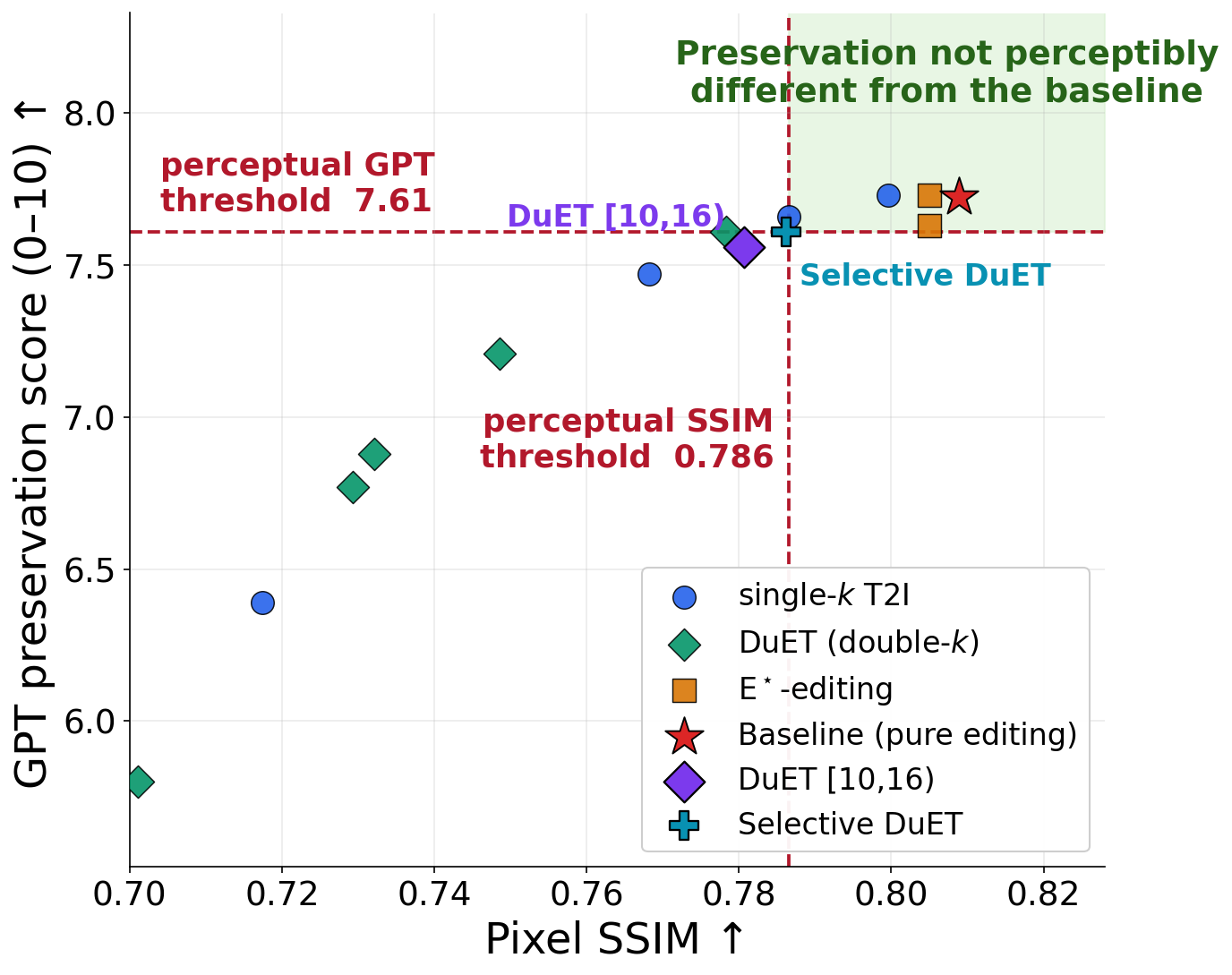}
		\caption{\textbf{Pixel SSIM overstates preservation loss, calibrated against a perceptual judge (FLUX2-Klein 4B configurations).} GIE-Bench's own preservation metric, pixel SSIM ($\uparrow$), vs.\ our GPT preservation score (a GEdit-style, preservation-only VLM judge; $0$--$10$, $\uparrow$) over all 4B configurations. The GPT judge saturates on a plateau: SSIM falls from the strict baseline ($\approx 0.815$) to $\approx 0.786$ with no perceptible preservation change. We read this knee as two thresholds (dashed red): $\mathrm{SSIM}{=}0.786$ and GPT preservation $=7.61$; above both (green) preservation is not perceptibly different from the baseline. \emph{Selective DuET} lands at the knee, while fixed DuET $[10,16)$ sits just left of it.}
		\label{fig:ssim-calibration}
	\end{minipage}\hfill
	\begin{minipage}[t]{0.49\linewidth}
		\centering
		\includegraphics[width=\linewidth]{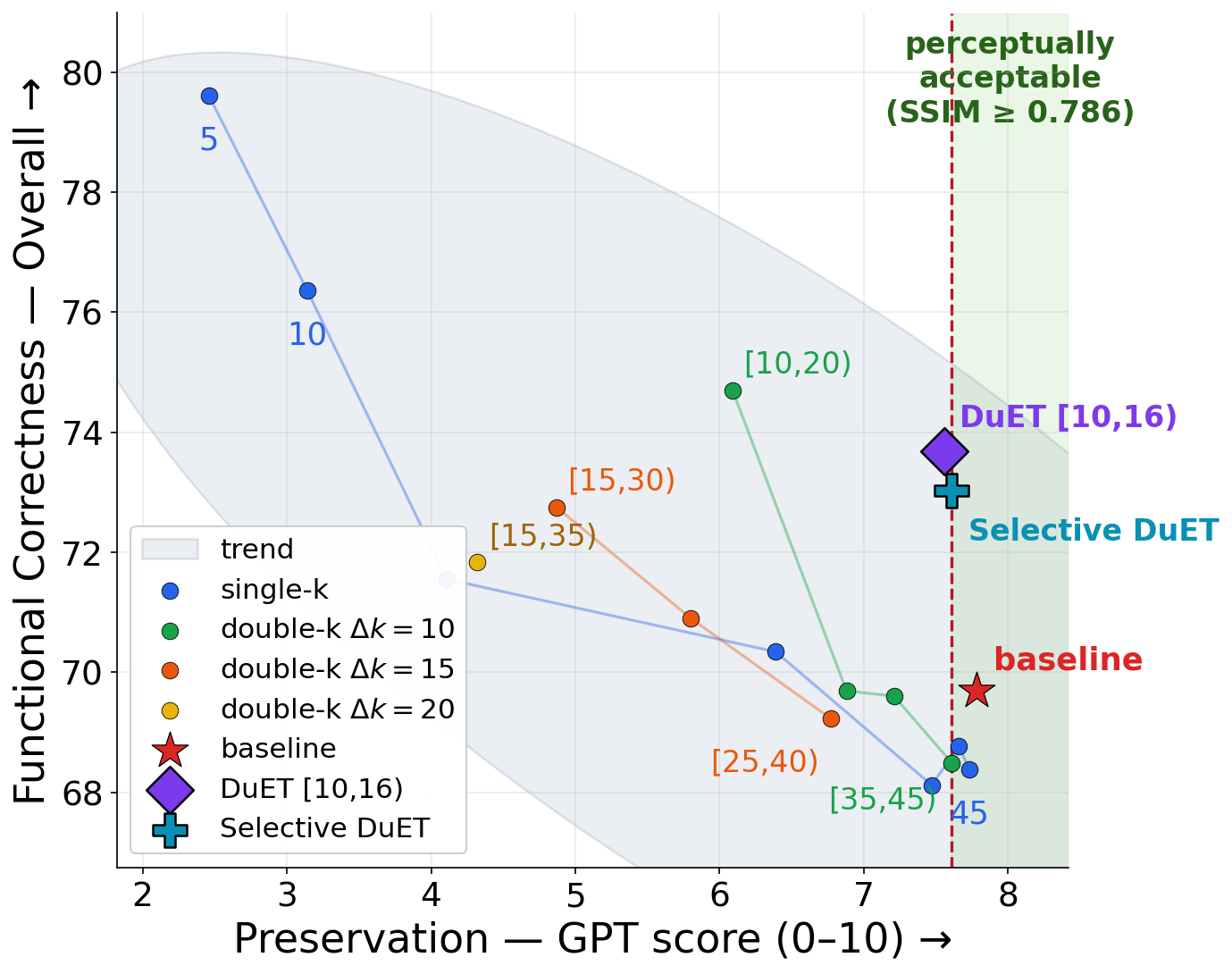}
		\caption{\textbf{Functional correctness vs.\ source preservation on GIE-Bench (FLUX2-Klein 4B configurations).} Every single-$k$ (blue) and double-$k$ (by width $\Delta k$) configuration, as functional correctness (Overall, Gemini-3-Flash judge; $\uparrow$) vs.\ our GPT preservation score ($0$--$10$; $\uparrow$). Correctness and preservation are anti-correlated (covariance ellipse); the baseline ($\star$) sits in the high-preservation corner. The green band is perceptually acceptable (GPT preservation $\ge 7.61$, the $\mathrm{SSIM}{=}0.786$ knee of \cref{fig:ssim-calibration}). Fixed DuET $[10,16)$ buys the most correctness but slips just left of the band; \emph{Selective DuET} keeps nearly the same correctness while staying inside it.}
		\label{fig:giebench}
	\end{minipage}
\end{figure}

\paragraph{A perceptual preservation threshold.}
We calibrate SSIM against this GPT preservation score over 35 FLUX2-Klein~4B configurations---DuET at various intervals, single-$k$ switches, the $\mathrm{E}^{*}$ control, and baseline seeds (\cref{fig:ssim-calibration}). The GPT judge stays on the pure-editing plateau as SSIM falls from the strict baseline ($\approx 0.815$) to a knee at $\approx 0.786$, and only registers a change below it. We take this knee---$\mathrm{SSIM}{=}0.786$, GPT preservation ${\approx}\,7.61$---as a perceptual threshold: configurations above it are not perceptibly worse at preservation than the baseline, even when their SSIM is measurably lower.

\paragraph{The correctness--preservation frontier.}
With preservation on a perceptual scale, we plot every single-$k$ and double-$k$ configuration in preservation--correctness space (\cref{fig:giebench}; functional correctness by Gemini-3-Flash). They fall on a clear frontier: raising correctness lowers preservation. The baseline ($\star$) sits in the high-preservation, low-correctness corner; stronger switches---earlier single-$k$ endpoints or wider double-$k$ windows---push toward higher correctness and lower preservation. Fixed DuET intervals lie on this frontier, and $[10,16)$ buys the most correctness but falls just below the perceptual band. This complements \cref{tab:main}, whose absolute FC values use the benchmark's GPT-4o judge.

\paragraph{Selective DuET.}
The frontier above is traced by \emph{fixed} intervals applied to every edit, but the probe signals of \cref{eq:pi-cv} suggest a per-edit alternative. \emph{Selective DuET} (\cref{sec:method}) reads $(\mathrm{cv},\pi)$ from the pure-editing pass at probe step $k{=}9$ and decides, per edit, whether to switch, following a preservation-first rule:
\begin{equation}
	\text{keep pure editing if } \mathrm{cv} > 0.50 \ \text{ or }\ \pi > 0.22, \qquad \text{else apply DuET } [10,16).
	\label{eq:selective-rule}
\end{equation}
Only edits that are both spatially uniform (low $\mathrm{cv}$) and low in preservation pressure (low $\pi$)---global rewrites that have already released the source---receive the excursion; localized (high $\mathrm{cv}$) or strongly source-anchored (high $\pi$) edits stay in pure editing. The thresholds are not learned but fixed empirically on a subset of the FLUX2-Klein~4B GEdit data (and remain config-tunable). Selective DuET thus raises functional correctness over the baseline while keeping preservation inside the perceptual band (\cref{fig:giebench,fig:ssim-calibration}, \cref{tab:main}), landing at the SSIM knee ($0.786$) rather than trading measurable preservation for correctness like the fixed intervals.

\FloatBarrier
\subsection{Ablation over conditioning variants and interval placement}
\label{sec:ablation}

\begin{figure}[!htbp]
	\centering
	\includegraphics[width=\linewidth]{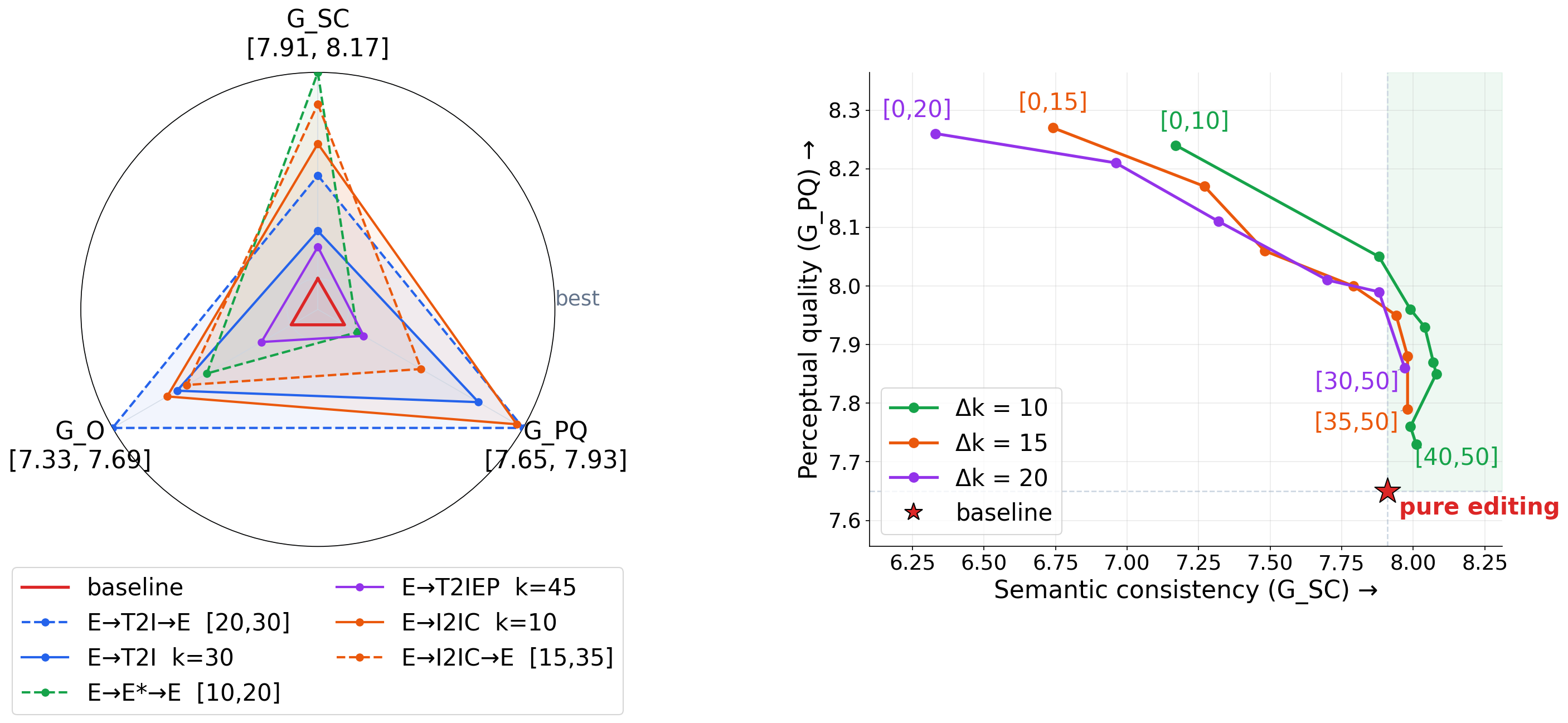}
	\caption{\textbf{GEdit ablations on FLUX2-Klein 4B (Qwen3.5 judge~\citep{qwen2026qwen35}).}
	\emph{Left:} radar over G\_SC, G\_PQ, G\_O (per-axis normalized; baseline as the red triangle near center, per-axis best at the edge; solid${=}$single-$k$, dashed${=}$double-$k$) of the variants whose best configuration beats the baseline on \emph{all three} axes. Six of nine clear this bar; DuET ($\mathrm{E}\to\mathrm{T2I}\to\mathrm{E}$), at its best interval $[20,30)$ here, is among the most balanced.
	\emph{Right:} double-$k$ Pareto curves of DuET in G\_SC vs.\ G\_PQ space, grouped by width $\Delta k$ (including the length-6 sweep); the shaded region beats the baseline ($\star$) on both axes. Sliding the interval later trades quality for consistency; narrowing it shifts the front up and right ($\Delta k{=}6 \succ 10 \succ 15 \succ 20$).}
	\label{fig:ablation}
\end{figure}

We run a factorial ablation on FLUX2-Klein 4B (Qwen3.5 GEdit judge; \cref{fig:ablation}) over nine condition-switching pipelines that vary \emph{(i)}~whether the source image is kept or dropped in the interval, \emph{(ii)}~whether the interval text is the target caption, the edit prompt, or the improved instruction $\mathrm{E}^{*}$, and \emph{(iii)}~switch ordering for single-$k$ endpoints (full grid in \cref{tab:appx-gedit-qwen}). Six of the nine beat the baseline on \emph{all three} GEdit axes at their best interval, with DuET $\mathrm{E}\to\mathrm{T2I}\to\mathrm{E}$ among the most balanced, while a few regress (e.g.\ $\mathrm{E}\to\mathrm{E}^{*}$ at $k{=}0$ collapses perceptual quality). Consistent with \cref{tab:main}, narrower windows dominate wider ones, and later placement favors semantic consistency while earlier placement favors perceptual quality.

\FloatBarrier
\subsection{What drives the gain: dropping image conditioning}
\label{sec:t2i-vs-i2c}

\begin{figure}[!htbp]
	\centering
	\includegraphics[width=\linewidth]{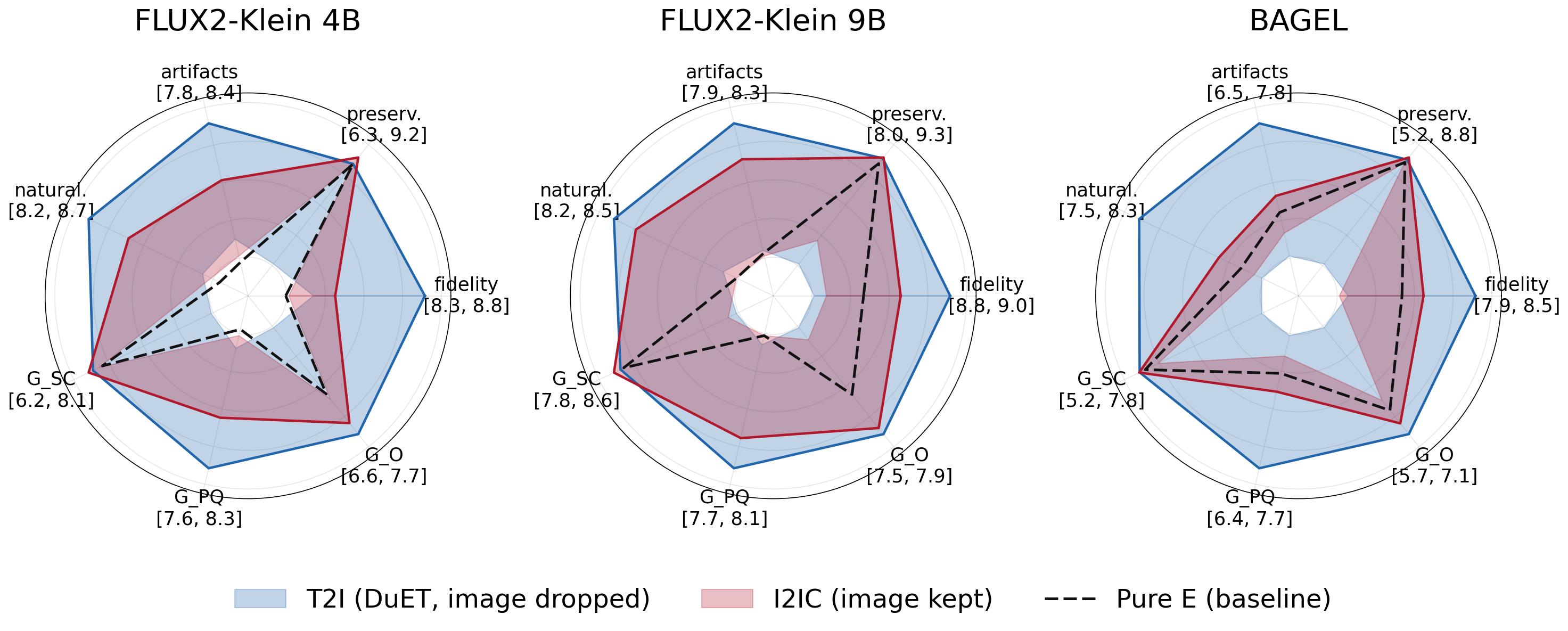}
	\caption{\textbf{Dropping the source image is what drives the gains (decomposed GEdit, Qwen3.5 judge).} One radar per base model (FLUX2-Klein 4B/9B, BAGEL) over the seven-way GEdit decomposition (fidelity, preservation, artifacts, naturalness, and aggregate G\_SC, G\_PQ, G\_O; each axis min--max normalized, raw span beneath its label). Each shaded band is a pipeline's min--max envelope over \emph{all} interval choices: \textbf{T2I} (image dropped, ${=}$ DuET; blue) vs.\ \textbf{I2IC} (image kept with the target caption; red), with Pure-E as the dashed outline. Dropping the image expands the reachable region on fidelity, naturalness, and artifacts (at a preservation cost); keeping it (I2IC) stays near baseline preservation but gains far less on fidelity, naturalness, and artifacts.}
	\label{fig:t2i-vs-i2c}
\end{figure}

To isolate the contribution of dropping the source image, we compare two variants that both use the target caption in the interval but differ only in whether the source image is kept: \textbf{T2I} (dropped, ${=}$ DuET) and \textbf{I2IC} (kept). \Cref{fig:t2i-vs-i2c} decomposes the GEdit metrics per base model, with min--max bands over interval choices and the pure-E baseline overlaid. Dropping the image (T2I) is what boosts fidelity, naturalness, and artifact scores---at a cost in preservation---whereas keeping it (I2IC) stays close to baseline preservation but under-executes global rewrites and gains far less on fidelity, naturalness, and artifacts. Together with the variant ablation above, this shows that \emph{both} dropping the source image \emph{and} switching to a target caption are necessary for DuET-scale gains: T2IEP (image release with the edit prompt) and I2IC (caption switch only) each improve some axes but cannot match the full $\mathrm{E}\to\mathrm{T2I}\to\mathrm{E}$ schedule, and improved instructions alone ($\mathrm{E}^{*}$) likewise fall short.

\subsection{Limitations and future work}
\label{sec:limitations}
Fixed DuET schedules trade measurable source preservation for edit fidelity: any single interval strong enough to raise functional correctness also lowers pixel SSIM relative to the baseline (\cref{fig:giebench}). \emph{Selective DuET} (\cref{sec:giebench}) is a proof of existence that per-edit routing can break this trade-off---improving fidelity, naturalness, and artifact scores while keeping preservation perceptually indistinguishable from the baseline (\cref{fig:ssim-calibration})---rather than a universal solution. Three limitations shape it and point to future work.

First, because we target an \emph{online} router that adds no extra forward passes and never looks ahead, it can only read signals available early in the trajectory---here the $(\mathrm{cv},\pi)$ probe at $k{=}9$---as starting the T2I phase as early as $[10,k_2)$ leaves no room for a later decision. A deeper analysis of which early signals predict a safe excursion could unlock further fidelity and perceptual-quality gains at no preservation cost. Second, the perceptual threshold is calibrated with a VLM judge rather than a formal human study, so the SSIM knee ($0.786$ on 4B) is an operating point, not a universal constant; it and the fixed intervals were tuned on FLUX2-Klein~4B and transferred unchanged to 9B and BAGEL. Third, DuET depends on an external VLM for target captions (and, in the $\mathrm{E}^{*}$ ablation, instruction rewriting): when the captioned target diverges from the instruction's implicit target, the T2I segment can steer toward a coherent but semantically mismatched result, since the caption is generated before editing and does not incorporate the editor's own notion of the final image.

We are actively developing better online task-routing, including learned, pre\-ser\-va\-tion-aware switching schedules; relaxing the strict no-extra-forwards constraint would let this direction be viewed as a form of inference-time scaling.

\section{Conclusion}
\label{sec:conclusion}

Instruction-based editors condition on the source image throughout denoising to preserve structure, yet persistent anchoring can limit instruction fidelity and output naturalness when the target scene diverges from the input. DuET is a training-free task-switching procedure for unified multimodal models (UMMs): it briefly releases source-image anchoring via caption-only text-to-image conditioning mid-trajectory, then resumes edit mode before the final steps---one forward pass, no weight updates.

Several switching regimes improve edit relevance and fidelity relative to standard instruction-based sampling. Fixed schedules do so at a predictable cost in source preservation, but this cost is not intrinsic: \emph{Selective DuET}, a per-edit router over the attention-probe signals $(\mathrm{cv},\pi)$, improves fidelity, naturalness, and artifact scores while holding preservation perceptually constant---a proof of existence that the trade-off can be broken, rather than a universal solution. We believe task switching is a useful and underused capability of UMMs, and that how, when, and how often to switch between conditioning modes warrants further systematic study. 

\FloatBarrier
\bibliographystyle{unsrtnat}
\bibliography{references}

\begin{thebibliography}{24}
\providecommand{\natexlab}[1]{#1}
\providecommand{\url}[1]{\texttt{#1}}
\expandafter\ifx\csname urlstyle\endcsname\relax
  \providecommand{\doi}[1]{doi: #1}\else
  \providecommand{\doi}{doi: \begingroup \urlstyle{rm}\Url}\fi

\bibitem[{Black Forest Labs}(2025)]{flux2025}
{Black Forest Labs}.
\newblock {FLUX.2}: Frontier visual intelligence.
\newblock \url{https://bfl.ai/blog/flux-2}, 2025.

\bibitem[Brooks et~al.(2023)Brooks, Holynski, and
  Efros]{brooks2023instructpix2pix}
Tim Brooks, Aleksander Holynski, and Alexei~A. Efros.
\newblock {InstructPix2Pix}: Learning to follow image editing instructions.
\newblock In \emph{Proceedings of the IEEE/CVF Conference on Computer Vision
  and Pattern Recognition (CVPR)}, 2023.

\bibitem[Rombach et~al.(2022)Rombach, Blattmann, Lorenz, Esser, and
  Ommer]{rombach2022ldm}
Robin Rombach, Andreas Blattmann, Dominik Lorenz, Patrick Esser, and Bj{\"o}rn
  Ommer.
\newblock High-resolution image synthesis with latent diffusion models.
\newblock In \emph{Proceedings of the IEEE/CVF Conference on Computer Vision
  and Pattern Recognition (CVPR)}, 2022.

\bibitem[Deng et~al.(2025)Deng, Zhu, Li, Gou, Li, Wang, Zhong, Yu, Nie, Song,
  Shi, and Fan]{deng2025bagel}
Chaorui Deng, Deyao Zhu, Kunchang Li, Chenhui Gou, Feng Li, Zeyu Wang, Shu
  Zhong, Weihao Yu, Xiaonan Nie, Ziang Song, Guang Shi, and Haoqi Fan.
\newblock Emerging properties in unified multimodal pretraining.
\newblock \emph{arXiv preprint arXiv:2505.14683}, 2025.

\bibitem[Balaji et~al.(2022)Balaji, Nah, Huang, Vahdat, Song, Zhang, Kreis,
  Aittala, Aila, Laine, Catanzaro, Karras, and Liu]{balaji2022ediffi}
Yogesh Balaji, Seungjun Nah, Xun Huang, Arash Vahdat, Jiaming Song, Qinsheng
  Zhang, Karsten Kreis, Miika Aittala, Timo Aila, Samuli Laine, Bryan
  Catanzaro, Tero Karras, and Ming-Yu Liu.
\newblock {eDiff-I}: Text-to-image diffusion models with an ensemble of expert
  denoisers.
\newblock \emph{arXiv preprint arXiv:2211.01324}, 2022.

\bibitem[Feng et~al.(2023)Feng, Zhang, Yu, Fang, Li, Chen, Lu, Liu, Yin, Feng,
  Sun, Chen, Tian, Wu, and Wang]{feng2023ernievilg}
Zhida Feng, Zhenyu Zhang, Xintong Yu, Yewei Fang, Lanxin Li, Xuyi Chen, Yuxiang
  Lu, Jiaxiang Liu, Weichong Yin, Shikun Feng, Yu~Sun, Li~Chen, Hao Tian, Hua
  Wu, and Haifeng Wang.
\newblock {ERNIE-ViLG 2.0}: Improving text-to-image diffusion model with
  knowledge-enhanced mixture-of-denoising-experts.
\newblock In \emph{Proceedings of the IEEE/CVF Conference on Computer Vision
  and Pattern Recognition (CVPR)}, 2023.

\bibitem[Park et~al.(2024)Park, Go, Kim, Woo, Ham, and Kim]{park2024switchdit}
Byeongjun Park, Hyojun Go, Jin-Young Kim, Sangmin Woo, Seokil Ham, and Changick
  Kim.
\newblock Switch diffusion transformer: Synergizing denoising tasks with sparse
  mixture-of-experts.
\newblock In \emph{European Conference on Computer Vision (ECCV)}, 2024.

\bibitem[Ma et~al.(2025{\natexlab{a}})Ma, Ning, Liu, Niu, and
  Zhang]{ma2024deme}
Qianli Ma, Xuefei Ning, Dongrui Liu, Li~Niu, and Linfeng Zhang.
\newblock Decouple-then-merge: Finetune diffusion models as multi-task
  learning.
\newblock In \emph{Proceedings of the IEEE/CVF Conference on Computer Vision
  and Pattern Recognition (CVPR)}, 2025{\natexlab{a}}.

\bibitem[Zhuang et~al.(2025)Zhuang, Guo, Ding, Li, Chen, Wang, Wang, Zhang, Li,
  and Wang]{timestepmaster2025}
Shaobin Zhuang, Yiwei Guo, Yanbo Ding, Kunchang Li, Xinyuan Chen, Yaohui Wang,
  Fangyikang Wang, Ying Zhang, Chen Li, and Yali Wang.
\newblock {TimeStep Master}: Asymmetrical mixture of timestep {LoRA} experts
  for versatile and efficient diffusion models in vision.
\newblock In \emph{International Conference on Machine Learning (ICML)}, 2025.

\bibitem[Wan et~al.(2025)Wan, Wang, Ai, Wen, Mao, Xie, Chen, Yu, Zhao, Yang,
  Zeng, Wang, Zhang, Zhou, Wang, Chen, Zhu, Zhao, Yan, Huang, Feng, Zhang, Li,
  Wu, Chu, Feng, Zhang, Sun, Fang, Wang, Gui, Weng, Shen, Lin, Wang, Wang,
  Zhou, Wang, Shen, Yu, Shi, Huang, Xu, Kou, Lv, Li, Liu, Wang, Zhang, Huang,
  Li, Wu, Liu, Pan, Zheng, Hong, Shi, Feng, Jiang, Han, Wu, and Liu]{wan2025}
Team Wan, Ang Wang, Baole Ai, Bin Wen, Chaojie Mao, Chen-Wei Xie, Di~Chen,
  Feiwu Yu, Haiming Zhao, Jianxiao Yang, Jianyuan Zeng, Jiayu Wang, Jingfeng
  Zhang, Jingren Zhou, Jinkai Wang, Jixuan Chen, Kai Zhu, Kang Zhao, Keyu Yan,
  Lianghua Huang, Mengyang Feng, Ningyi Zhang, Pandeng Li, Pingyu Wu, Ruihang
  Chu, Ruili Feng, Shiwei Zhang, Siyang Sun, Tao Fang, Tianxing Wang, Tianyi
  Gui, Tingyu Weng, Tong Shen, Wei Lin, Wei Wang, Wei Wang, Wenmeng Zhou, Wente
  Wang, Wenting Shen, Wenyuan Yu, Xianzhong Shi, Xiaoming Huang, Xin Xu, Yan
  Kou, Yangyu Lv, Yifei Li, Yijing Liu, Yiming Wang, Yingya Zhang, Yitong
  Huang, Yong Li, You Wu, Yu~Liu, Yulin Pan, Yun Zheng, Yuntao Hong, Yupeng
  Shi, Yutong Feng, Zeyinzi Jiang, Zhen Han, Zhi-Fan Wu, and Ziyu Liu.
\newblock {Wan}: Open and advanced large-scale video generative models.
\newblock \emph{arXiv preprint arXiv:2503.20314}, 2025.

\bibitem[Ma{\~n}as et~al.(2024)Ma{\~n}as, Astolfi, Hall, Ross, Urbanek,
  Williams, Agrawal, Romero-Soriano, and Drozdzal]{manas2024opt2i}
Oscar Ma{\~n}as, Pietro Astolfi, Melissa Hall, Candace Ross, Jack Urbanek,
  Adina Williams, Aishwarya Agrawal, Adriana Romero-Soriano, and Michal
  Drozdzal.
\newblock Improving text-to-image consistency via automatic prompt
  optimization.
\newblock \emph{arXiv preprint arXiv:2403.17804}, 2024.

\bibitem[Khan et~al.(2025)Khan, Jain, Bhattacharyya, and Vineet]{khan2025tir}
Mohammad Abdul~Hafeez Khan, Yash Jain, Siddhartha Bhattacharyya, and Vibhav
  Vineet.
\newblock Test-time prompt refinement for text-to-image models.
\newblock \emph{arXiv preprint arXiv:2507.22076}, 2025.

\bibitem[Li et~al.(2025)Li, Kallidromitis, Gokul, Koneru, Kato, Kozuka, and
  Grover]{li2025reflectdit}
Shufan Li, Konstantinos Kallidromitis, Akash Gokul, Arsh Koneru, Yusuke Kato,
  Kazuki Kozuka, and Aditya Grover.
\newblock {Reflect-DiT}: Inference-time scaling for text-to-image diffusion
  transformers via in-context reflection.
\newblock \emph{arXiv preprint arXiv:2503.12271}, 2025.

\bibitem[Xie et~al.(2025)Xie, Chen, Zhao, Yu, Zhu, Wu, Lin, Zhang, Li, Chen,
  Cai, Liu, Zhou, and Han]{xie2025sana15}
Enze Xie, Junsong Chen, Yuyang Zhao, Jincheng Yu, Ligeng Zhu, Chengyue Wu,
  Yujun Lin, Zhekai Zhang, Muyang Li, Junyu Chen, Han Cai, Bingchen Liu, Daquan
  Zhou, and Song Han.
\newblock {SANA} 1.5: Efficient scaling of training-time and inference-time
  compute in linear diffusion transformer.
\newblock \emph{arXiv preprint arXiv:2501.18427}, 2025.

\bibitem[Ma et~al.(2025{\natexlab{b}})Ma, Tong, Jia, Hu, Su, Zhang, Yang, Li,
  Jaakkola, Jia, and Xie]{ma2025inferencescaling}
Nanye Ma, Shangyuan Tong, Haolin Jia, Hexiang Hu, Yu-Chuan Su, Mingda Zhang,
  Xuan Yang, Yandong Li, Tommi Jaakkola, Xuhui Jia, and Saining Xie.
\newblock Inference-time scaling for diffusion models beyond scaling denoising
  steps.
\newblock \emph{arXiv preprint arXiv:2501.09732}, 2025{\natexlab{b}}.

\bibitem[Comanici et~al.(2025)Comanici, Bieber, Schaekermann,
  et~al.]{comanici2025gemini25}
Gheorghe Comanici, Eric Bieber, Mike Schaekermann, et~al.
\newblock {Gemini} 2.5: Pushing the frontier with advanced reasoning,
  multimodality, long context, and next generation agentic capabilities.
\newblock \emph{arXiv preprint arXiv:2507.06261}, 2025.

\bibitem[Liu et~al.(2025)Liu, Han, Xing, Yin, Wang, Cheng, Liao, Wang, Fu, Han,
  Li, Peng, Sun, Wu, Cai, Ge, Ming, Xia, Zeng, Zhu, Jiao, Zhang, Yu, and
  Jiang]{liu2025gedit}
Shiyu Liu, Yucheng Han, Peng Xing, Fukun Yin, Rui Wang, Wei Cheng, Jiaqi Liao,
  Yingming Wang, Honghao Fu, Chunrui Han, Guopeng Li, Yuang Peng, Quan Sun,
  Jingwei Wu, Yan Cai, Zheng Ge, Ranchen Ming, Lei Xia, Xianfang Zeng, Yibo
  Zhu, Binxing Jiao, Xiangyu Zhang, Gang Yu, and Daxin Jiang.
\newblock {Step1X-Edit}: A practical framework for general image editing.
\newblock \emph{arXiv preprint arXiv:2504.17761}, 2025.

\bibitem[{OpenAI}(2025)]{openai2025gpt41}
{OpenAI}.
\newblock Introducing {GPT-4.1} in the {API}.
\newblock \url{https://openai.com/index/gpt-4-1/}, 2025.

\bibitem[Ye et~al.(2025)Ye, He, Li, Lin, Yuan, Yan, Hou, and
  Yuan]{ye2025imgedit}
Yang Ye, Xianyi He, Zongjian Li, Bin Lin, Shenghai Yuan, Zhiyuan Yan, Bohan
  Hou, and Li~Yuan.
\newblock {ImgEdit}: A unified image editing dataset and benchmark.
\newblock \emph{arXiv preprint arXiv:2505.20275}, 2025.

\bibitem[{OpenAI}(2024)]{openai2024gpt4o}
{OpenAI}.
\newblock {GPT-4o} system card.
\newblock \emph{arXiv preprint arXiv:2410.21276}, 2024.

\bibitem[Zhang et~al.(2023)Zhang, Mo, Chen, Sun, and Su]{zhang2023magicbrush}
Kai Zhang, Lingbo Mo, Wenhu Chen, Huan Sun, and Yu~Su.
\newblock {MagicBrush}: A manually annotated dataset for instruction-guided
  image editing.
\newblock In \emph{Advances in Neural Information Processing Systems}, 2023.

\bibitem[Qian et~al.(2025)Qian, Lu, Fu, Wang, Chen, Yang, Hu, and
  Gan]{qian2025giebench}
Yusu Qian, Jiasen Lu, Tsu-Jui Fu, Xinze Wang, Chen Chen, Yinfei Yang, Wenze Hu,
  and Zhe Gan.
\newblock {GIE-Bench}: Towards grounded evaluation for text-guided image
  editing.
\newblock \emph{arXiv preprint arXiv:2505.11493}, 2025.

\bibitem[{Google DeepMind}(2025)]{google2025gemini3flash}
{Google DeepMind}.
\newblock Introducing {Gemini} 3 {Flash}: Benchmarks, global availability.
\newblock
  \url{https://blog.google/products-and-platforms/products/gemini/gemini-3-flash/},
  2025.

\bibitem[{Qwen Team}(2026)]{qwen2026qwen35}
{Qwen Team}.
\newblock {Qwen3.5}: Towards native multimodal agents.
\newblock \url{https://qwen.ai/blog?id=qwen3.5}, 2026.

\end{thebibliography}

\clearpage
\appendix
\section{Extended numerical results}
\label{app:results}

This appendix collects the full FLUX2-Klein~4B measurement grids behind the main-text figures, plus cross-model ImgEdit breakdowns omitted from \cref{tab:main} for space. Beyond the endpoints in the main text, the GIE-Bench, MagicBrush, GEdit (GPT-4.1), and ImgEdit grids also report the tightened $[10,16)$ interval and the adaptive \emph{Selective DuET} schedule. Table~\ref{tab:appx-giebench-fc} gives GIE-Bench per-category functional correctness (GPT-4o judge) and preservation for the baseline, representative single-$k$ $\mathrm{E}\to\mathrm{T2I}$ endpoints, DuET double-$k$ intervals, the $\mathrm{E}\to\mathrm{E}^{*}\to\mathrm{E}$ control, and Selective DuET. Table~\ref{tab:appx-magicbrush} lists the complete MagicBrush CLIP-T-gen sweep. Table~\ref{tab:appx-gedit-gpt} extends the GEdit grid (GPT-4.1 judge) with all double-$k$ intervals used in \cref{fig:existence}. Table~\ref{tab:appx-gedit-qwen} reports the factorial ablation over conditioning variants (Qwen3.5 judge) underlying \cref{fig:ablation}; the paragraph following it compares best-per-column GEdit scores across the partial switches. Table~\ref{tab:appx-imgedit} breaks ImgEdit down by edit category for FLUX2-Klein 4B/9B and BAGEL, including extra 9B/BAGEL intervals beyond \cref{tab:main}.


\begin{table}[!ht]
\centering
\caption{\textbf{GIE-Bench (FLUX2-Klein 4B, GPT-4o judge).} Functional-correctness per-category scores (left) and preservation metrics (right) for the baseline, selected single-$k$ E$\to$T2I endpoints, DuET double-$k$ intervals, the $\mathrm{E}\to\mathrm{E}^{*}\to\mathrm{E}$ control, and Selective DuET. A row's preservation cells are shaded red when its SSIM drops more than $0.029$ (the perceptual margin) below the baseline SSIM.}
\label{tab:appx-giebench-fc}
\vspace{0.75em}
{\footnotesize
\setlength{\tabcolsep}{3pt}
\resizebox{\linewidth}{!}{
\begin{tabular}{ll|cccccccccc|ccccc}
\toprule
 & & \multicolumn{10}{c|}{\textbf{Functional correctness}} & \multicolumn{5}{c}{\textbf{Preservation}} \\
Pipeline & Interval & \shortstack{Add\\Object} & \shortstack{Attribute\\Change} & \shortstack{Color\\Change} & \shortstack{Layout\\Modif.} & \shortstack{Object\\Replace} & \shortstack{Remove\\Object} & \shortstack{Scene /\\Bkgd\\Change} & \shortstack{Size\\Change} & \shortstack{Textual\\Edit} & Overall & SSIM$\uparrow$ & CLIP$\uparrow$ & \shortstack{Unmasked\\CLIP$\uparrow$} & PSNR$\uparrow$ & MSE$\downarrow$ \\
\midrule
Baseline & --- & 91.67 & 85.00 & 91.60 & 48.33 & 93.33 & 67.50 & 95.00 & 23.33 & 70.83 & 74.05 & 0.82 & 0.97 & 0.92 & 20.99 & 1892.61 \\
E$\to$T2I & $k=5$ & \cellcolor[RGB]{198,239,206}95.00 & \cellcolor[RGB]{205,241,212}95.83 & \cellcolor[RGB]{240,250,242}92.44 & \cellcolor[RGB]{198,239,206}59.17 & \cellcolor[RGB]{239,250,241}94.17 & \cellcolor[RGB]{198,239,206}82.50 & \cellcolor[RGB]{216,244,222}96.67 & \cellcolor[RGB]{198,239,206}38.33 & \cellcolor[RGB]{198,239,206}82.50 & \cellcolor[RGB]{198,239,206}81.84 & \cellcolor[RGB]{255,199,206}0.54 & \cellcolor[RGB]{255,199,206}0.93 & \cellcolor[RGB]{255,199,206}0.86 & \cellcolor[RGB]{255,199,206}9.21 & \cellcolor[RGB]{255,199,206}8892.47 \\
E$\to$T2I & $k=15$ & \cellcolor[RGB]{226,247,230}93.33 & \cellcolor[RGB]{228,247,232}90.83 & \cellcolor[RGB]{212,242,218}94.12 & \cellcolor[RGB]{255,199,206}35.00 & \cellcolor[RGB]{240,250,242}94.12 & \cellcolor[RGB]{210,242,216}79.17 & \cellcolor[RGB]{236,249,238}95.83 & \cellcolor[RGB]{255,199,206}20.83 & \cellcolor[RGB]{222,245,226}77.50 & \cellcolor[RGB]{243,251,245}75.60 & \cellcolor[RGB]{255,199,206}0.64 & \cellcolor[RGB]{255,199,206}0.95 & \cellcolor[RGB]{255,199,206}0.88 & \cellcolor[RGB]{255,199,206}9.78 & \cellcolor[RGB]{255,199,206}8177.44 \\
DuET ($\mathrm{E}\to\mathrm{T2I}\to\mathrm{E}$) & $[10,16)$ & \cellcolor[RGB]{212,242,218}94.17 & \cellcolor[RGB]{213,243,219}94.17 & \cellcolor[RGB]{198,239,206}94.96 & \cellcolor[RGB]{255,241,242}45.00 & \cellcolor[RGB]{224,246,228}95.00 & \cellcolor[RGB]{242,251,244}70.83 & \cellcolor[RGB]{198,239,206}97.50 & \cellcolor[RGB]{239,250,241}27.50 & \cellcolor[RGB]{255,241,242}69.17 & \cellcolor[RGB]{237,250,239}76.46 & \cellcolor[RGB]{255,199,206}0.78 & \cellcolor[RGB]{255,199,206}0.97 & \cellcolor[RGB]{255,199,206}0.91 & \cellcolor[RGB]{255,199,206}18.45 & \cellcolor[RGB]{255,199,206}2569.91 \\
DuET ($\mathrm{E}\to\mathrm{T2I}\to\mathrm{E}$) & $[10,20)$ & \cellcolor[RGB]{226,247,230}93.33 & \cellcolor[RGB]{198,239,206}97.50 & \cellcolor[RGB]{226,246,230}93.28 & \cellcolor[RGB]{255,216,221}39.17 & \cellcolor[RGB]{255,199,206}92.50 & \cellcolor[RGB]{204,240,211}80.83 & \cellcolor[RGB]{255,224,228}94.17 & \cellcolor[RGB]{223,246,227}31.67 & \cellcolor[RGB]{255,233,236}68.33 & \cellcolor[RGB]{235,249,238}76.74 & \cellcolor[RGB]{255,199,206}0.68 & \cellcolor[RGB]{255,199,206}0.95 & \cellcolor[RGB]{255,199,206}0.89 & \cellcolor[RGB]{255,199,206}12.54 & \cellcolor[RGB]{255,199,206}5568.90 \\
DuET ($\mathrm{E}\to\mathrm{T2I}\to\mathrm{E}$) & $[20,35)$ & \cellcolor[RGB]{226,247,230}93.33 & \cellcolor[RGB]{235,249,238}89.17 & \cellcolor[RGB]{240,250,242}92.44 & \cellcolor[RGB]{255,202,209}35.83 & \cellcolor[RGB]{224,246,228}95.00 & \cellcolor[RGB]{207,241,214}80.00 & \cellcolor[RGB]{236,249,238}95.83 & 23.33 & \cellcolor[RGB]{234,249,237}75.00 & \cellcolor[RGB]{244,251,245}75.53 & \cellcolor[RGB]{255,199,206}0.70 & \cellcolor[RGB]{255,199,206}0.96 & \cellcolor[RGB]{255,199,206}0.89 & \cellcolor[RGB]{255,199,206}10.95 & \cellcolor[RGB]{255,199,206}7090.06 \\
DuET ($\mathrm{E}\to\mathrm{T2I}\to\mathrm{E}$) & $[25,35)$ & \cellcolor[RGB]{255,252,253}91.59 & \cellcolor[RGB]{229,247,232}90.65 & \cellcolor[RGB]{238,250,241}92.55 & \cellcolor[RGB]{255,220,225}40.19 & \cellcolor[RGB]{228,247,232}94.79 & \cellcolor[RGB]{223,246,228}75.73 & \cellcolor[RGB]{247,252,248}95.33 & \cellcolor[RGB]{255,230,233}22.22 & \cellcolor[RGB]{247,253,248}72.28 & \cellcolor[RGB]{247,252,248}75.03 & \cellcolor[RGB]{255,199,206}0.75 & \cellcolor[RGB]{255,199,206}0.96 & \cellcolor[RGB]{255,199,206}0.90 & \cellcolor[RGB]{255,199,206}12.93 & \cellcolor[RGB]{255,199,206}5518.98 \\
DuET ($\mathrm{E}\to\mathrm{T2I}\to\mathrm{E}$) & $[35,45)$ & \cellcolor[RGB]{238,250,240}92.65 & \cellcolor[RGB]{238,250,241}88.52 & \cellcolor[RGB]{255,199,206}91.30 & \cellcolor[RGB]{255,218,223}39.71 & \cellcolor[RGB]{198,239,206}96.47 & \cellcolor[RGB]{218,244,223}77.11 & \cellcolor[RGB]{255,208,214}93.75 & \cellcolor[RGB]{251,254,252}24.14 & \cellcolor[RGB]{255,200,206}64.29 & \cellcolor[RGB]{255,199,206}73.13 & \cellcolor[RGB]{255,199,206}0.78 & \cellcolor[RGB]{255,199,206}0.97 & \cellcolor[RGB]{255,199,206}0.90 & \cellcolor[RGB]{255,199,206}13.96 & \cellcolor[RGB]{255,199,206}5005.39 \\
$\mathrm{E}\to\mathrm{E}^{*}\to\mathrm{E}$ & $[25,35)$ & \cellcolor[RGB]{255,199,206}89.52 & \cellcolor[RGB]{241,251,243}88.00 & \cellcolor[RGB]{233,248,236}92.86 & \cellcolor[RGB]{255,242,244}45.45 & \cellcolor[RGB]{237,250,240}94.29 & \cellcolor[RGB]{255,199,206}66.34 & \cellcolor[RGB]{255,199,206}93.48 & \cellcolor[RGB]{254,254,254}23.58 & \cellcolor[RGB]{255,228,231}67.65 & \cellcolor[RGB]{255,199,206}73.13 & 0.81 & 0.97 & 0.91 & 20.76 & 1897.70 \\
$\mathrm{E}\to\mathrm{E}^{*}\to\mathrm{E}$ & $[35,45)$ & \cellcolor[RGB]{255,233,235}90.83 & \cellcolor[RGB]{235,249,238}89.17 & 91.60 & 48.33 & \cellcolor[RGB]{255,251,252}93.28 & 67.50 & \cellcolor[RGB]{236,249,238}95.83 & \cellcolor[RGB]{248,253,249}25.00 & \cellcolor[RGB]{255,199,206}64.17 & \cellcolor[RGB]{255,247,248}73.93 & 0.81 & 0.97 & 0.91 & 20.84 & 1878.95 \\
Selective DuET & adaptive & \cellcolor[RGB]{212,242,218}94.17 & \cellcolor[RGB]{217,244,222}93.33 & \cellcolor[RGB]{198,239,206}94.96 & \cellcolor[RGB]{255,230,233}42.50 & \cellcolor[RGB]{239,250,241}94.17 & \cellcolor[RGB]{239,250,241}71.67 & 95.00 & \cellcolor[RGB]{242,251,244}26.67 & \cellcolor[RGB]{255,241,242}69.17 & \cellcolor[RGB]{242,251,244}75.72 & 0.79 & 0.97 & 0.91 & 19.00 & 2370.95 \\
\bottomrule
\end{tabular}
}
}
\end{table}

\begin{table}[!ht]
\centering
\noindent
\begin{minipage}[t]{0.38\linewidth}
\centering
\captionof{table}{\textbf{MagicBrush CLIP-T-gen (FLUX2-Klein 4B).} Metric reported in Table~2.}
\label{tab:appx-magicbrush}
\vspace{0.75em}
{\footnotesize
\setlength{\tabcolsep}{3pt}
\centering
\sbox0{\begin{tabular}{llc}
\toprule
Pipeline & Interval & CLIP-T-gen \\
\midrule
Baseline & --- & 0.317 \\
E$\to$T2I & $k=5$ & \cellcolor[RGB]{219,245,224}0.322 \\
E$\to$T2I & $k=10$ & \cellcolor[RGB]{205,241,212}0.324 \\
E$\to$T2I & $k=15$ & \cellcolor[RGB]{198,239,206}\textbf{0.325} \\
E$\to$T2I & $k=25$ & \cellcolor[RGB]{226,247,230}0.321 \\
E$\to$T2I & $k=35$ & \cellcolor[RGB]{247,253,248}0.318 \\
E$\to$T2I & $k=40$ & \cellcolor[RGB]{247,253,248}0.318 \\
E$\to$T2I & $k=45$ & 0.317 \\
DuET ($\mathrm{E}\to\mathrm{T2I}\to\mathrm{E}$) & $[10,16)$ & \cellcolor[RGB]{219,245,224}0.322 \\
DuET ($\mathrm{E}\to\mathrm{T2I}\to\mathrm{E}$) & $[10,20)$ & \cellcolor[RGB]{205,241,212}0.324 \\
DuET ($\mathrm{E}\to\mathrm{T2I}\to\mathrm{E}$) & $[15,30)$ & \cellcolor[RGB]{198,239,206}\textbf{0.325} \\
DuET ($\mathrm{E}\to\mathrm{T2I}\to\mathrm{E}$) & $[15,35)$ & \cellcolor[RGB]{198,239,206}\textbf{0.325} \\
DuET ($\mathrm{E}\to\mathrm{T2I}\to\mathrm{E}$) & $[20,30)$ & \cellcolor[RGB]{219,245,224}0.322 \\
DuET ($\mathrm{E}\to\mathrm{T2I}\to\mathrm{E}$) & $[20,35)$ & \cellcolor[RGB]{212,243,218}0.323 \\
DuET ($\mathrm{E}\to\mathrm{T2I}\to\mathrm{E}$) & $[25,35)$ & \cellcolor[RGB]{226,247,230}0.321 \\
DuET ($\mathrm{E}\to\mathrm{T2I}\to\mathrm{E}$) & $[25,40)$ & \cellcolor[RGB]{226,247,230}0.321 \\
DuET ($\mathrm{E}\to\mathrm{T2I}\to\mathrm{E}$) & $[35,45)$ & \cellcolor[RGB]{247,253,248}0.318 \\
Selective DuET & adaptive & \cellcolor[RGB]{240,251,242}0.319 \\
\bottomrule
\end{tabular}}
\ifdim\wd0>\linewidth\relax
  \resizebox{\linewidth}{!}{\usebox0}
\else
  \usebox0
\fi
}

\end{minipage}
\hfill
\begin{minipage}[t]{0.60\linewidth}
\centering
{\captionsetup{width=0.78\linewidth, margin={1.2em,0pt}}
\captionof{table}{\textbf{GEdit sweep (FLUX2-Klein 4B, GPT-4.1 judge).} Single-$k$ E$\to$T2I and double-$k$ DuET intervals beyond those in Table~2.}
\label{tab:appx-gedit-gpt}
}
\vspace{0.75em}
{\footnotesize
\setlength{\tabcolsep}{3pt}
\centering
\sbox0{\begin{tabular}{llccc}
\toprule
Pipeline & Interval & G\_SC & G\_PQ & G\_O \\
\midrule
Baseline & --- & 7.72 & 7.46 & 7.11 \\
E$\to$T2I & $k=5$ & \cellcolor[RGB]{255,199,206}5.31 & \cellcolor[RGB]{198,239,206}\textbf{8.50} & \cellcolor[RGB]{255,199,206}6.15 \\
E$\to$T2I & $k=10$ & \cellcolor[RGB]{255,210,215}5.79 & \cellcolor[RGB]{200,239,208}8.45 & \cellcolor[RGB]{255,217,222}6.47 \\
E$\to$T2I & $k=15$ & \cellcolor[RGB]{255,224,227}6.39 & \cellcolor[RGB]{210,242,216}8.28 & \cellcolor[RGB]{255,232,235}6.72 \\
E$\to$T2I & $k=25$ & \cellcolor[RGB]{255,244,245}7.26 & \cellcolor[RGB]{218,244,223}8.12 & \cellcolor[RGB]{253,254,254}7.12 \\
E$\to$T2I & $k=35$ & \cellcolor[RGB]{255,246,247}7.36 & \cellcolor[RGB]{231,248,234}7.89 & \cellcolor[RGB]{255,249,250}7.02 \\
E$\to$T2I & $k=40$ & \cellcolor[RGB]{255,244,246}7.28 & \cellcolor[RGB]{240,250,242}7.73 & \cellcolor[RGB]{255,240,242}6.86 \\
E$\to$T2I & $k=45$ & \cellcolor[RGB]{255,242,244}7.18 & \cellcolor[RGB]{247,253,248}7.59 & \cellcolor[RGB]{255,231,234}6.71 \\
DuET ($\mathrm{E}\to\mathrm{T2I}\to\mathrm{E}$) & $[10,16)$ & \cellcolor[RGB]{198,239,206}\textbf{8.08} & \cellcolor[RGB]{230,248,234}7.90 & \cellcolor[RGB]{198,239,206}\textbf{7.67} \\
DuET ($\mathrm{E}\to\mathrm{T2I}\to\mathrm{E}$) & $[10,20)$ & \cellcolor[RGB]{255,250,251}7.53 & \cellcolor[RGB]{213,243,219}8.21 & \cellcolor[RGB]{212,243,218}7.53 \\
DuET ($\mathrm{E}\to\mathrm{T2I}\to\mathrm{E}$) & $[15,30)$ & \cellcolor[RGB]{255,231,234}6.70 & \cellcolor[RGB]{210,242,216}8.27 & \cellcolor[RGB]{255,242,243}6.89 \\
DuET ($\mathrm{E}\to\mathrm{T2I}\to\mathrm{E}$) & $[15,35)$ & \cellcolor[RGB]{255,228,232}6.59 & \cellcolor[RGB]{211,242,217}8.26 & \cellcolor[RGB]{255,243,245}6.92 \\
DuET ($\mathrm{E}\to\mathrm{T2I}\to\mathrm{E}$) & $[20,30)$ & \cellcolor[RGB]{255,253,253}7.67 & \cellcolor[RGB]{224,246,229}8.01 & \cellcolor[RGB]{222,245,227}7.43 \\
DuET ($\mathrm{E}\to\mathrm{T2I}\to\mathrm{E}$) & $[20,35)$ & \cellcolor[RGB]{255,244,245}7.26 & \cellcolor[RGB]{215,243,221}8.18 & \cellcolor[RGB]{232,248,235}7.33 \\
DuET ($\mathrm{E}\to\mathrm{T2I}\to\mathrm{E}$) & $[25,35)$ & \cellcolor[RGB]{236,249,238}7.84 & \cellcolor[RGB]{224,246,229}8.01 & \cellcolor[RGB]{211,242,217}7.54 \\
$\mathrm{E}\to\mathrm{E}^{*}\to\mathrm{E}$ & $[25,35)$ & \cellcolor[RGB]{247,252,248}7.77 & \cellcolor[RGB]{254,254,254}7.47 & \cellcolor[RGB]{249,253,250}7.16 \\
DuET ($\mathrm{E}\to\mathrm{T2I}\to\mathrm{E}$) & $[25,40)$ & \cellcolor[RGB]{255,252,253}7.63 & \cellcolor[RGB]{220,245,225}8.09 & \cellcolor[RGB]{218,244,223}7.47 \\
DuET ($\mathrm{E}\to\mathrm{T2I}\to\mathrm{E}$) & $[35,45)$ & \cellcolor[RGB]{239,250,241}7.82 & \cellcolor[RGB]{231,248,234}7.89 & \cellcolor[RGB]{224,246,228}7.41 \\
$\mathrm{E}\to\mathrm{E}^{*}\to\mathrm{E}$ & $[35,45)$ & \cellcolor[RGB]{253,254,253}7.73 & 7.46 & \cellcolor[RGB]{255,253,253}7.09 \\
Selective DuET & adaptive & \cellcolor[RGB]{221,245,226}7.93 & \cellcolor[RGB]{241,251,243}7.70 & \cellcolor[RGB]{223,246,227}7.42 \\
\bottomrule
\end{tabular}}
\ifdim\wd0>0.74\linewidth\relax
  \resizebox{0.74\linewidth}{!}{\usebox0}
\else
  \usebox0
\fi
}

\end{minipage}
\smallskip
\end{table}

\begin{table}[!ht]
\centering
\caption{\textbf{GEdit ablation grid (FLUX2-Klein 4B, Qwen3.5 judge).} Factorial sweeps over conditioning variants; feeds Figure~6. T2IEP and I2IC denote the partial switches defined in \S3. Qwen baseline differs slightly from the GPT-4.1 baseline in Table~2.}
\label{tab:appx-gedit-qwen}
\noindent\textbf{Single-$k$ variants.}\par\smallskip

\noindent
\begin{minipage}[t]{0.58\linewidth}
\centering
{\footnotesize
\renewcommand{\arraystretch}{0.95}
\setlength{\tabcolsep}{3pt}
\setlength{\aboverulesep}{2pt}
\setlength{\belowrulesep}{1pt}
\centering
\sbox0{\begin{tabular}{l|ccc|ccc|ccc}
\toprule
 & \multicolumn{3}{c|}{E$\to$T2I} & \multicolumn{3}{c|}{T2I$\to$E} & \multicolumn{3}{c}{E$\to$T2IEP} \\
Interval & G\_SC & G\_PQ & G\_O & G\_SC & G\_PQ & G\_O & G\_SC & G\_PQ & G\_O \\
\midrule
--- & 7.91 & 7.65 & 7.33 & \textbf{7.91} & 7.65 & 7.33 & 7.91 & 7.65 & 7.33 \\
$k=5$ & \cellcolor[RGB]{255,199,206}6.09 & \cellcolor[RGB]{198,239,206}\textbf{8.27} & \cellcolor[RGB]{255,199,206}6.53 & \cellcolor[RGB]{255,252,252}7.79 & \cellcolor[RGB]{221,245,226}8.07 & \cellcolor[RGB]{198,239,206}\textbf{7.52} & \cellcolor[RGB]{255,199,206}0.61 & \cellcolor[RGB]{198,239,206}\textbf{8.38} & \cellcolor[RGB]{255,199,206}0.82 \\
$k=10$ & \cellcolor[RGB]{255,218,223}6.73 & \cellcolor[RGB]{201,240,209}8.23 & \cellcolor[RGB]{255,234,237}7.04 & \cellcolor[RGB]{255,236,238}7.17 & \cellcolor[RGB]{207,241,214}8.24 & \cellcolor[RGB]{255,251,251}7.26 & \cellcolor[RGB]{255,209,215}2.03 & \cellcolor[RGB]{213,243,219}8.18 & \cellcolor[RGB]{255,213,219}2.55 \\
$k=15$ & \cellcolor[RGB]{255,230,233}7.12 & \cellcolor[RGB]{209,242,216}8.14 & \cellcolor[RGB]{255,248,248}7.23 & \cellcolor[RGB]{255,225,229}6.74 & \cellcolor[RGB]{205,241,212}8.27 & \cellcolor[RGB]{255,239,241}7.04 & \cellcolor[RGB]{255,225,229}4.12 & \cellcolor[RGB]{230,247,233}7.97 & \cellcolor[RGB]{255,232,235}4.70 \\
$k=20$ & \cellcolor[RGB]{255,246,247}7.62 & \cellcolor[RGB]{220,245,225}8.02 & \cellcolor[RGB]{210,242,216}7.51 & \cellcolor[RGB]{255,215,220}6.33 & \cellcolor[RGB]{206,241,212}8.26 & \cellcolor[RGB]{255,223,227}6.73 & \cellcolor[RGB]{255,238,240}5.81 & \cellcolor[RGB]{235,249,238}7.90 & \cellcolor[RGB]{255,245,246}6.17 \\
$k=25$ & \cellcolor[RGB]{255,253,253}7.85 & \cellcolor[RGB]{226,247,230}7.96 & \cellcolor[RGB]{200,239,208}7.55 & \cellcolor[RGB]{255,208,214}6.04 & \cellcolor[RGB]{203,240,210}8.29 & \cellcolor[RGB]{255,213,218}6.54 & \cellcolor[RGB]{255,247,248}6.88 & \cellcolor[RGB]{243,251,244}7.80 & \cellcolor[RGB]{255,251,251}6.88 \\
$k=30$ & \cellcolor[RGB]{223,246,228}7.97 & \cellcolor[RGB]{235,249,238}7.86 & \cellcolor[RGB]{198,239,206}\textbf{7.56} & \cellcolor[RGB]{255,206,212}5.98 & \cellcolor[RGB]{204,240,211}8.28 & \cellcolor[RGB]{255,210,215}6.48 & \cellcolor[RGB]{255,251,252}7.51 & \cellcolor[RGB]{251,253,251}7.70 & \cellcolor[RGB]{255,253,254}7.20 \\
$k=35$ & \cellcolor[RGB]{218,244,223}7.98 & \cellcolor[RGB]{242,251,243}7.79 & \cellcolor[RGB]{217,244,223}7.48 & \cellcolor[RGB]{255,204,211}5.90 & \cellcolor[RGB]{200,239,208}8.33 & \cellcolor[RGB]{255,208,214}6.45 & \cellcolor[RGB]{255,253,253}7.74 & \cellcolor[RGB]{251,253,251}7.70 & \cellcolor[RGB]{255,254,254}7.32 \\
$k=40$ & \cellcolor[RGB]{203,240,210}8.01 & \cellcolor[RGB]{247,252,248}7.73 & \cellcolor[RGB]{217,244,223}7.48 & \cellcolor[RGB]{255,199,206}5.66 & \cellcolor[RGB]{198,239,206}8.35 & \cellcolor[RGB]{255,199,206}6.27 & \cellcolor[RGB]{255,254,254}7.86 & \cellcolor[RGB]{255,199,206}7.63 & \cellcolor[RGB]{245,252,246}7.34 \\
$k=45$ & \cellcolor[RGB]{198,239,206}\textbf{8.02} & \cellcolor[RGB]{254,254,254}7.66 & \cellcolor[RGB]{235,249,237}7.41 & \cellcolor[RGB]{255,200,207}5.73 & \cellcolor[RGB]{198,239,206}\textbf{8.36} & \cellcolor[RGB]{255,203,209}6.35 & \cellcolor[RGB]{198,239,206}\textbf{7.95} & \cellcolor[RGB]{252,254,252}7.68 & \cellcolor[RGB]{198,239,206}\textbf{7.39} \\
\bottomrule
\end{tabular}}
\ifdim\wd0>0.98\linewidth\relax
  \resizebox{0.98\linewidth}{!}{\usebox0}
\else
  \usebox0
\fi
}
\end{minipage}
\hfill
\begin{minipage}[t]{0.40\linewidth}
\centering
{\footnotesize
\renewcommand{\arraystretch}{0.95}
\setlength{\tabcolsep}{3pt}
\setlength{\aboverulesep}{2pt}
\setlength{\belowrulesep}{1pt}
\centering
\sbox0{\begin{tabular}{l|ccc|ccc}
\toprule
 & \multicolumn{3}{c|}{E$\to\mathrm{E}^{*}$} & \multicolumn{3}{c}{E$\to$I2IC} \\
Interval & G\_SC & G\_PQ & G\_O & G\_SC & G\_PQ & G\_O \\
\midrule
--- & 7.91 & 7.65 & 7.33 & 7.91 & 7.65 & 7.33 \\
$k=0$ & \cellcolor[RGB]{198,239,206}\textbf{8.44} & \cellcolor[RGB]{255,199,206}7.53 & \cellcolor[RGB]{198,239,206}\textbf{7.72} & \cellcolor[RGB]{255,199,206}7.57 & \cellcolor[RGB]{198,239,206}\textbf{8.18} & \cellcolor[RGB]{255,199,206}7.28 \\
$k=5$ & \cellcolor[RGB]{215,243,220}8.28 & \cellcolor[RGB]{198,239,206}\textbf{7.66} & \cellcolor[RGB]{214,243,219}7.61 & \cellcolor[RGB]{230,248,233}8.01 & \cellcolor[RGB]{220,245,225}7.97 & \cellcolor[RGB]{216,244,221}7.50 \\
$k=10$ & \cellcolor[RGB]{233,248,236}8.11 & \cellcolor[RGB]{198,239,206}\textbf{7.66} & \cellcolor[RGB]{231,248,234}7.49 & \cellcolor[RGB]{212,243,218}8.08 & \cellcolor[RGB]{225,246,230}7.92 & \cellcolor[RGB]{198,239,206}\textbf{7.58} \\
$k=15$ & \cellcolor[RGB]{234,249,237}8.10 & \cellcolor[RGB]{255,240,242}7.62 & \cellcolor[RGB]{238,250,241}7.44 & \cellcolor[RGB]{198,239,206}\textbf{8.14} & \cellcolor[RGB]{232,248,235}7.86 & \cellcolor[RGB]{204,240,211}7.55 \\
$k=25$ & \cellcolor[RGB]{233,248,236}8.11 & \cellcolor[RGB]{255,231,234}7.60 & \cellcolor[RGB]{241,251,243}7.42 & \cellcolor[RGB]{220,245,225}8.05 & \cellcolor[RGB]{252,254,253}7.67 & \cellcolor[RGB]{236,249,239}7.41 \\
$k=35$ & \cellcolor[RGB]{246,252,247}7.99 & \cellcolor[RGB]{255,245,246}7.63 & \cellcolor[RGB]{253,254,253}7.34 & \cellcolor[RGB]{242,251,244}7.96 & \cellcolor[RGB]{248,253,249}7.71 & \cellcolor[RGB]{241,251,243}7.39 \\
$k=40$ & \cellcolor[RGB]{245,252,246}8.00 & \cellcolor[RGB]{255,245,246}7.63 & \cellcolor[RGB]{249,253,249}7.37 & \cellcolor[RGB]{235,249,237}7.99 & 7.65 & \cellcolor[RGB]{241,251,243}7.39 \\
$k=45$ & \cellcolor[RGB]{251,254,252}7.94 & \cellcolor[RGB]{255,231,234}7.60 & \cellcolor[RGB]{255,199,206}7.32 & \cellcolor[RGB]{255,248,249}7.87 & \cellcolor[RGB]{255,199,206}7.63 & \cellcolor[RGB]{255,210,215}7.29 \\
\bottomrule
\end{tabular}}
\ifdim\wd0>0.98\linewidth\relax
  \resizebox{0.98\linewidth}{!}{\usebox0}
\else
  \usebox0
\fi
}
\end{minipage}
\smallskip

\smallskip
\noindent\textbf{Double-$k$ intervals.}\par\smallskip

\noindent
\begin{minipage}[t]{0.32\linewidth}
\centering
{\footnotesize
\renewcommand{\arraystretch}{0.95}
\setlength{\tabcolsep}{3pt}
\setlength{\aboverulesep}{2pt}
\setlength{\belowrulesep}{1pt}
\centering
\sbox0{\begin{tabular}{llccc}
\toprule
Variant & Interval & G\_SC & G\_PQ & G\_O \\
\midrule
Baseline & --- & 7.91 & 7.65 & 7.33 \\
DuET & $[0,10)$ & \cellcolor[RGB]{255,228,232}7.17 & \cellcolor[RGB]{200,239,208}8.24 & \cellcolor[RGB]{255,248,249}7.26 \\
DuET & $[0,15)$ & \cellcolor[RGB]{255,213,218}6.74 & \cellcolor[RGB]{198,239,206}\textbf{8.27} & \cellcolor[RGB]{255,227,231}7.04 \\
DuET & $[0,20)$ & \cellcolor[RGB]{255,199,206}6.33 & \cellcolor[RGB]{198,239,206}8.26 & \cellcolor[RGB]{255,199,206}6.73 \\
DuET & $[10,20)$ & \cellcolor[RGB]{255,253,254}7.88 & \cellcolor[RGB]{218,244,223}8.05 & \cellcolor[RGB]{201,239,208}7.67 \\
DuET & $[10,25)$ & \cellcolor[RGB]{255,232,235}7.27 & \cellcolor[RGB]{207,241,213}8.17 & \cellcolor[RGB]{245,252,246}7.39 \\
DuET & $[10,30)$ & \cellcolor[RGB]{255,221,225}6.96 & \cellcolor[RGB]{203,240,210}8.21 & \cellcolor[RGB]{255,243,245}7.21 \\
DuET & $[15,25)$ & \cellcolor[RGB]{228,247,231}7.99 & \cellcolor[RGB]{226,247,230}7.96 & \cellcolor[RGB]{199,239,207}7.68 \\
DuET & $[15,30)$ & \cellcolor[RGB]{255,239,241}7.48 & \cellcolor[RGB]{217,244,222}8.06 & \cellcolor[RGB]{236,249,238}7.45 \\
DuET & $[15,35)$ & \cellcolor[RGB]{255,234,236}7.32 & \cellcolor[RGB]{212,243,218}8.11 & \cellcolor[RGB]{248,253,249}7.37 \\
DuET & $[20,30)$ & \cellcolor[RGB]{211,242,217}8.04 & \cellcolor[RGB]{229,247,232}7.93 & \cellcolor[RGB]{198,239,206}\textbf{7.69} \\
DuET & $[20,35)$ & \cellcolor[RGB]{255,250,251}7.79 & \cellcolor[RGB]{222,245,227}8.00 & \cellcolor[RGB]{217,244,222}7.57 \\
DuET & $[20,40)$ & \cellcolor[RGB]{255,247,248}7.70 & \cellcolor[RGB]{221,245,226}8.01 & \cellcolor[RGB]{223,246,227}7.53 \\
DuET & $[25,35)$ & \cellcolor[RGB]{201,239,208}8.07 & \cellcolor[RGB]{234,249,237}7.87 & \cellcolor[RGB]{207,241,214}7.63 \\
DuET & $[25,40)$ & \cellcolor[RGB]{244,252,246}7.94 & \cellcolor[RGB]{227,247,231}7.95 & \cellcolor[RGB]{210,242,216}7.61 \\
DuET & $[25,45)$ & \cellcolor[RGB]{255,253,254}7.88 & \cellcolor[RGB]{223,246,228}7.99 & \cellcolor[RGB]{210,242,216}7.61 \\
DuET & $[30,40)$ & \cellcolor[RGB]{198,239,206}\textbf{8.08} & \cellcolor[RGB]{236,249,239}7.85 & \cellcolor[RGB]{209,242,215}7.62 \\
DuET & $[30,45)$ & \cellcolor[RGB]{231,248,234}7.98 & \cellcolor[RGB]{233,249,236}7.88 & \cellcolor[RGB]{215,243,220}7.58 \\
DuET & $[30,50)$ & \cellcolor[RGB]{234,249,237}7.97 & \cellcolor[RGB]{235,249,238}7.86 & \cellcolor[RGB]{218,244,223}7.56 \\
DuET & $[35,45)$ & \cellcolor[RGB]{228,247,231}7.99 & \cellcolor[RGB]{244,252,246}7.76 & \cellcolor[RGB]{228,247,231}7.50 \\
DuET & $[35,50)$ & \cellcolor[RGB]{231,248,234}7.98 & \cellcolor[RGB]{242,251,243}7.79 & \cellcolor[RGB]{231,248,234}7.48 \\
DuET & $[40,50)$ & \cellcolor[RGB]{221,245,226}8.01 & \cellcolor[RGB]{247,252,248}7.73 & \cellcolor[RGB]{231,248,234}7.48 \\
\bottomrule
\end{tabular}}
\ifdim\wd0>\linewidth\relax
  \resizebox{\linewidth}{!}{\usebox0}
\else
  \usebox0
\fi
}
\end{minipage}
\hfill
\begin{minipage}[t]{0.66\linewidth}
\centering
{\footnotesize
\renewcommand{\arraystretch}{0.95}
\setlength{\tabcolsep}{3pt}
\setlength{\aboverulesep}{2pt}
\setlength{\belowrulesep}{1pt}
\centering
\sbox0{\begin{tabular}{l|ccc|ccc|ccc}
\toprule
 & \multicolumn{3}{c|}{$\mathrm{E}\to\mathrm{E}^{*}\to\mathrm{E}$} & \multicolumn{3}{c|}{E$\to$T2IEP$\to$E} & \multicolumn{3}{c}{E$\to$I2IC$\to$E} \\
Interval & G\_SC & G\_PQ & G\_O & G\_SC & G\_PQ & G\_O & G\_SC & G\_PQ & G\_O \\
\midrule
--- & 7.91 & 7.65 & 7.33 & \textbf{7.91} & 7.65 & 7.33 & 7.91 & 7.65 & 7.33 \\
$[0,10)$ & \cellcolor[RGB]{211,242,217}8.11 & \cellcolor[RGB]{226,247,230}7.66 & \cellcolor[RGB]{201,239,208}7.49 & \cellcolor[RGB]{255,199,206}2.03 & \cellcolor[RGB]{198,239,206}\textbf{8.18} & \cellcolor[RGB]{255,199,206}2.55 & \cellcolor[RGB]{212,243,218}8.08 & \cellcolor[RGB]{198,239,206}\textbf{7.92} & \cellcolor[RGB]{198,239,206}\textbf{7.58} \\
$[0,15)$ & \cellcolor[RGB]{213,243,219}8.10 & \cellcolor[RGB]{255,227,230}7.62 & \cellcolor[RGB]{218,244,223}7.44 & \cellcolor[RGB]{255,218,223}4.12 & \cellcolor[RGB]{220,245,225}7.97 & \cellcolor[RGB]{255,224,228}4.70 & \cellcolor[RGB]{198,239,206}\textbf{8.14} & \cellcolor[RGB]{210,242,216}7.86 & \cellcolor[RGB]{204,240,211}7.55 \\
$[10,20)$ & \cellcolor[RGB]{198,239,206}\textbf{8.17} & \cellcolor[RGB]{198,239,206}\textbf{7.67} & \cellcolor[RGB]{198,239,206}\textbf{7.50} & \cellcolor[RGB]{255,229,232}5.21 & \cellcolor[RGB]{237,250,240}7.81 & \cellcolor[RGB]{255,233,235}5.47 & \cellcolor[RGB]{200,239,208}8.13 & \cellcolor[RGB]{225,246,229}7.79 & \cellcolor[RGB]{209,242,215}7.53 \\
$[15,30)$ & \cellcolor[RGB]{224,246,228}8.05 & 7.65 & \cellcolor[RGB]{221,245,226}7.43 & \cellcolor[RGB]{255,226,230}4.94 & \cellcolor[RGB]{225,246,230}7.92 & \cellcolor[RGB]{255,233,235}5.46 & \cellcolor[RGB]{225,246,229}8.03 & \cellcolor[RGB]{233,249,236}7.75 & \cellcolor[RGB]{227,247,231}7.45 \\
$[15,35)$ & \cellcolor[RGB]{204,240,211}8.14 & \cellcolor[RGB]{255,236,238}7.63 & \cellcolor[RGB]{208,241,214}7.47 & \cellcolor[RGB]{255,222,226}4.46 & \cellcolor[RGB]{220,245,225}7.97 & \cellcolor[RGB]{255,227,231}4.99 & \cellcolor[RGB]{200,239,208}8.13 & \cellcolor[RGB]{229,247,233}7.77 & \cellcolor[RGB]{207,241,213}7.54 \\
$[20,30)$ & \cellcolor[RGB]{219,245,224}8.07 & \cellcolor[RGB]{255,199,206}7.59 & \cellcolor[RGB]{238,250,240}7.38 & \cellcolor[RGB]{255,249,250}7.36 & \cellcolor[RGB]{233,248,236}7.85 & \cellcolor[RGB]{255,253,253}7.19 & \cellcolor[RGB]{230,248,233}8.01 & \cellcolor[RGB]{244,252,245}7.70 & \cellcolor[RGB]{236,249,239}7.41 \\
$[20,35)$ & \cellcolor[RGB]{219,245,224}8.07 & \cellcolor[RGB]{255,199,206}7.59 & \cellcolor[RGB]{238,250,240}7.38 & \cellcolor[RGB]{255,239,241}6.31 & \cellcolor[RGB]{233,248,236}7.85 & \cellcolor[RGB]{255,245,246}6.51 & \cellcolor[RGB]{227,247,231}8.02 & \cellcolor[RGB]{250,253,251}7.67 & \cellcolor[RGB]{239,250,241}7.40 \\
$[25,35)$ & \cellcolor[RGB]{228,247,232}8.03 & \cellcolor[RGB]{255,227,230}7.62 & \cellcolor[RGB]{241,251,243}7.37 & \cellcolor[RGB]{255,252,252}7.61 & \cellcolor[RGB]{244,251,245}7.75 & \cellcolor[RGB]{255,254,254}7.32 & \cellcolor[RGB]{227,247,231}8.02 & \cellcolor[RGB]{255,236,238}7.64 & \cellcolor[RGB]{243,251,245}7.38 \\
$[25,40)$ & \cellcolor[RGB]{230,248,234}8.02 & \cellcolor[RGB]{255,236,238}7.63 & \cellcolor[RGB]{244,252,246}7.36 & \cellcolor[RGB]{255,247,248}7.13 & \cellcolor[RGB]{241,251,242}7.78 & \cellcolor[RGB]{255,251,251}7.02 & \cellcolor[RGB]{237,250,240}7.98 & \cellcolor[RGB]{255,199,206}7.62 & \cellcolor[RGB]{245,252,247}7.37 \\
$[35,45)$ & \cellcolor[RGB]{226,247,230}8.04 & \cellcolor[RGB]{255,208,214}7.60 & \cellcolor[RGB]{241,251,243}7.37 & \cellcolor[RGB]{255,253,253}7.78 & \cellcolor[RGB]{250,253,251}7.69 & 7.33 & \cellcolor[RGB]{242,251,244}7.96 & \cellcolor[RGB]{246,252,247}7.69 & \cellcolor[RGB]{241,251,243}7.39 \\
$[35,50)$ & \cellcolor[RGB]{237,250,239}7.99 & \cellcolor[RGB]{255,236,238}7.63 & \cellcolor[RGB]{251,254,252}7.34 & \cellcolor[RGB]{255,253,253}7.74 & \cellcolor[RGB]{249,253,250}7.70 & \cellcolor[RGB]{255,254,254}7.32 & \cellcolor[RGB]{242,251,244}7.96 & \cellcolor[RGB]{242,251,244}7.71 & \cellcolor[RGB]{241,251,243}7.39 \\
$[40,50)$ & \cellcolor[RGB]{235,249,238}8.00 & \cellcolor[RGB]{255,236,238}7.63 & \cellcolor[RGB]{241,251,243}7.37 & \cellcolor[RGB]{255,254,254}7.86 & \cellcolor[RGB]{255,199,206}7.63 & \cellcolor[RGB]{198,239,206}\textbf{7.34} & \cellcolor[RGB]{235,249,237}7.99 & 7.65 & \cellcolor[RGB]{241,251,243}7.39 \\
\bottomrule
\end{tabular}}
\ifdim\wd0>\linewidth\relax
  \resizebox{\linewidth}{!}{\usebox0}
\else
  \usebox0
\fi
}
\end{minipage}
\smallskip

\end{table}


\paragraph{Reading the factorial ablations (Table~\ref{tab:appx-gedit-qwen}).}
For each variant, we consider the best score attained in each GEdit metric across all tested single-$k$ and double-$k$ intervals. As reference points, the pure-editing baseline achieves $7.91/7.65/7.33$ (G\_SC/G\_PQ/G\_O), while the best $\mathrm{T2I}$ schedule reaches $8.09/8.27/7.69$.

The ablations reveal complementary roles for the two components of DuET's switch:

\begin{itemize}
\item \textbf{Better instructions alone.} The $\mathrm{E}^{*}$ variant, which replaces the original instruction with a VLM-improved edit prompt, substantially increases semantic correctness, achieving the highest G\_SC among all variants ($8.44$). However, G\_PQ remains essentially unchanged relative to the baseline ($7.66$--$7.67$ versus $7.65$). Better instructions help the model understand \emph{what} edit to perform, but persistent image conditioning continues to constrain image quality and realism.

\item \textbf{Caption substitution without image release.} The $\mathrm{I2IC}$ variant replaces $p_{\mathrm{edit}}$ with $c_{\mathrm{tgt}}$ while retaining source-image conditioning. This improves both semantic fidelity and perceptual quality, reaching G\_SC $=8.14$ and G\_PQ $=8.18$. The result suggests that target-scene captions provide a stronger conditioning signal than edit instructions alone. However, performance remains below full $\mathrm{T2I}$ in G\_PQ, indicating that retaining the source-image condition still limits the model's ability to generate the most natural-looking outcomes.

\item \textbf{Image release without caption substitution.} The $\mathrm{T2IEP}$ variant drops the source image while retaining only $p_{\mathrm{edit}}$. This produces the opposite behavior: G\_PQ becomes very strong, reaching $8.38$, but semantic fidelity deteriorates sharply, with G\_SC falling as low as $0.61$ for early switches and peaking at only $7.95$. Once the source image is removed, the model loses both the visual reference and any explicit description of the intended target scene; the edit instruction specifies only the desired \emph{change}, leaving the final image under-constrained.
\end{itemize}

Taken together, these results suggest that the two components address different failure modes. Target-scene captions primarily improve semantic specification, while temporary removal of image conditioning primarily improves perceptual quality. The full $\mathrm{E}\rightarrow\mathrm{T2I}\rightarrow\mathrm{E}$ schedule is the only variant that combines both effects, leading to simultaneous gains in G\_SC, G\_PQ, and G\_O beyond those achieved by any single-factor control.

\begin{table}[!ht]
\centering
\caption{\textbf{ImgEdit per-category scores (GPT-4o judge).} Category breakdown for FLUX2-Klein 4B, 9B, and BAGEL: E$\to$T2I, DuET, and selected ablation rows (T2IEP and I2IC as in \S3); 9B/BAGEL blocks include additional double-$k$ intervals not shown in Table~2.}
\label{tab:appx-imgedit}
\vspace{0.3em}
{\footnotesize
\setlength{\tabcolsep}{3pt}
\renewcommand{\arraystretch}{0.85}
\resizebox{\linewidth}{!}{
\begin{tabular}{llcccccccccc}
\toprule
Pipeline & Interval & background & adjust & style & extract & remove & add & replace & compose & action & overall \\
\midrule
\multicolumn{12}{l}{\textbf{FLUX2-Klein 4B}} \\
\midrule

Baseline & --- & 4.21 & 3.85 & 4.80 & 1.47 & 2.99 & 4.40 & 4.18 & 3.26 & 4.70 & 3.74 \\
T2I$\to$E & $k=25$ & \cellcolor[RGB]{224,246,229}4.40 & \cellcolor[RGB]{255,199,206}3.04 & \cellcolor[RGB]{255,236,239}4.70 & \cellcolor[RGB]{198,239,206}\textbf{3.90} & \cellcolor[RGB]{218,244,223}3.78 & \cellcolor[RGB]{245,252,247}4.45 & \cellcolor[RGB]{255,199,206}3.55 & \cellcolor[RGB]{255,199,206}2.01 & \cellcolor[RGB]{255,236,239}4.36 & \cellcolor[RGB]{227,247,231}3.94 \\
E$\to$T2I & $k=5$ & \cellcolor[RGB]{220,245,225}4.43 & \cellcolor[RGB]{255,208,214}3.18 & \cellcolor[RGB]{255,247,248}4.76 & \cellcolor[RGB]{205,241,212}3.58 & \cellcolor[RGB]{198,239,206}\textbf{4.21} & \cellcolor[RGB]{242,251,243}4.47 & \cellcolor[RGB]{255,212,217}3.70 & \cellcolor[RGB]{255,205,211}2.14 & \cellcolor[RGB]{255,199,206}3.65 & \cellcolor[RGB]{225,246,229}3.96 \\
E$\to$T2I & $k=10$ & \cellcolor[RGB]{215,243,220}4.46 & \cellcolor[RGB]{255,241,243}3.66 & \cellcolor[RGB]{243,251,245}4.83 & \cellcolor[RGB]{228,247,232}2.58 & \cellcolor[RGB]{227,247,231}3.58 & \cellcolor[RGB]{221,245,226}4.58 & \cellcolor[RGB]{255,221,225}3.80 & \cellcolor[RGB]{255,199,206}1.99 & \cellcolor[RGB]{255,213,219}3.93 & \cellcolor[RGB]{237,250,239}3.87 \\
E$\to$T2I & $k=15$ & \cellcolor[RGB]{210,242,216}4.49 & \cellcolor[RGB]{240,250,242}4.06 & \cellcolor[RGB]{224,246,228}4.88 & \cellcolor[RGB]{242,251,244}2.00 & \cellcolor[RGB]{235,249,238}3.40 & \cellcolor[RGB]{210,242,217}4.64 & \cellcolor[RGB]{255,242,244}4.04 & \cellcolor[RGB]{255,239,241}2.91 & \cellcolor[RGB]{255,229,233}4.23 & \cellcolor[RGB]{229,247,232}3.93 \\
E$\to$T2I & $k=25$ & \cellcolor[RGB]{215,243,220}4.46 & \cellcolor[RGB]{202,240,209}4.62 & \cellcolor[RGB]{205,241,212}4.93 & \cellcolor[RGB]{243,251,245}1.94 & \cellcolor[RGB]{240,250,242}3.30 & \cellcolor[RGB]{209,242,215}4.65 & \cellcolor[RGB]{230,248,234}4.30 & \cellcolor[RGB]{205,241,212}3.70 & \cellcolor[RGB]{255,254,254}4.69 & \cellcolor[RGB]{208,242,215}4.08 \\
E$\to$T2IEP & $k=25$ & \cellcolor[RGB]{255,252,252}4.17 & \cellcolor[RGB]{239,250,242}4.07 & \cellcolor[RGB]{220,245,225}4.89 & \cellcolor[RGB]{246,252,247}1.82 & \cellcolor[RGB]{245,252,246}3.19 & \cellcolor[RGB]{229,247,232}4.54 & \cellcolor[RGB]{255,251,251}4.14 & \cellcolor[RGB]{255,244,246}3.03 & \cellcolor[RGB]{255,225,228}4.14 & \cellcolor[RGB]{240,250,242}3.85 \\
E$\to$I2IC & $k=25$ & \cellcolor[RGB]{255,254,254}4.20 & \cellcolor[RGB]{247,252,248}3.96 & \cellcolor[RGB]{247,252,248}4.82 & \cellcolor[RGB]{255,248,249}1.46 & \cellcolor[RGB]{252,254,252}3.05 & \cellcolor[RGB]{245,252,247}4.45 & \cellcolor[RGB]{242,251,244}4.24 & \cellcolor[RGB]{232,248,235}3.46 & \cellcolor[RGB]{222,245,226}4.82 & \cellcolor[RGB]{249,253,250}3.78 \\
E$\to$T2I & $k=35$ & \cellcolor[RGB]{243,251,245}4.28 & \cellcolor[RGB]{215,243,220}4.43 & \cellcolor[RGB]{209,242,215}4.92 & \cellcolor[RGB]{251,253,251}1.63 & \cellcolor[RGB]{245,252,246}3.20 & \cellcolor[RGB]{231,248,234}4.53 & \cellcolor[RGB]{234,249,237}4.28 & \cellcolor[RGB]{208,241,214}3.68 & \cellcolor[RGB]{241,251,243}4.75 & \cellcolor[RGB]{229,247,232}3.93 \\
E$\to$T2I & $k=40$ & \cellcolor[RGB]{239,250,241}4.31 & \cellcolor[RGB]{231,248,234}4.19 & \cellcolor[RGB]{213,243,219}4.91 & \cellcolor[RGB]{251,254,252}1.61 & \cellcolor[RGB]{247,252,248}3.15 & \cellcolor[RGB]{234,249,237}4.51 & \cellcolor[RGB]{236,249,239}4.27 & \cellcolor[RGB]{227,247,230}3.51 & \cellcolor[RGB]{219,245,224}4.83 & \cellcolor[RGB]{236,249,238}3.88 \\
E$\to$T2I & $k=45$ & \cellcolor[RGB]{243,251,245}4.28 & \cellcolor[RGB]{240,250,242}4.06 & \cellcolor[RGB]{228,247,232}4.87 & \cellcolor[RGB]{254,254,254}1.48 & \cellcolor[RGB]{247,252,248}3.15 & \cellcolor[RGB]{247,252,248}4.44 & \cellcolor[RGB]{248,253,249}4.21 & \cellcolor[RGB]{243,251,245}3.36 & \cellcolor[RGB]{203,240,210}4.89 & \cellcolor[RGB]{244,251,245}3.82 \\
DuET ($\mathrm{E}\to\mathrm{T2I}\to\mathrm{E}$) & $[10,16)$ & \cellcolor[RGB]{207,241,214}4.51 & \cellcolor[RGB]{236,249,239}4.12 & \cellcolor[RGB]{220,245,225}4.89 & \cellcolor[RGB]{245,252,246}1.88 & \cellcolor[RGB]{235,249,238}3.41 & \cellcolor[RGB]{223,246,228}4.57 & \cellcolor[RGB]{242,251,244}4.24 & \cellcolor[RGB]{201,239,208}3.74 & \cellcolor[RGB]{211,242,217}4.86 & \cellcolor[RGB]{221,245,225}3.99 \\
DuET ($\mathrm{E}\to\mathrm{T2I}\to\mathrm{E}$) & $[10,20)$ & \cellcolor[RGB]{209,242,215}4.50 & \cellcolor[RGB]{223,246,227}4.31 & \cellcolor[RGB]{198,239,206}\textbf{4.95} & \cellcolor[RGB]{233,249,236}2.37 & \cellcolor[RGB]{217,244,222}3.79 & \cellcolor[RGB]{198,239,206}\textbf{4.71} & \cellcolor[RGB]{198,239,206}\textbf{4.46} & \cellcolor[RGB]{230,248,233}3.48 & \cellcolor[RGB]{236,249,238}4.77 & \cellcolor[RGB]{198,239,206}\textbf{4.16} \\
E$\to$T2IEP$\to$E & $[10,20)$ & \cellcolor[RGB]{255,199,206}3.28 & \cellcolor[RGB]{255,212,218}3.24 & \cellcolor[RGB]{255,199,206}4.49 & \cellcolor[RGB]{247,252,248}1.81 & \cellcolor[RGB]{248,253,249}3.12 & \cellcolor[RGB]{255,199,206}4.26 & \cellcolor[RGB]{255,226,230}3.86 & \cellcolor[RGB]{255,232,234}2.74 & \cellcolor[RGB]{255,236,238}4.35 & \cellcolor[RGB]{255,199,206}3.48 \\
E$\to$I2IC$\to$E & $[10,20)$ & \cellcolor[RGB]{223,246,227}4.41 & \cellcolor[RGB]{250,253,251}3.91 & \cellcolor[RGB]{247,252,248}4.82 & \cellcolor[RGB]{251,254,252}1.60 & \cellcolor[RGB]{252,254,252}3.04 & \cellcolor[RGB]{227,247,231}4.55 & \cellcolor[RGB]{250,253,251}4.20 & \cellcolor[RGB]{248,253,249}3.32 & \cellcolor[RGB]{225,246,229}4.81 & \cellcolor[RGB]{244,251,245}3.82 \\
DuET ($\mathrm{E}\to\mathrm{T2I}\to\mathrm{E}$) & $[15,30)$ & \cellcolor[RGB]{198,239,206}\textbf{4.57} & \cellcolor[RGB]{202,240,209}4.62 & \cellcolor[RGB]{205,241,212}4.93 & \cellcolor[RGB]{242,251,244}2.00 & \cellcolor[RGB]{226,247,230}3.59 & \cellcolor[RGB]{199,239,207}4.70 & \cellcolor[RGB]{255,251,251}4.14 & \cellcolor[RGB]{230,248,233}3.48 & \cellcolor[RGB]{246,252,247}4.73 & \cellcolor[RGB]{203,240,210}4.12 \\
DuET ($\mathrm{E}\to\mathrm{T2I}\to\mathrm{E}$) & $[15,35)$ & \cellcolor[RGB]{199,239,207}4.56 & \cellcolor[RGB]{217,244,222}4.40 & \cellcolor[RGB]{213,243,219}4.91 & \cellcolor[RGB]{239,250,241}2.13 & \cellcolor[RGB]{234,249,237}3.42 & \cellcolor[RGB]{209,242,215}4.65 & \cellcolor[RGB]{255,248,249}4.11 & \cellcolor[RGB]{244,252,246}3.35 & \cellcolor[RGB]{255,254,254}4.69 & \cellcolor[RGB]{211,242,217}4.06 \\
DuET ($\mathrm{E}\to\mathrm{T2I}\to\mathrm{E}$) & $[20,30)$ & \cellcolor[RGB]{218,244,223}4.44 & \cellcolor[RGB]{205,241,212}4.57 & \cellcolor[RGB]{216,244,222}4.90 & \cellcolor[RGB]{244,252,246}1.91 & \cellcolor[RGB]{238,250,240}3.34 & \cellcolor[RGB]{214,243,220}4.62 & \cellcolor[RGB]{230,248,234}4.30 & \cellcolor[RGB]{201,239,208}3.74 & \cellcolor[RGB]{217,244,222}4.84 & \cellcolor[RGB]{210,242,216}4.07 \\
DuET ($\mathrm{E}\to\mathrm{T2I}\to\mathrm{E}$) & $[20,35)$ & \cellcolor[RGB]{213,243,219}4.47 & \cellcolor[RGB]{211,242,217}4.49 & \cellcolor[RGB]{205,241,212}4.93 & \cellcolor[RGB]{244,252,246}1.90 & \cellcolor[RGB]{241,251,243}3.27 & \cellcolor[RGB]{207,241,213}4.66 & \cellcolor[RGB]{246,252,248}4.22 & \cellcolor[RGB]{198,239,206}\textbf{3.77} & \cellcolor[RGB]{241,251,243}4.75 & \cellcolor[RGB]{214,243,220}4.04 \\
DuET ($\mathrm{E}\to\mathrm{T2I}\to\mathrm{E}$) & $[25,35)$ & \cellcolor[RGB]{220,245,225}4.43 & \cellcolor[RGB]{207,241,214}4.54 & \cellcolor[RGB]{209,242,215}4.92 & \cellcolor[RGB]{245,252,247}1.86 & \cellcolor[RGB]{239,250,241}3.32 & \cellcolor[RGB]{210,242,217}4.64 & \cellcolor[RGB]{236,249,239}4.27 & \cellcolor[RGB]{200,239,207}3.75 & \cellcolor[RGB]{225,246,229}4.81 & \cellcolor[RGB]{214,243,220}4.04 \\
$\mathrm{E}\to\mathrm{E}^{*}\to\mathrm{E}$ & $[25,35)$ & \cellcolor[RGB]{255,247,248}4.08 & \cellcolor[RGB]{255,253,253}3.83 & \cellcolor[RGB]{255,244,245}4.74 & \cellcolor[RGB]{255,199,206}1.38 & \cellcolor[RGB]{255,199,206}2.98 & \cellcolor[RGB]{255,239,241}4.36 & \cellcolor[RGB]{255,251,251}4.14 & \cellcolor[RGB]{255,251,252}3.19 & \cellcolor[RGB]{198,239,206}\textbf{4.91} & \cellcolor[RGB]{255,244,245}3.69 \\
DuET ($\mathrm{E}\to\mathrm{T2I}\to\mathrm{E}$) & $[25,40)$ & \cellcolor[RGB]{221,245,226}4.42 & \cellcolor[RGB]{198,239,206}\textbf{4.68} & \cellcolor[RGB]{201,240,209}4.94 & \cellcolor[RGB]{248,253,249}1.74 & \cellcolor[RGB]{237,250,240}3.36 & \cellcolor[RGB]{201,240,209}4.69 & \cellcolor[RGB]{234,249,237}4.28 & \cellcolor[RGB]{211,242,217}3.65 & \cellcolor[RGB]{246,252,247}4.73 & \cellcolor[RGB]{210,242,216}4.07 \\
DuET ($\mathrm{E}\to\mathrm{T2I}\to\mathrm{E}$) & $[35,45)$ & \cellcolor[RGB]{236,249,238}4.33 & \cellcolor[RGB]{220,245,225}4.35 & \cellcolor[RGB]{205,241,212}4.93 & \cellcolor[RGB]{250,253,251}1.66 & \cellcolor[RGB]{244,252,246}3.21 & \cellcolor[RGB]{236,249,239}4.50 & \cellcolor[RGB]{230,248,234}4.30 & \cellcolor[RGB]{235,249,238}3.43 & \cellcolor[RGB]{214,243,220}4.85 & \cellcolor[RGB]{230,248,234}3.92 \\
$\mathrm{E}\to\mathrm{E}^{*}\to\mathrm{E}$ & $[35,45)$ & \cellcolor[RGB]{255,251,251}4.15 & \cellcolor[RGB]{255,252,252}3.81 & 4.80 & \cellcolor[RGB]{253,254,253}1.55 & \cellcolor[RGB]{254,254,254}3.01 & 4.40 & \cellcolor[RGB]{250,253,251}4.20 & 3.26 & \cellcolor[RGB]{227,247,231}4.80 & \cellcolor[RGB]{255,252,253}3.73 \\
Selective DuET & adaptive & \cellcolor[RGB]{247,252,248}4.26 & \cellcolor[RGB]{251,254,252}3.90 & \cellcolor[RGB]{228,247,232}4.87 & \cellcolor[RGB]{247,252,248}1.81 & \cellcolor[RGB]{239,250,241}3.33 & \cellcolor[RGB]{238,250,240}4.49 & \cellcolor[RGB]{228,247,232}4.31 & \cellcolor[RGB]{218,244,223}3.59 & \cellcolor[RGB]{222,245,226}4.82 & \cellcolor[RGB]{234,249,237}3.89 \\
\midrule
\multicolumn{12}{l}{\textbf{FLUX2-Klein 9B}} \\
\midrule

Baseline & --- & 4.22 & 4.17 & 4.92 & 1.94 & 4.14 & 4.61 & 4.47 & 3.29 & 4.81 & 4.02 \\
DuET ($\mathrm{E}\to\mathrm{T2I}\to\mathrm{E}$) & $[10,16)$ & \cellcolor[RGB]{230,248,234}4.34 & \cellcolor[RGB]{237,250,239}4.29 & \cellcolor[RGB]{240,251,242}4.93 & \cellcolor[RGB]{224,246,228}2.50 & \cellcolor[RGB]{220,245,225}4.36 & \cellcolor[RGB]{243,251,245}4.62 & \cellcolor[RGB]{198,239,206}\textbf{4.58} & \cellcolor[RGB]{226,247,230}3.61 & \cellcolor[RGB]{198,239,206}\textbf{4.92} & \cellcolor[RGB]{216,244,221}4.23 \\
DuET ($\mathrm{E}\to\mathrm{T2I}\to\mathrm{E}$) & $[10,20)$ & \cellcolor[RGB]{198,239,206}\textbf{4.50} & \cellcolor[RGB]{222,245,227}4.39 & \cellcolor[RGB]{255,199,206}4.87 & \cellcolor[RGB]{198,239,206}\textbf{2.98} & \cellcolor[RGB]{198,239,206}\textbf{4.50} & \cellcolor[RGB]{220,245,225}4.64 & \cellcolor[RGB]{255,199,206}4.36 & \cellcolor[RGB]{239,250,241}3.46 & \cellcolor[RGB]{249,253,250}4.82 & \cellcolor[RGB]{198,239,206}\textbf{4.33} \\
DuET ($\mathrm{E}\to\mathrm{T2I}\to\mathrm{E}$) & $[20,30)$ & \cellcolor[RGB]{220,245,225}4.39 & \cellcolor[RGB]{198,239,206}\textbf{4.56} & \cellcolor[RGB]{198,239,206}\textbf{4.96} & \cellcolor[RGB]{231,248,235}2.36 & \cellcolor[RGB]{236,249,238}4.26 & \cellcolor[RGB]{198,239,206}\textbf{4.66} & \cellcolor[RGB]{244,252,246}4.49 & \cellcolor[RGB]{214,243,220}3.74 & \cellcolor[RGB]{218,244,223}4.88 & \cellcolor[RGB]{210,242,217}4.26 \\
DuET ($\mathrm{E}\to\mathrm{T2I}\to\mathrm{E}$) & $[25,35)$ & \cellcolor[RGB]{214,243,220}4.42 & \cellcolor[RGB]{203,240,211}4.52 & \cellcolor[RGB]{212,242,218}4.95 & \cellcolor[RGB]{233,249,236}2.33 & \cellcolor[RGB]{234,249,237}4.27 & \cellcolor[RGB]{209,242,215}4.65 & \cellcolor[RGB]{255,234,237}4.43 & \cellcolor[RGB]{209,242,215}3.80 & \cellcolor[RGB]{223,246,228}4.87 & \cellcolor[RGB]{214,243,220}4.24 \\
$\mathrm{E}\to\mathrm{E}^{*}\to\mathrm{E}$ & $[25,35)$ & \cellcolor[RGB]{240,251,242}4.29 & \cellcolor[RGB]{249,253,249}4.21 & \cellcolor[RGB]{255,232,235}4.90 & \cellcolor[RGB]{248,253,249}2.05 & \cellcolor[RGB]{255,242,244}4.12 & \cellcolor[RGB]{255,199,206}4.55 & \cellcolor[RGB]{234,249,237}4.51 & \cellcolor[RGB]{229,247,232}3.58 & 4.81 & \cellcolor[RGB]{243,251,245}4.08 \\
DuET ($\mathrm{E}\to\mathrm{T2I}\to\mathrm{E}$) & $[35,45)$ & \cellcolor[RGB]{230,248,234}4.34 & \cellcolor[RGB]{218,244,223}4.42 & \cellcolor[RGB]{226,246,230}4.94 & \cellcolor[RGB]{239,250,241}2.23 & \cellcolor[RGB]{250,253,250}4.17 & \cellcolor[RGB]{220,245,225}4.64 & \cellcolor[RGB]{255,219,223}4.40 & \cellcolor[RGB]{198,239,206}\textbf{3.93} & \cellcolor[RGB]{255,231,234}4.78 & \cellcolor[RGB]{225,246,229}4.18 \\
$\mathrm{E}\to\mathrm{E}^{*}\to\mathrm{E}$ & $[35,45)$ & \cellcolor[RGB]{252,254,253}4.23 & \cellcolor[RGB]{244,252,246}4.24 & \cellcolor[RGB]{255,232,235}4.90 & \cellcolor[RGB]{246,252,247}2.09 & \cellcolor[RGB]{255,199,206}4.05 & \cellcolor[RGB]{255,217,222}4.57 & \cellcolor[RGB]{249,253,250}4.48 & \cellcolor[RGB]{252,254,252}3.32 & \cellcolor[RGB]{255,199,206}4.74 & \cellcolor[RGB]{245,252,247}4.07 \\
\midrule
\multicolumn{12}{l}{\textbf{BAGEL}} \\
\midrule

Baseline & --- & 3.38 & 3.51 & 4.43 & 1.38 & 3.00 & 3.93 & 3.64 & 2.53 & 4.22 & 3.32 \\
DuET ($\mathrm{E}\to\mathrm{T2I}\to\mathrm{E}$) & $[10,16)$ & \cellcolor[RGB]{219,245,224}3.53 & \cellcolor[RGB]{255,250,251}3.45 & \cellcolor[RGB]{208,241,215}4.65 & \cellcolor[RGB]{218,244,223}1.54 & \cellcolor[RGB]{237,250,240}3.21 & \cellcolor[RGB]{228,247,232}4.05 & \cellcolor[RGB]{231,248,234}3.77 & \cellcolor[RGB]{204,240,211}2.75 & \cellcolor[RGB]{244,251,245}4.28 & \cellcolor[RGB]{227,247,231}3.44 \\
DuET ($\mathrm{E}\to\mathrm{T2I}\to\mathrm{E}$) & $[10,20)$ & \cellcolor[RGB]{214,243,220}3.55 & \cellcolor[RGB]{255,199,206}2.75 & \cellcolor[RGB]{204,240,211}4.67 & \cellcolor[RGB]{198,239,206}\textbf{1.63} & \cellcolor[RGB]{231,248,234}3.29 & \cellcolor[RGB]{246,252,247}3.97 & \cellcolor[RGB]{245,252,247}3.69 & \cellcolor[RGB]{255,199,206}2.35 & \cellcolor[RGB]{255,217,222}4.20 & \cellcolor[RGB]{248,253,249}3.35 \\
DuET ($\mathrm{E}\to\mathrm{T2I}\to\mathrm{E}$) & $[20,30)$ & \cellcolor[RGB]{224,246,228}3.51 & \cellcolor[RGB]{255,243,245}3.36 & \cellcolor[RGB]{202,240,209}4.68 & \cellcolor[RGB]{225,246,229}1.51 & \cellcolor[RGB]{231,248,234}3.29 & \cellcolor[RGB]{217,244,222}4.10 & \cellcolor[RGB]{218,244,223}3.84 & \cellcolor[RGB]{232,248,235}2.63 & \cellcolor[RGB]{238,250,241}4.31 & \cellcolor[RGB]{223,246,227}3.46 \\
DuET ($\mathrm{E}\to\mathrm{T2I}\to\mathrm{E}$) & $[25,35)$ & \cellcolor[RGB]{205,241,212}3.59 & \cellcolor[RGB]{255,251,251}3.46 & \cellcolor[RGB]{198,239,206}\textbf{4.70} & \cellcolor[RGB]{227,247,231}1.50 & \cellcolor[RGB]{224,246,228}3.38 & \cellcolor[RGB]{198,239,206}\textbf{4.19} & \cellcolor[RGB]{207,241,213}3.90 & \cellcolor[RGB]{255,242,244}2.49 & \cellcolor[RGB]{221,245,225}4.41 & \cellcolor[RGB]{211,242,217}3.51 \\
$\mathrm{E}\to\mathrm{E}^{*}\to\mathrm{E}$ & $[25,35)$ & \cellcolor[RGB]{233,248,236}3.47 & \cellcolor[RGB]{247,252,248}3.54 & 4.43 & \cellcolor[RGB]{255,199,206}1.35 & 3.00 & \cellcolor[RGB]{255,247,248}3.92 & \cellcolor[RGB]{255,236,238}3.63 & \cellcolor[RGB]{200,239,207}2.77 & \cellcolor[RGB]{251,253,251}4.24 & \cellcolor[RGB]{252,254,253}3.33 \\
DuET ($\mathrm{E}\to\mathrm{T2I}\to\mathrm{E}$) & $[35,45)$ & \cellcolor[RGB]{198,239,206}\textbf{3.62} & \cellcolor[RGB]{198,239,206}\textbf{3.73} & \cellcolor[RGB]{219,244,224}4.60 & \cellcolor[RGB]{252,254,253}1.39 & \cellcolor[RGB]{198,239,206}\textbf{3.70} & \cellcolor[RGB]{200,239,207}4.18 & \cellcolor[RGB]{198,239,206}\textbf{3.95} & \cellcolor[RGB]{198,239,206}\textbf{2.78} & \cellcolor[RGB]{198,239,206}\textbf{4.54} & \cellcolor[RGB]{198,239,206}\textbf{3.57} \\
$\mathrm{E}\to\mathrm{E}^{*}\to\mathrm{E}$ & $[35,45)$ & \cellcolor[RGB]{243,251,244}3.43 & \cellcolor[RGB]{255,249,249}3.43 & \cellcolor[RGB]{242,251,244}4.49 & \cellcolor[RGB]{245,252,247}1.42 & \cellcolor[RGB]{254,254,254}3.01 & \cellcolor[RGB]{255,199,206}3.85 & \cellcolor[RGB]{255,199,206}3.61 & \cellcolor[RGB]{241,251,243}2.59 & \cellcolor[RGB]{255,199,206}4.19 & \cellcolor[RGB]{255,199,206}3.30 \\
\bottomrule
\end{tabular}
}
}
\end{table}

\end{document}